%% file: main.tex
\documentclass{article}

\usepackage[preprint]{neurips_2026}

\usepackage[utf8]{inputenc} 
\usepackage[T1]{fontenc}    
\usepackage[colorlinks=true]{hyperref}       
\usepackage{url}            
\usepackage{booktabs}       
\usepackage{amsfonts}       
\usepackage{nicefrac}       
\usepackage{microtype}      
\usepackage{xcolor}         
\definecolor{lightblue}{rgb}{0.1, 0.4, 0.6}
\usepackage{array,multirow}
\usepackage{graphicx}
\usepackage{wrapfig}
\usepackage{amsmath}
\usepackage{amssymb}
\usepackage{booktabs}
\usepackage{multirow}
\usepackage{verbatim}
\usepackage{graphicx}
\usepackage[table]{xcolor} 
\usepackage{colortbl}
\usepackage{array}         
\usepackage{booktabs}
\usepackage{amssymb}
\definecolor{lightblue}{RGB}{86,156,214}
\definecolor{headergray}{RGB}{245,248,252}
\definecolor{fullrow}{RGB}{235,245,255}
\definecolor{lossbg}{gray}{0.96}
\usepackage{pifont}
\usepackage{caption}
\newcommand{\MyTitle}{SS3D}
\usepackage{svg}
\usepackage{placeins}
\usepackage{float}
\usepackage{pifont}
\newcommand{\cmark}{\textcolor{green}{\ding{51}}}
\newcommand{\xmark}{\textcolor{red}{\ding{55}}}

\definecolor{lightblue}{rgb}{0.1, 0.4, 0.6}
\title{\MyTitle: End2End Self-Supervised 3D from Web Videos}

\author{
Marwane Hariat$^{1}$ \quad Gianni Franchi$^{2}$ \quad David Filliat$^{2}$ \quad Antoine Manzanera$^{1}$\thanks{Corresponding author.}\\[2mm]
$^{1}$U2IS, ENSTA -- Institut Polytechnique de Paris, Palaiseau, France\\
$^{2}$Pôle Recherche, Agence Ministérielle pour l'IA de Défense, Palaiseau, France\\
\texttt{\{marwane.hariat, antoine.manzanera, gianni.franchi\}@ensta.fr}\\
\texttt{\{david.filliat\}@polytechnique.edu}
}

\begin{document}

\maketitle

\input{my_sec/abstract}    
\input{my_sec/intro}
\input{my_sec/related_work}

\input{my_sec/method}
\input{my_sec/experiments}

\input{my_sec/conclusion}

\newpage
{
\small
\bibliographystyle{abbrv}
\bibliography{main}
}

\input{my_sec/supp}

\newpage
\newpage

\end{document}

%% file: my_sec/abstract.tex
\begin{abstract}
We present SS3D, a web-scale SfM-based self-supervision pretraining pipeline for feed-forward 3D estimation from monocular video which jointly predicts depth, ego-motion, and intrinsics in a single forward pass.
Scaling SfM self-supervision to unconstrained web video is challenging due to weak multi-view observability and strong corpus heterogeneity; we address these with a multi-view signal proxy (MVSP) used for filtering and curriculum sampling, and with expert training distilled into a single student. Additionally, joint learning in stabilized with an intrinsics-first two-stage schedule. Pretraining on YouTube-8M yields strong cross-domain zero-shot transfer and improved fine-tuning performance over prior self-supervised baselines. We release the pretrained checkpoint and code.
\end{abstract}

%% file: my_sec/intro.tex
\section{Introduction}
\label{sec:intro}


\begin{wrapfigure}{r}{6.5cm}
  \centering
  \includegraphics[width=\linewidth]{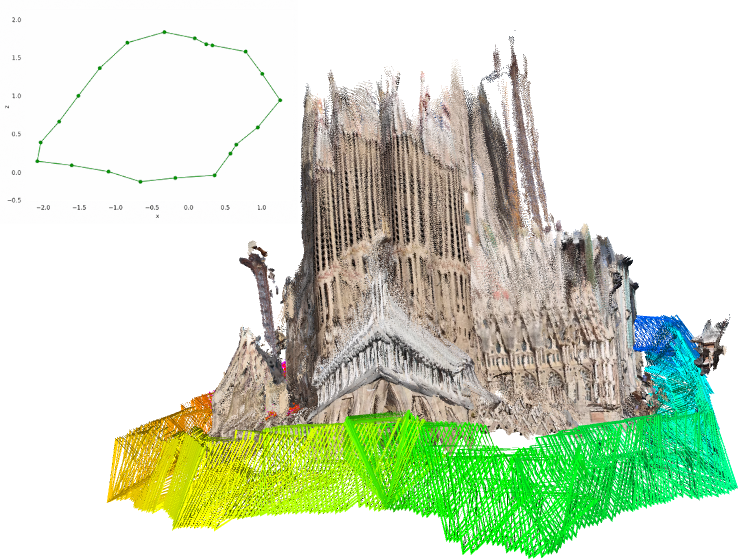}
  \caption{Example 3d reconstruction, poses and odometry predicted by SS3D from the in-the-wild video \url{https://www.youtube.com/watch?v=_M2Q4lfoUHo&t=543s}. }
  \label{fig:sagrada_intro}
\end{wrapfigure}

Recent progress in 3D estimation, such as VGGT~\cite{wang2025vggt} and MapAnything~\cite{keetha2025mapanything}, shows that a single network can infer depth, camera motion, and intrinsics from monocular video without explicit geometric pipelines like SfM \cite{schonberger2016structure}. However, these models rely on large-scale curated datasets with high-quality geometric annotations, whose collection is costly and limits scalability \cite{rohan2025systematicreviewdeepdepth, rajapaksha2024acmsurvey}. This contrasts with 2D and vision-language models, which scale via self-supervised learning on massive uncurated data \cite{oquab2023dinov2, assran2025v}.

Motivated by the scalability of self-supervised learning, we focus on SfM-based \cite{zhou2017unsupervised} (reprojection-based) self-supervision from monocular video, where depth, pose, and intrinsics are learned via view synthesis
and we ask the question: 
\emph{Can a 3D-estimation feed-forward model be trained at scale without any explicit 3D annotations ?}

SfM-based self-supervision ultimately induces a 3D reconstruction from depth, pose, and intrinsics, yet prior work has largely emphasized depth estimation, with pose and intrinsics often treated as supporting variables or assumed known. This makes it unclear whether the learned components can be combined into a coherent end-to-end 3D estimator. We therefore reframe the objective as end-to-end 3D estimation, where the goal is the quality and consistency of the downstream induced 3D reconstruction. Concretely, we train a single model that jointly predicts depth, pose, and intrinsics, and stabilize joint learning with a two-stage schedule: we first learn intrinsics, then freeze them while learning depth and pose. We show that this tri-task optimization yields more consistent 3D reconstructions.

A second challenge is training at scale: naively applying reprojection losses to unconstrained web video is unstable due to weak multi-view signal (e.g., low parallax) and heterogeneous video statistics. We propose a web-scale recipe that stabilizes training along two axes: (i) we introduce a lightweight multi-view signal strength proxy (MVSP) that estimates how well a scene is observed from multiple views, and we shape the MVSP distribution over training by enforcing a minimum threshold via filtering and using a curriculum that gradually broadens the distribution as learning stabilizes, (ii) we address heterogeneity by automatically restructuring the corpus into homogeneous sub-domains where SfM-based self-supervision is reliable, training self-supervised experts on each sub-domain, and distilling them into a single deployable student model.

Using this recipe, we introduce \textit{\MyTitle}, a model pretrained on a multi-million-video corpus derived from YouTube-8M \cite{abu2016youtube} ($\sim$6M videos) and release the resulting weights and code. 
The model achieves strong cross-domain zero-shot transfer and provides a substantially stronger initialization for in-domain fine-tuning than prior self-supervised baselines, supporting both a single foundation model with broad generalization and efficient specialization to target distributions.




\textbf{Contributions.} (1) \textit{\MyTitle}, a unified feed-forward model that jointly predicts depth, ego-motion, and intrinsics (see Fig.\ref{fig:sagrada_intro}), evaluated from a single checkpoint. (2) A web-scale SfM self-supervision pipeline using MVSP-based curriculum sampling and expert distillation to handle data heterogeneity. (3) Release of pretrained models, code, and data processing tools.

%% file: my_sec/related_work.tex
\section{Related works}
\label{sec:related_Works}

\noindent
\textbf{Depth-centric SfM-based self-supervision.} Following the view-synthesis framework of Zhou et al. \cite{zhou2017unsupervised}, many self-supervised methods jointly learn depth and ego-motion (and sometimes intrinsics \cite{gordon2019depth}) by reconstructing a target frame from a source frame using a photometric loss. A large body of work improves robustness through mechanisms such as uncertainty modeling \cite{poggi2020uncertainty}, indoor-specific adaptations \cite{fan2023deeper}, semantic guidance \cite{guizilini2020semantically, li2023learning, saeedan2021boosting, zhu2020edge}, architectural changes \cite{zhao2022monovit, guizilini20203d, zhou_diffnet, he2022ra, watson2021temporal, lyu2021hr, jia2021self}, and handling photometric violations including occlusions \cite{godard2019digging}, dynamic objects \cite{hariat2023rebalancing, lee2021learning, klingner2020self}, and low-texture regions \cite{hariat2025improved, xu2021self, shu2020feature}. Despite this coupled formulation, evaluation is often fragmented: depth, pose, and intrinsics are frequently treated as auxiliary variables, estimated by separate networks and assessed under different protocols. Some works report depth on benchmarks like KITTI, while pose is evaluated using separately trained variants on KITTI Odometry, and intrinsics on yet other datasets such as EuRoC \cite{gordon2019depth}; even when trained jointly, results may rely on task-specific checkpoint selection \cite{ranjan2019competitive}. These practices obscure whether the components form a coherent end-to-end 3D estimator, motivating joint evaluation of depth, pose, and intrinsics from a single feed-forward checkpoint under a unified protocol.

\textbf{Scaling to a large and heterogeneous video corpus.} Recent supervised depth methods have advanced significantly by training on large mixtures of labeled datasets \cite{ranftl2020towards, yang2024depth, piccinelli2024unidepth} and scaling model capacity \cite{ranftl2021vision, bhat2023zoedepth}, while “foundation-style” 3D models leverage extensive annotated multi-domain data to achieve state-of-the-art performance \cite{keetha2025mapanything, wang2025vggt, wang2024dust3r, lin2025depth, wimbauer2025anycam}. In contrast, SfM-based self-supervision remains largely confined to narrowly curated datasets or limited mixtures (e.g., KITTI + Cityscapes \cite{bian2019unsupervised} or indoor-only datasets \cite{chen2020improving}). Scaling such methods to unconstrained, multi-domain web video is challenging for two main reasons. First, reprojection-based supervision is reliable only when sufficient multi-view signal (e.g., parallax) is present, which is often weak or unstable in web video; prior in-the-wild approaches therefore rely on heavy post-processing \cite{chen2019self} or restrict training to specific subsets (e.g., quadcopter footage \cite{gordon2019depth}). Second, in heterogeneous multi-domain corpora, reprojection losses exhibit inconsistent scales and statistics, leading to conflicting gradients that a single model struggles to reconcile \cite{klingner2020self}. These limitations motivate our web-scale approach, which introduces a multi-view signal proxy to drive curriculum-based training and uses expert distillation to address domain heterogeneity. While curriculum learning \cite{wang2023efficienttrain} and mixture-of-experts distillation \cite{xie2024mode} are well-established, we adapt them to the specific failure modes of SfM-based self-supervision on unconstrained web video. We note that dynamic objects are also prevalent in such data, though effective mitigation strategies exist \cite{hariat2023rebalancing, lee2021learning}.

\textbf{Zero-shot Abilities.} Most SfM-based self-supervision methods are developed and evaluated in an in-domain regime: models are typically trained and tested on the same benchmark distribution (e.g., KITTI for outdoor driving \cite{zou2018df, yin2018geonet}, NYUv2 for indoor \cite{fan2023deeper}), or on closely related datasets (e.g., Cityscapes→KITTI \cite{hariat2023rebalancing, li2021unsupervised} or NYUv2→ScanNet \cite{hariat2025improved, li2022monoindoor++}), with the goal of maximizing benchmark performance and producing a benchmark-specialized model. As a consequence, robustness to large domain shifts has not been a primary design target in this literature, and performance often degrades out-of-domain \cite{pinard2023does, chen2020improving}. In contrast, our multi-domain web-video pretraining yields a single released model that exhibits strong cross-domain zero-shot transfer.

\textbf{In-Domain specialization.} Fine-tuning to a specific target domain remains essential in many practical settings, yet traditional SfM-based self-supervision pipelines are commonly trained from scratch or initialized from generic visual pretraining \cite{hariat2023rebalancing, godard2019digging, zhao2022monovit} (e.g., ImageNet \cite{russakovsky2015imagenet} or more recently from vision transformer models \cite{dosovitskiy2020image, oquab2023dinov2, he2022masked}). As a result, these models start with limited transferable 3D geometric prior, which compromises accuracy and sample-efficiency. We find that our pretrained model consistently improves in-domain fine-tuning performance compared to prior self-supervised baselines.

%% file: my_sec/method.tex
\section{Method}
\label{sec:method}

\begin{figure}[t]
  \centering
  \includegraphics[width=\linewidth, height=7.5cm]{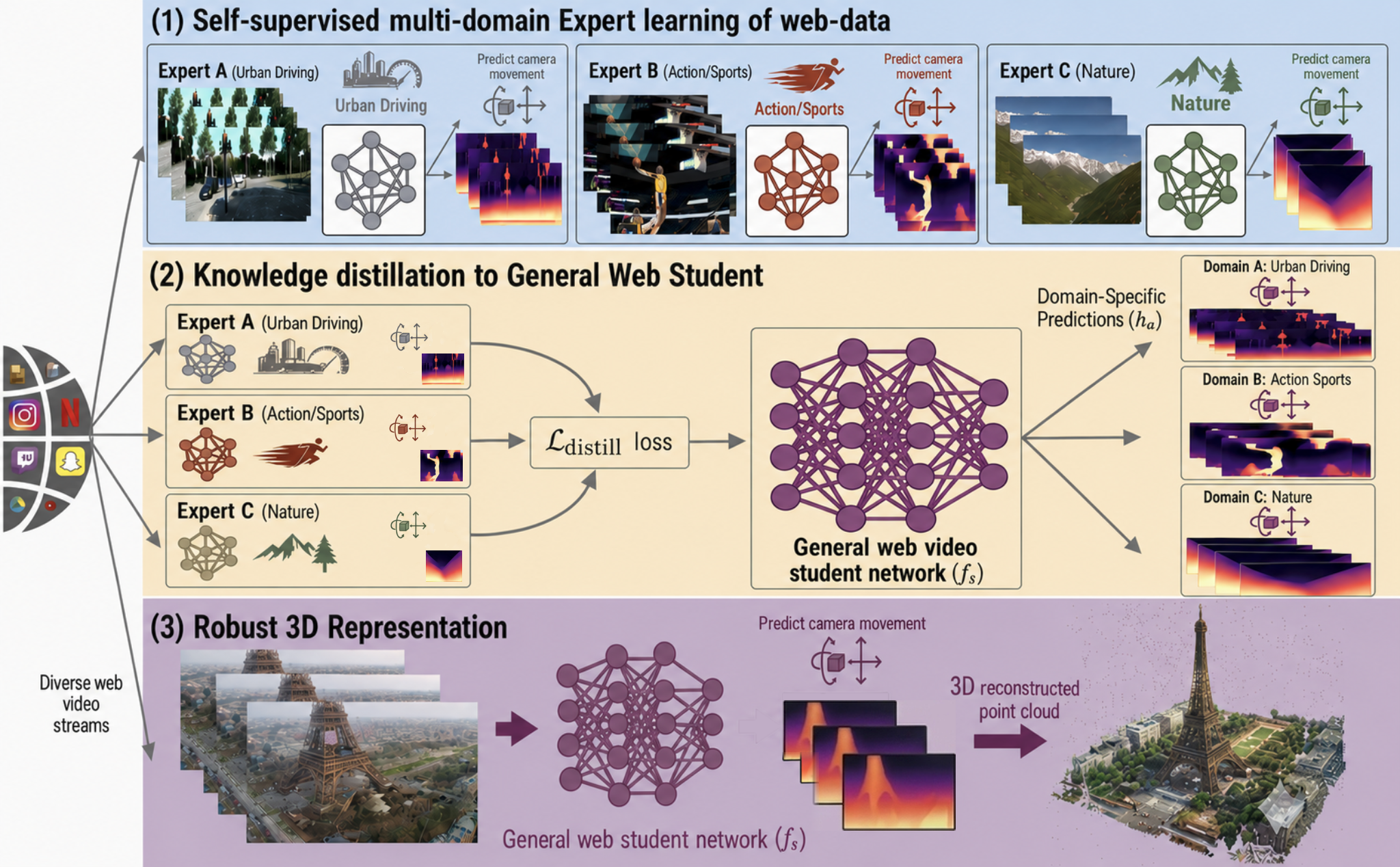}
  \caption{\textbf{Overview of \MyTitle} (1) Train self-supervised, sub-domain experts on web video. (2) Distill expert into a single student model. (3) At inference, student predicts depth, pose, and intrinsics, inducing a 3D reconstruction (e.g., point cloud on bottom right). Domain names (sport, driving,...) are chosen for illustration and may not match the unsupervised clustering used in training.}
  \label{fig:full_pipeline}
\end{figure}

Our objective is to learn via structure-from-motion (SfM) self-supervision (see Sec.~\ref{sec:method:problem}) a single feed-forward model that jointly estimates depth, camera motion, and intrinsics from raw video (Sec.~\ref{sec:method:unified_3d}).
Our approach (Fig.~\ref{fig:full_pipeline}) consists of three key components:
(i) a unified 3D estimator trained with multi-view photometric supervision,
(ii) a web-scale stabilization strategy based on multi-view observability,
and (iii) expert distillation to address large-scale domain heterogeneity.

\subsection{Problem Setup and Self-Supervised Objective}
\label{sec:method:problem}

Let $I_t : \Omega \rightarrow \mathbb{R}^3$ denote a RGB image at time $t$, defined over pixel domain $\Omega \subset \mathbb{R}^2$. A pixel $p \in \Omega$ is written in homogeneous coordinates $p = (u,v,1)^\top$. We denote $D_t(p) \in \mathbb{R}_{+}$ the predicted depth at pixel $p$, $T_t = [R_t \mid t_t] \in SE(3)$ the camera pose with $R_t \in SO(3)$ and $t_t \in \mathbb{R}^3$, and $K_t \in \mathbb{R}^{3\times 3}$ the camera intrinsics matrix, assuming a pinhole camera with centered principal point:
\begin{equation}
K_t =
\begin{bmatrix}
f_x^t & 0 & W/2 \\
0 & f_y^t & H/2 \\
0 & 0 & 1
\end{bmatrix}.
\end{equation}

A pixel $p$ from an image is lifted to 3D in camera coordinates as:
\begin{equation}
X_t(p) = D_t(p) K_t^{-1} p \in \mathbb{R}^3.
\label{eq:backprojection}
\end{equation}

Given two frames $i$ and $j$, we compute the relative transformation as:
\begin{equation}
T_{i \rightarrow j} = 
T_j T_i^{-1}.
\end{equation}

We can then synthesize frame $I_i$ from $I_j$ via differentiable warping:
\begin{equation}
\hat{I}_{i}(p)
=
I_j \!\left(
\pi\!\left(
K_j T_{i\rightarrow j} X_i(p)
\right)
\right),
\label{eq:view_synthesis}
\end{equation}
where $\pi(x,y,z) = (x/z, y/z)$ is perspective projection and we use bilinear interpolation.

\paragraph{Photometric Self-Supervision.}
The reconstruction loss between synthesized and target images is:
\begin{equation}
\Psi_{i\rightarrow j}
=
\sum_{p \in \Omega}
m_{i,j}(p)
\rho\big(
\hat{I}_i(p) - I_i(p)
\big),
\label{eq:photo_loss}
\end{equation}
where $\rho(\cdot)$ iss a robust penalty (Charbonnier + SSIM)\cite{wang2004image},
and \(m_{i,j}\) is a mask function to invalidate unreliable correspondences, computed by comparing the reconstructions induced by depth + pose reprojection with those induced by optical flow \cite{hariat2023rebalancing}. This masking is important as large discrepancies indicate pixels where the geometric warp is unreliable, such as dynamic objects that are frequent in unconstrained videos, and are therefore excluded from the photometric loss.

Unlike classical pairwise training, we sample all ordered pairs $(i,j)$ within a clip of length $N$, leading to the multi-view objective which implicitly reconstructs a consistent 3D scene across the clip:
\begin{equation}
\Psi = 
\sum_{i \neq j}
\Psi_{i\rightarrow j}.
\end{equation}

\subsection{Unified 3D Foundation Model}
\label{sec:method:unified_3d}

Prior SfM-based self-supervision methods typically use separate networks for depth and pose and assume known intrinsics. In contrast, we train a \emph{single feed-forward transformer} that jointly predicts depth, pose, and intrinsics and always use the same checkpoint to evaluate all components:
\begin{equation}
f_\theta\big(\{I_i\}_{i=1}^{N}\big)
=
\{(D_i, T_i, K_i)\}_{i=1}^{N}.
\label{eq:model}
\end{equation}

Intrinsics are predicted per frame but assumed constant within a clip (see below).

\paragraph{Two-Stage Optimization.}
Joint optimization of $(D,T,K)$ is unstable. We therefore  adopt a two stage approach:
 (\textbf{Stage 1})~We begin with a brief warm-up that trains depth and pose while keeping intrinsics fixed to default calibration; and then optimize intrinsics head while freezing depth and pose.
  (\textbf{Stage 2})~Freezing intrinsics, we optimize depth and pose.
At inference and during Stage 2, we use averaged intrinsics for each clip:
$\bar{K} = \frac{1}{N}\sum_{i=1}^{N} K_i$.



\subsection{Scaling Self-Supervision to Web-Scale Video}
\label{sec:method:scaling}

Naively scaling this model to unconstrained web video fails (as shown on Fig.~\ref{fig:naive_scaling} and Tab.~\ref{table:ablation_parallax_curr_kitti_nyu}) for two main reasons. \textbf{(A) Weak geometric signal.}
Many videos contain little useful multi-view information. For example, the camera may be almost static, or it may only rotate or zoom. In such cases, reprojection provides weak or misleading training signals. \textbf{(B) Strong domain heterogeneity.}
Web videos cover many different scenes, motions, and camera settings. As a result, loss statistics vary significantly across videos, and naïvely mixing everything together often leads to unstable training.\\
We address (A) by estimating the \emph{multi-view observability} of each video using a simple geometric score. This score is used to filter and schedule training data through a curriculum.  
We address (B) by training domain-specific experts and then distilling them into a single model.

\paragraph{\textbf{Multi-View Signal Proxy (MVSP)}:}
\label{sec:method:mvs}

We introduce 
this simple score
to measure whether a video contains enough parallax to provide useful geometric supervision. Intuitively, when there is strong parallax, correspondences between frames are better explained by an epipolar model (fundamental matrix $F$) than by a planar model (homography $H$). 
Let $\mathcal{C}$ be our video corpus and $v \in \mathcal{C}$ a video.  
For two consecutive frames $(I_t, I_{t+1})$, we extract $M$ correspondences:
$\{(x_i^t, x_i^{t+1})\}_{i=1}^M$,
and use them to
estimate a fundamental matrix $F$ and a homography $H$. We compute the geometric error $d_F(i)$ and $d_H(i)$ of each model:
\begin{equation}
d_F(i) =
\frac{
({x_i^{t+1}}^\top F x_i^t)^2
}{
(F x_i^t)_1^2 + (F x_i^t)_2^2
+
(F^\top x_i^{t+1})_1^2 + (F^\top x_i^{t+1})_2^2}\\
 \end{equation}
 \begin{equation}
d_H(i) =
\|x_i^{t+1} - H x_i^t\|_2^2
+
\|x_i^t - H^{-1} x_i^{t+1}\|_2^2.
\end{equation}

A frame-level parallax score is computed by the ratio $\frac{r_H}{r_F}$, where $r_H$ and $r_F$ are the average errors over the $M$ correspondences for Homography and Fundamental matrix respectively, and finally the video-level $\text{MVSP}(v)$ score is the temporal average of $\frac{r_H}{r_F}$ over the frames of $v$.
High MVSP indicates strong multi-view geometry.
Low MVSP indicates weak 
translation
or mostly planar 
scene. 

\paragraph{\textbf{Curriculum sampling}.} We use MVSP to guide training: First, we remove videos with extremely weak geometric signal.  
Then, we train using high-MVSP videos first, and progressively include lower-MVSP videos as training becomes stable.
Let $F(\cdot)$ be the cumulative distribution function of MVSP over the corpus.  
For a threshold $\alpha \in [0,1]$, we define:
\begin{equation}
\mathcal{S}_\alpha
=
\{ v \in \mathcal{C} \mid F(\text{MVSP}(v)) \ge \alpha \}.
\end{equation}
Training starts with large $\alpha$ (strong parallax only) and gradually decreases $\alpha$ to include harder videos.

\subsection{Distillation of Experts}
\label{sec:method:distillation}

Even with MVSP-based curriculum, web videos remain very diverse. Training a single model directly on the full mixture can be unstable, as different domains may interfere.
We address this in two steps.

\paragraph{\textbf{Domain-Specific Experts.}} We automatically partition the corpus $\mathcal{C}$ into $K$ more homogeneous subsets $\{\mathcal{C}_k\}_{k=1}^K$, using a simple K-Means clustering on CLIP features (more details in Sec.~\ref{sec:method:youtube} and Appendix~\ref{sec:supp:ytb_preprocessing}).
For each subset $\mathcal{C}_k$, we train an expert model $\mathcal{E}_k$ using the same self-supervised objective.  
Since each subset is more consistent, training is more stable and each expert can specialize to its domain.

\paragraph{\textbf{Distillation into a Single Student.}} We then train a single student model on the full corpus. For each training clip, we: (1) identify its sub-domain $k$, (2) obtain the expert predictions $(D^*, T^*, K^*)$, (3) use them as soft targets.
The student is trained with the photometric loss plus a distillation loss: $\mathcal{L} = \Psi + \lambda_{\text{distill}} \mathcal{L}_{\text{distill}}$.
The distillation loss encourages the student to match the expert predictions up to scale ambiguity:
\begin{equation}
\mathcal{L}_{\text{distill}} =
\kappa(D, D^*)
+
\kappa(t, t^*)
+
\|R^\top R^* - I\|_F
+
\|K^{-1}K^* - I\|_F.
\end{equation}
where $\kappa(\cdot,\cdot)$ is a scale-and-shift invariant matching function similar to \cite{ranftl2020towards}. At inference time, we discard experts and use only the distilled student model.

\subsection{Training on YouTube-8M}
\label{sec:method:youtube}
Our training pipeline combines the unified 3D estimator (Sec.~\ref{sec:method:unified_3d}) with MVSP-based filtering/curriculum (Sec.~\ref{sec:method:mvs}) and expert distillation (Sec.~\ref{sec:method:distillation}), yielding a single deployable student model trained entirely without ground-truth 3D labels. Below we describe the details used to train \MyTitle\ at scale on YouTube-8M to produce our pretrained checkpoint.

\textbf{Implementation details}
Our architecture follows a VGGT-style transformer design, initialized with a pretrained DINOv2 image encoder, and uses three lightweight heads to predict depth, ego-motion, and intrinsics. The network is composed of 24 attention blocks, each containing a frame-wise self-attention layer and a global self-attention layer. Following the ViT-L configuration used in DINOv2, every layer has 16 attention heads of dimension 1024.  To improve training stability, 
QKNorm and LayerScale are incorporated in each layer, initializing the LayerScale parameters to 0.01.

For image tokenization, we use DINOv2 features and augment them with positional embeddings. Tokens from the 4th, 11th, 17th, and 23rd blocks are forwarded to DPT for the upsampling stage.\
We train on \(518 \times 518\) crops and sample clips at 10 FPS. Videos are segmented into short clips using shot boundary detection, which supports our assumption of approximately constant intrinsics within each clip. 
After shot segmentation and MVSP-based filtering, our training set contains \textasciitilde100M frames.

We form \(K=5\) sub-domains by clustering the corpus videos using K-Means over frame-level CLIP~\cite{radford2021learning} embeddings, averaged per clip.  We use a sequence length of \(N=8\). We use a batch size 8 composed of 4 different sub-domains and we follow the method proposed in \cite{sener2018multi} to compute Pareto-optimal gradients. We first train the \(K\) experts and then distill them into a single student that share the same architecture.  Experts were trained on 2 NVIDIA RTX A6000 GPUs, and the student was trained on 64 H100 GPUs. We use standard augmentations (horizontal flips; random brightness/contrast/saturation/hue) following \cite{godard2019digging}. 
Training uses the Adam optimiser~\cite{kingma2014adam} with \(\beta_{1} = 0.99\) and \(\beta_{2} = 0.999\).  We use \(\lambda_{\text{distill}} = 0.2\).
More details in Appendix~\ref{sec:supp:ytb_preprocessing}.


%% file: my_sec/experiments.tex
\vspace{-0.3cm}

\section{Experiments}
\label{sec:experiments}
We report performance of {\MyTitle}  across the three outputs our model jointly predicts: depth, camera motion, and intrinsics, using a unified protocol. In particular, all zero-shot results are produced from a single YouTube-8M pretrained checkpoint, and fine-tuning results start from that same checkpoint unless stated otherwise. This section validates the three claims of the paper: (i) training a unified depth–pose–intrinsics estimator yields more coherent induced 3D reconstructions, (ii) stable web-scale training requires MVSP-aware sampling and heterogeneity handling, and (iii) the resulting pretrained model transfers zero-shot across domains and provides a strong fine-tuning initialization

\vspace{-0.2cm}
\subsection{Preliminaries}
\noindent
{\bf Data:}
We evaluate SS3D on benchmarks spanning diverse scenarios, using the commonly used evaluation protocol for each dataset. For depth, we focus on \textbf{KITTI} \cite{geiger2013vision} and \textbf{NYUv2} \cite{silberman2012indoor}, the most widely used outdoor and indoor depth benchmarks. We intentionally pair them because they differ sharply in scene scale, camera motion patterns, and intrinsics, making them a strong test of cross-domain generalization.
For camera motion, we evaluate in more challenging regimes: Sintel~\cite{butler2012naturalistic} (synthetic, complex motion and appearance changes) and TUM-RGBD \cite{sturm2012benchmark}  (real-world, dynamic scenes).


\begin{figure}[htb]
  \centering
  \includegraphics[width=\linewidth, height=3cm]{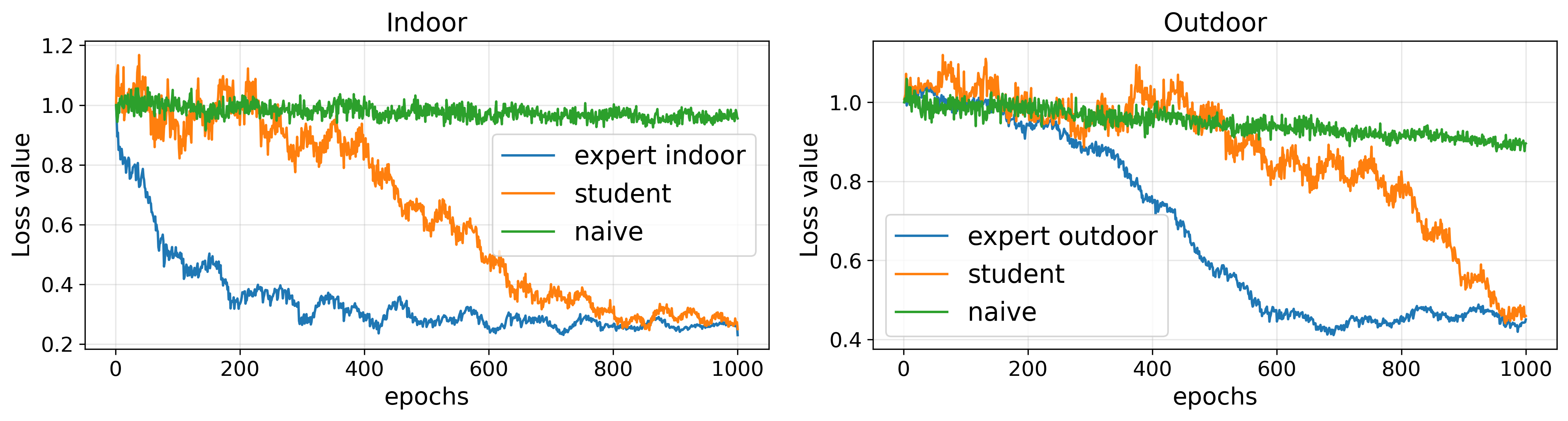}
  \vspace{-13pt}
  \caption{{\bf Training dynamics} 
  on two
  contexts. We can see that naively scaling to more data yields few to no gains. Experiments done with distillation of two experts: indoor and outdoor.}
  \label{fig:naive_scaling}
\end{figure}

\noindent
{\bf Naïve scaling fails without stability mechanisms:}
Before presenting final results, we first validate a core motivation of {\MyTitle}: simply feeding more web videos to SfM-based self-supervision does not reliably improve performance. Fig.~\ref{fig:naive_scaling} illustrates this failure mode: when trained naively on a heterogeneous mixture, optimization becomes unstable, and the model struggles to learn from all regimes simultaneously. In contrast, handling heterogeneity with expert training followed by distillation enables the student to improve on both indoor and outdoor contexts. Results in Tab.~\ref{table:ablation_parallax_curr_kitti_nyu} quantify this limitation : simply pre-training on YouTube-8M (line +YT) show limited benefits (vs baseline B2).

\begin{table}[htb]
\centering
\setlength{\tabcolsep}{2.2pt}
\renewcommand{\arraystretch}{0.98}
\scriptsize

\newcommand{\gcheck}{\textcolor{green!60!black}{\ensuremath{\checkmark}}}

\resizebox{0,7\columnwidth}{!}{%
\begin{tabular}{c c c c c  c c  c c}
\toprule
\rowcolor{headergray}
\textbf{ID} &
\textbf{MVSP+Curr} &
$\boldsymbol{\mathcal{L}_{\text{distill}}}$ &
\textbf{YTB8M} &
\textbf{Unified 3D} &
\multicolumn{2}{c}{\textcolor{lightblue}{\textbf{KITTI Abs Rel} $\downarrow$}} &
\multicolumn{2}{c}{\textcolor{lightblue}{\textbf{NYU Abs Rel} $\downarrow$}} \\
\rowcolor{headergray}
 & & & & & \textbf{ZS} & \textbf{FT} & \textbf{ZS} & \textbf{FT} \\
\midrule

B1 &  &  &  &  & -- & 0.082 & -- & 0.115 \\
B2 &  &  &  &  & -- & 0.080 & -- & 0.117 \\
+U &  &  &  & \gcheck & -- & 0.078 & -- & 0.111 \\
+YT &  &  & \gcheck & \gcheck & 0.19 & 0.079 & 0.26 & 0.112 \\
+D &  & \gcheck & \gcheck & \gcheck & 0.101 & 0.072 & 0.125 & 0.098 \\
\textbf{Full} & \gcheck & \gcheck & \gcheck & \gcheck & \textbf{0.092} & \textbf{0.064} & \textbf{0.116} & \textbf{0.090} \\
\bottomrule
\end{tabular}%
}

\vspace{5pt}
\caption{\textbf{Ablation study} on KITTI and NYU in zero-shot (ZS) and fine-tuning (FT) w/ VGGT architecture (B2). B1 follows \cite{hariat2025improved} with a slightly different head. B1/B2: no pretraining on YTB8M (FT only), no distillation, no MVSP+Curr, no Unified 3D. +U: add Unified 3D. +YT: naive YTB8M mixing. +D: add distillation. Full: add MVSP+Curr filtering.}
\label{table:ablation_parallax_curr_kitti_nyu}
\end{table}

\vspace{-0.2cm}
\subsection{Ablation: each component matters}
Tab.~\ref{table:ablation_parallax_curr_kitti_nyu} decomposes the full pipeline and shows that performance improves monotonically as we add our proposed components: the unified estimator (+U), distillation (+D), and finally MVSP-based filtering and curriculum (Full). Importantly, improvements are visible both in zero-shot transfer (ZS) and after fine-tuning (FT), demonstrating that our method is not merely “more data,” but rather a recipe that makes web-scale SfM training reliable.

\begin{table}[b]
\centering

\begin{minipage}[t]{0.45\linewidth}
    \centering
    \small
    \renewcommand{\arraystretch}{1.15}
\begin{tabular}{lcccc}
\toprule
\textbf{Variant} 
& \textbf{AbsRel $\downarrow$} 
& \textbf{$\delta_1$ $\uparrow$} 
\\
\midrule
No strategy \cite{hariat2025improved}                  
& 0.104 
& 0.885 
\\
Strong Parallax only
    
& 0.126 
& 0.862 
\\
Full Curriculum (Ours)        
& \textbf{0.098}
& \textbf{0.890}
\\
\bottomrule
\end{tabular}
\vspace{5pt}
\caption{\textbf{Ablation of our MVSP-based curriculum.} 
$\Delta$ values denote improvement relative to the no-curriculum baseline. Strong Parallax only refers to an intuitive strategy consisting of training only successive frames with high parallax.}
\label{tab:mvs_curriculum}
\end{minipage}
\hfill
\begin{minipage}[t]{0.52\linewidth}
    \centering
    \small
    \renewcommand{\arraystretch}{1.15}
    \begin{tabular}{lcc}
        \toprule
        \textbf{\# Experts} & \textbf{AbsRel $\downarrow$} & \textbf{\boldmath{$\delta_1$} (\%) $\uparrow$} \\
        \midrule
        Baseline      & 0.080 & 0.928 \\
        No expert     & 0.079 & 0.930 \\
        \(2\) Experts & 0.069 & 0.939 \\
        \(5\) Experts & \textbf{0.064} & \textbf{0.946} \\
        \bottomrule
    \end{tabular}
    \vspace{5pt}
    \caption{\textbf{Expert ablation.} Ablation on the number of experts distilled to the student during pretraining on YouTube8M. Performance is evaluated by fine-tuning and evaluating on KITTI. The baseline corresponds to no pretraining on YouTube8M.}
    \label{tab:expert_ablation}
\end{minipage}

\end{table}

We evaluate the effect of our MVSP-based curriculum in isolation using a standard SfM-based reprojection baseline (ResNet-50, trained in-domain on KITTI). Importantly, we do not use any other component of SS3D (no unified depth–pose–in\-trinsics transformer, no experts, no distillation). The only change is to replace uniform sampling with our MVSP-based curriculum, which prioritizes high-parallax frame pairs early in training and gradually includes lower-parallax pairs. Even on top of a strong baseline \cite{hariat2025improved}, the curriculum yields consistent improvements.
Tab.~\ref{tab:mvs_curriculum} presents the results of this study. 
We compare three strategies: the original baseline without any sampling strategy, a simple heuristic that only selects frame pairs with high parallax, and our full curriculum.The results highlight several observations. 

First, simply selecting frame pairs with high parallax (\textit{Strong Parallax only}) does not improve performance. 
In fact, it degrades both the depth error (AbsRel) and the accuracy metric ($\delta_1$). 
This suggests that restricting training to only large-parallax pairs reduces the diversity of training examples.
In contrast, our full curriculum improves performance over the baseline. 
It reduces the depth error 
and slightly increases the accuracy. 
These results indicate that gradually introducing different levels of parallax during training provides better supervision and leads to more stable learning.

Tab.~\ref{tab:expert_ablation} further quantifies the role of experts and distillation: increasing the number of experts improves downstream fine-tuning performance, consistent with the hypothesis that experts reduce cross-domain interference during pretraining and that distillation successfully consolidates these domain-specific cues into a single student. Marginal improvements were observed above 5 experts, on top of computational resources constraints.

\begin{table*}[htb]
\centering
\setlength{\tabcolsep}{2.2pt}
\renewcommand{\arraystretch}{1.05}
\scriptsize

\begin{minipage}[t]{0.49\textwidth}

\resizebox{\linewidth}{!}{%
\begin{tabular}{l c c c c c c c c}
\toprule
\multirow{2}{*}{\textbf{Method}} 
& \multirow{2}{*}{\textbf{Self-Sup}}
& \multicolumn{4}{c}{\textit{Lower is better} $\downarrow$} 
& \multicolumn{3}{c}{\textit{Higher is better} $\uparrow$} \\
\cmidrule(r){3-6} \cmidrule(l){7-9}
& 
& \textcolor{lightblue}{Abs Rel} 
& Sq Rel 
& RMSE 
& \textcolor{lightblue}{RMSE log}
& \textcolor{lightblue}{$\delta_{1}$} 
& $\delta_{2}$ 
& $\delta_{3}$ \\
\midrule

DPT~\cite{ranftl2021vision}             & \xmark & 0.100 & \texttt{N/A} & \texttt{N/A} & \texttt{N/A} & 0.901 & \texttt{N/A} & \texttt{N/A} \\
MidASv3.1~\cite{birkl2023midas}         & \xmark & 0.127 & \texttt{N/A} & \texttt{N/A} & \texttt{N/A} & 0.850 & \texttt{N/A} & \texttt{N/A} \\
DepthAnything~\cite{yang2024depth}      & \xmark & 0.076 & \texttt{N/A} & \texttt{N/A} & \texttt{N/A} & 0.947 & \texttt{N/A} & \texttt{N/A} \\
VGGT(Depth+Cam)~\cite{wang2025vggt}     & \xmark & 0.093 & \texttt{N/A} & \texttt{N/A} & \texttt{N/A} & 0.917 & \texttt{N/A} & \texttt{N/A} \\
\midrule
\textbf{SS3D (Ours)}                           & \cmark & 0.092 & 0.678 & 4.016 & 0.166 & 0.928 & 0.968 & 0.984 \\
\bottomrule
\end{tabular}%
}
\caption{\textbf{KITTI ZS} - Zero-shot on KITTI. Light-blue metrics are the most challenging ones. }
\label{table:results_depth_kitti_zs}


\end{minipage}
\hfill
\begin{minipage}[t]{0.49\textwidth}

\resizebox{\linewidth}{!}{%
\begin{tabular}{l c c c c c c c}
\toprule
\multirow{2}{*}{\textbf{Method}} 
& \multirow{2}{*}{\textbf{Self-Sup}}
& \multicolumn{3}{c}{\textit{Lower is better} $\downarrow$} 
& \multicolumn{3}{c}{\textit{Higher is better} $\uparrow$} \\
\cmidrule(r){3-5} \cmidrule(l){6-8}
& 
& \textcolor{lightblue}{Abs Rel} 
& RMSE
& \textcolor{lightblue}{RMSE log}
& \textcolor{lightblue}{$\delta_{1}$} 
& $\delta_{2}$ 
& $\delta_{3}$ \\
\midrule

DPT~\cite{ranftl2021vision}             & \xmark & 0.098 & \texttt{N/A} & \texttt{N/A} & 0.903 & \texttt{N/A} & \texttt{N/A} \\
MiDaS v3.1~\cite{birkl2023midas}        & \xmark & 0.048 & \texttt{N/A} & \texttt{N/A} & 0.980 & \texttt{N/A} & \texttt{N/A} \\
DepthAnything~\cite{yang2024depth}      & \xmark & 0.043 & \texttt{N/A} & \texttt{N/A} & 0.981 & \texttt{N/A} & \texttt{N/A} \\
VGGT (Depth+Cam)~\cite{wang2025vggt}    & \xmark & 0.036 & \texttt{N/A} & \texttt{N/A} & 0.980 & \texttt{N/A} & \texttt{N/A} \\
\midrule
\textbf{SS3D (Ours)}                    & \cmark & 0.116 & 0.523 & 0.055 & 0.822 & 0.947 & 0.967 \\
\bottomrule
\end{tabular}%
}
\caption{\textbf{NYU ZS} -- Zero-shot evaluation on NYUv2. N/A: metrics non available from the authors' paper.}
\label{table:results_depth_nyu_zs}

\end{minipage}
\end{table*}

\begin{table*}[hb]
\centering
\resizebox{\textwidth}{!}{%
\begin{tabular}{lcccccccccccc}
\toprule
\textbf{Methods} & \textbf{00} & \textbf{01} & \textbf{02} & \textbf{03} & \textbf{04} & \textbf{05} & \textbf{06} & \textbf{07} & \textbf{08} & \textbf{09} & \textbf{10} & \textbf{Avg.} \\
\midrule
\textit{Seq. Length (m)} 
& 3724.19m & 2453.20m & 5067.23m & 560.89m & 393.65m 
& 2205.58m & 1232.88m & 649.70m & 3222.80m & 1705.05m & 919.52m & -- \\
\midrule

DROID-SLAM \cite{teed2021droid}
& 92.10 & 344.60 & 107.61 & 2.38 & 1.00 & 118.50 & 62.47 & 21.78 & 161.60 & 72.32 & 118.70 & 100.28 \\
DPV-SLAM \cite{lipson2024deep}
& 112.80 & 11.50 & 123.53 & 2.50 & 0.81 & 57.80 & 54.86 & 18.77 & 110.49 & 76.66 & 13.65 & 53.03 \\
DPV-SLAM++ \cite{lipson2024deep}
& 8.30 & 11.86 & 39.64 & 2.50 & 0.78 & 5.74 & 11.60 & 1.52 & 110.90 & 76.70 & 13.70 & 25.75 \\
\midrule

InfiniteVGGT \cite{yuan2026infinitevggt}
& 186.46 & 623.62 & 289.16 & 166.74 & 68.00 & 143.84 & 117.57 & 85.33 & 221.56 & 215.41 & 156.92 & 206.78 \\
CuT3R \cite{wang2025continuous}
& 190.38 & 90.59 & 264.39 & 20.40 & 7.31 & 92.25 & 67.54 & 22.48 & 145.08 & 67.42 & 40.00 & 91.62 \\
TTT3R \cite{chen2025ttt3r}
& 119.94 & 99.59 & 238.07 & 16.83 & 3.98 & 36.38 & 47.20 & 11.62 & 107.33 & 86.96 & 33.58 & 72.86 \\
LoGeR \cite{zhang2026loger}
& 62.34 & 41.64 & 39.64 & 4.89 & 1.82 & 41.27 & 13.99 & 16.24 & 26.46 & 22.71 & 8.84 & 25.44 \\
VGGT-Long \cite{deng2025vggt}
& 11.97 & 40.51 & 49.85 & 5.41 & 2.86 & 9.88 & 6.07 & 3.47 & 66.27 & 32.27 & 21.33 & 22.718 \\
{\bf SS3D (Ours)}
& 11.94 & 48.57 & 49.74 & 8.56 & 2.00 & 8.10 & 7.70 & 2.57 & 57.05 & 24.76 & 14.04 & \textbf{21.37} \\
\bottomrule
\end{tabular}%
}
\caption{
{\bf Odometry} - Comparison of Absolute Trajectory Error (ATE$\downarrow$, m) on KITTI. The upper block reports optimization-based methods, while the lower block reports feed-forward methods, which are the most directly comparable to ours.
}
\label{tab:kitti_ate}
\end{table*}

\begin{table}[t]
\centering
\setlength{\tabcolsep}{2.5pt} 
\renewcommand{\arraystretch}{1.05}
\scriptsize
\resizebox{0,85\columnwidth}{!}{%
\begin{tabular}{l l c c c c c c c c}
\toprule
Category & Method & No sup. & Time &
\multicolumn{3}{c}{Sintel} &
\multicolumn{3}{c}{TUM-RGBD (dyn.)} \\
\cmidrule(lr){5-7}\cmidrule(lr){8-10}
& & & &
ATE$\downarrow$ & RPE$_t\downarrow$ & RPE$_r\downarrow$ &
ATE$\downarrow$ & RPE$_t\downarrow$ & RPE$_r\downarrow$ \\
\midrule
& AnyCam w/o ref. & \xmark & $<20$s & 0.099 & 0.045 & \textbf{0.567} & 0.095 & \textbf{0.025} & \textbf{1.050} \\
& Ours w/o ref.   & \cmark & $<20$s & \textbf{0.090} & \textbf{0.043} & 0.601 & \textbf{0.092} & 0.026 & 1.064 \\
\bottomrule
\end{tabular}%
}
\vspace{5pt}
\caption{Pose estimation comparison to AnyCam~\cite{wimbauer2025anycam} in zero-shot. Absolute trajectory error (ATE) and relative pose error for translation (RPE$_t$) and rotation (RPE$_r$) on Sintel and TUM-RGBD.}
\label{tab:pose}
\vspace{-10pt}
\end{table}

\vspace{-0.2cm}
\subsection{Zero-shot cross-domain transfer}
We now evaluate generalization without any target-domain training. To ensure comparison fairness, let us clarify that, at inference, depth is predicted on single frames independently.

\noindent
{\bf Depth:}
Tabs.~\ref{table:results_depth_kitti_zs} and ~\ref{table:results_depth_nyu_zs} show that \MyTitle\ achieves strong zero-shot depth estimation across both outdoor and indoor domains. While fully supervised methods remain an upper bound, our self-supervised pretrained model approaches the performance of strong supervised baselines, and does so without any 3D ground truth. Compared to prior self-supervised methods (which are typically evaluated in-domain), our zero-shot results are already competitive, indicating that web-scale pretraining learns transferable geometric priors rather than benchmark-specific shortcuts. The fact that this holds on both KITTI and NYUv2, despite their substantial differences in scene scale and camera intrinsics, supports our main claim that \MyTitle\ learns domain-robust geometry from raw video.

\noindent
{\bf Camera motion:}
We further report odometry results in Tab.~\ref{tab:kitti_ate}, showing that \textbf{SS3D} achieves clear gains over supervised baselines and remains highly competitive with feed-forward 3D models. 
(qualitative results are provided in Appendix).
These results show that \textbf{SS3D} accurately captures the overall camera trajectory, even on long sequences spanning several kilometers. Tab.~\ref{tab:pose} reports zero-shot pose results on Sintel and dynamic TUM-RGBD and compares to AnyCam, which is designed to leverage strong pretrained depth/flow estimators. \MyTitle\ is competitive with AnyCam, despite being trained fully self-supervised and predicting depth, pose, and intrinsics jointly.

\noindent
{\bf Intrinsics:}
Fig.~\ref{fig:intrinsic_stability} shows focal estimation on Sintel (quantitative results in appendix). \MyTitle\ 
achieves competitive focal estimation accuracy without any calibration supervision. These results further reinforce the strength of our whole framework and show that we can infer camera intrinsics quite precisely 
from raw monocular videos.

\begin{figure}[H]
    \centering
    \includegraphics[width=\textwidth]{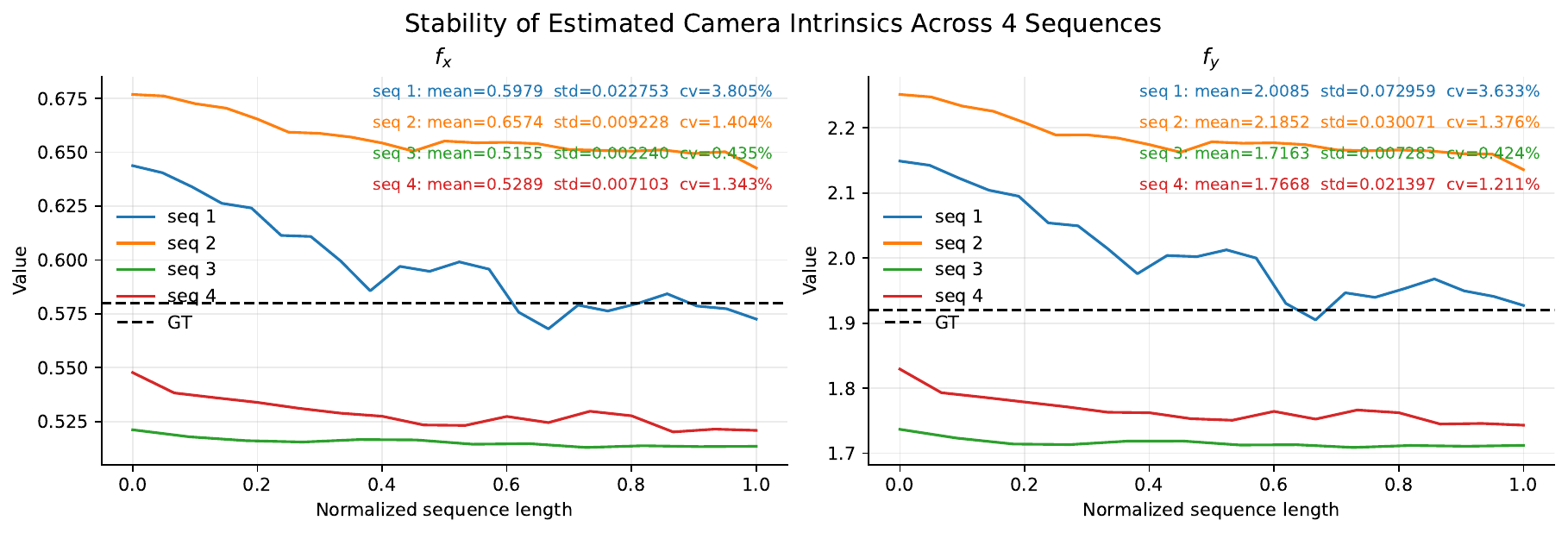}
    \caption{Stability of the learned intrinsic parameters across four sequences on Kitti. Dashed lines indicate the ground-truth values for $f_x$ and $f_y$.}
    \label{fig:intrinsic_stability}
\end{figure}

\noindent
{\bf 3D reconstruction:}
We evaluate the reconstructed 3D geometry by aligning the predicted point cloud to the ground truth scale and global pose (see protocol in Appendix \ref{sec:app:pointcloudquality}) and  compute the bidirectional RMSE, i.e., the nearest-neighbor RMSE from prediction to ground truth and reversely, and report their average as the final reconstruction error. Results are shown in Tab.~\ref{tab:seq_results}. 

\begin{table}[htb]
\centering
\resizebox{0.7\textwidth}{!}{%
\begin{tabular}{l|c|c|c|c}
\hline
Method & Sequence 1 & Sequence 2 & Sequence 3 & Sequence 4 \\
\hline
FeatDepth & 6.29 & 5.76 & 4.27 & 7.16 \\
MonoDepth & 5.18 & 4.20 & 4.43 & 3.52 \\
SS3D      & \textbf{2.63} & \textbf{2.95} & \textbf{3.80} & \textbf{2.31} \\
\hline
\end{tabular}
}
\vspace{5pt}
\caption{Per-sequence bidirectional RMSE between predicted and ground-truth point clouds after ICP alignment (in meters). } \label{tab:seq_results}
\vspace{-10pt}
\end{table}

\vspace{-0.2cm}
\subsection{Further results}

\noindent
{\bf In-domain fine-tuning and specialization:}
We now evaluate whether our pretraining also improves the standard in-domain benchmark protocol. In these experiments, we initialize from the YouTube-8M pretrained checkpoint and fine-tune on the target dataset.

\begin{table*}[hb]
\centering
\setlength{\tabcolsep}{2.2pt}
\renewcommand{\arraystretch}{1.05}
\scriptsize

\begin{minipage}[t]{0.49\textwidth}

\resizebox{\linewidth}{!}{%
\begin{tabular}{l c c c c c c c c}
\toprule
\multirow{2}{*}{\textbf{Method}} 
& \multirow{2}{*}{\textbf{Self-Sup}}
& \multicolumn{4}{c}{\textit{Lower is better} $\downarrow$} 
& \multicolumn{3}{c}{\textit{Higher is better} $\uparrow$} \\
\cmidrule(r){3-6} \cmidrule(l){7-9}
& 
& \textcolor{lightblue}{Abs Rel} 
& Sq Rel 
& RMSE 
& \textcolor{lightblue}{RMSE log}
& \textcolor{lightblue}{$\delta_{1}$} 
& $\delta_{2}$ 
& $\delta_{3}$ \\
\midrule

Monodepth2~\cite{godard2019digging}              & \cmark & 0.110 & 0.831 & 4.642 & 0.187 & 0.883 & 0.962 & 0.982 \\
MonoViT~\cite{zhao2022monovit}                   & \cmark & 0.099 & 0.708 & 4.372 & 0.175 & 0.900 & 0.967 & 0.984 \\
HR-Depth~\cite{lyu2021hr}                        & \cmark & 0.109 & 0.792 & 4.632 & 0.185 & 0.884 & 0.962 & 0.983 \\
RA-Depth~\cite{he2022ra}                         & \cmark & 0.096 & 0.613 & 4.216 & 0.171 & 0.903 & 0.968 & 0.985 \\
DIFFNet~\cite{zhou_diffnet}                      & \cmark & 0.102 & 0.764 & 4.483 & 0.180 & 0.896 & 0.965 & 0.983 \\
Hariat \textit{et al.}~\cite{hariat2025improved} & \cmark & 0.082 & 0.604 & 4.108 & 0.162 & 0.928 & 0.968 & 0.985 \\
\midrule
Adabins~\cite{bhat2021adabins}                   & \xmark & 0.058 & \texttt{N/A} & 2.360 & 0.088 & 0.964 & 0.995 & 0.999 \\
DPT~\cite{ranftl2021vision}                      & \xmark & 0.062 & \texttt{N/A} & 2.573 & 0.092 & 0.959 & 0.9995 & 0.999 \\
DepthAnything~\cite{yang2024depth}            & \xmark & 0.046 & \texttt{N/A} & 1.896 & 0.069 & 0.982 & 0.9998 & 0.9999 \\
\midrule
\textbf{SS3D (Ours)}                                    & \cmark & 0.064 & 0.530 & 3.212 & 0.138 & 0.946 & 0.977 & 0.986 \\
\bottomrule
\end{tabular}%
}
\caption{\textbf{KITTI FT} - Fine-tuning on KITTI~\cite{geiger2013vision} and evaluating on KITTI. }
\label{table:results_depth_kitti_ft}

\end{minipage}
\hfill
\begin{minipage}[t]{0.49\textwidth}

\resizebox{\linewidth}{!}{%
\begin{tabular}{l c c c c c c c}
\toprule
\multirow{2}{*}{\textbf{Method}} 
& \multirow{2}{*}{\textbf{Self-Sup}}
& \multicolumn{3}{c}{\textit{Lower is better} $\downarrow$} 
& \multicolumn{3}{c}{\textit{Higher is better} $\uparrow$} \\
\cmidrule(r){3-5} \cmidrule(l){6-8}
& 
& \textcolor{lightblue}{Abs Rel} 
& RMSE
& \textcolor{lightblue}{RMSE log}
& \textcolor{lightblue}{$\delta_{1}$} 
& $\delta_{2}$ 
& $\delta_{3}$ \\
\midrule

MovingIndoor~\cite{zhou2019moving}                    & \cmark & 0.208 & 0.712 & 0.086 & 0.674 & 0.900 & 0.968 \\
StructDepth~\cite{li2021structdepth}                  & \cmark & 0.140 & 0.540 & 0.060 & 0.817 & 0.955 & 0.988 \\
MonoIndoor++~\cite{li2022monoindoor++}                & \cmark & 0.132 & 0.517 & \texttt{N/A} & 0.834 & 0.961 & 0.990 \\
IndoorDepth~\cite{fan2023deeper}                      & \cmark & 0.126 & 0.494 & 0.054 & 0.845 & 0.965 & 0.991 \\
Hariat \textit{et al.}~\cite{hariat2025improved}      & \cmark & 0.115 & 0.458 & 0.054 & 0.859 & 0.970 & 0.992 \\
\midrule
Adabins~\cite{bhat2021adabins}                        & \xmark & 0.103 & 0.364 & 0.044 & 0.903 & 0.984 & 0.997 \\
DPT~\cite{ranftl2021vision}                           & \xmark & 0.110 & 0.357 & 0.045 & 0.904 & 0.988 & 0.998 \\
DepthAnything~\cite{yang2024depth}                    & \xmark & 0.056 & 0.206 & 0.024 & 0.984 & 0.998 & 1.000 \\
\midrule
\textbf{SS3D (Ours)}                                         & \cmark & 0.090 & 0.418 & 0.049 & 0.866 & 0.970 & 0.992 \\
\bottomrule
\end{tabular}%
}
\caption{\textbf{NYU FT} - Fine-tuning on NYUv2~\cite{silberman2012indoor} and evaluating on NYUv2.}
\label{table:results_depth_nyu_ft}

\end{minipage}
\end{table*}

Tabs.~\ref{table:results_depth_kitti_ft} and \ref{table:results_depth_nyu_ft} show that \MyTitle\      
provides a substantially stronger initialization than prior self-supervised baselines, yielding clear gains after fine-tuning on both KITTI and NYUv2. Notably, these improvements hold in both outdoor and indoor settings, suggesting that the pretrained model captures transferable geometric structure that improves sample efficiency and final accuracy. 

\noindent
{\bf Qualitative Results:}
The
depth prediction and the point-clouds resulting from the joint depth/pose/intrinsics predictions 
show
that SS3D produces sharper depth and, more importantly, substantially cleaner and more geometrically consistent 3D reconstructions than the baselines. 
Significant examples from the KITTI sequences and from casual internet videos
can be found in Appendix.

%% file: my_sec/conclusion.tex
\vspace{-0.3cm}
\section{Conclusion}

In this work, we have presented SS3D, the first framework to successfully scale self-supervised depth, pose, and intrinsics estimation using a web-scale video dataset (YouTube-8M). 
Our experiments demonstrate that while "more data" is often viewed as a panacea, naïve scaling in the context of Structure-from-Motion (SfM) self-supervision leads to optimization instability and poor generalization due to scene heterogeneity. Ultimately, SS3D proves that the vast, non curated repository of web videos can serve as a powerful signal for learning universal geometric priors that lead to state of the art performance. This shifts the bottleneck of 3D computer vision from expensive 3D annotated data collection to the intelligent curation and stabilization of existing video data. 


%% file: my_sec/supp.tex
\clearpage

\appendix

\section{Qualitative results}

\begin{figure}[h]
  \centering
  \includegraphics[width=\linewidth]{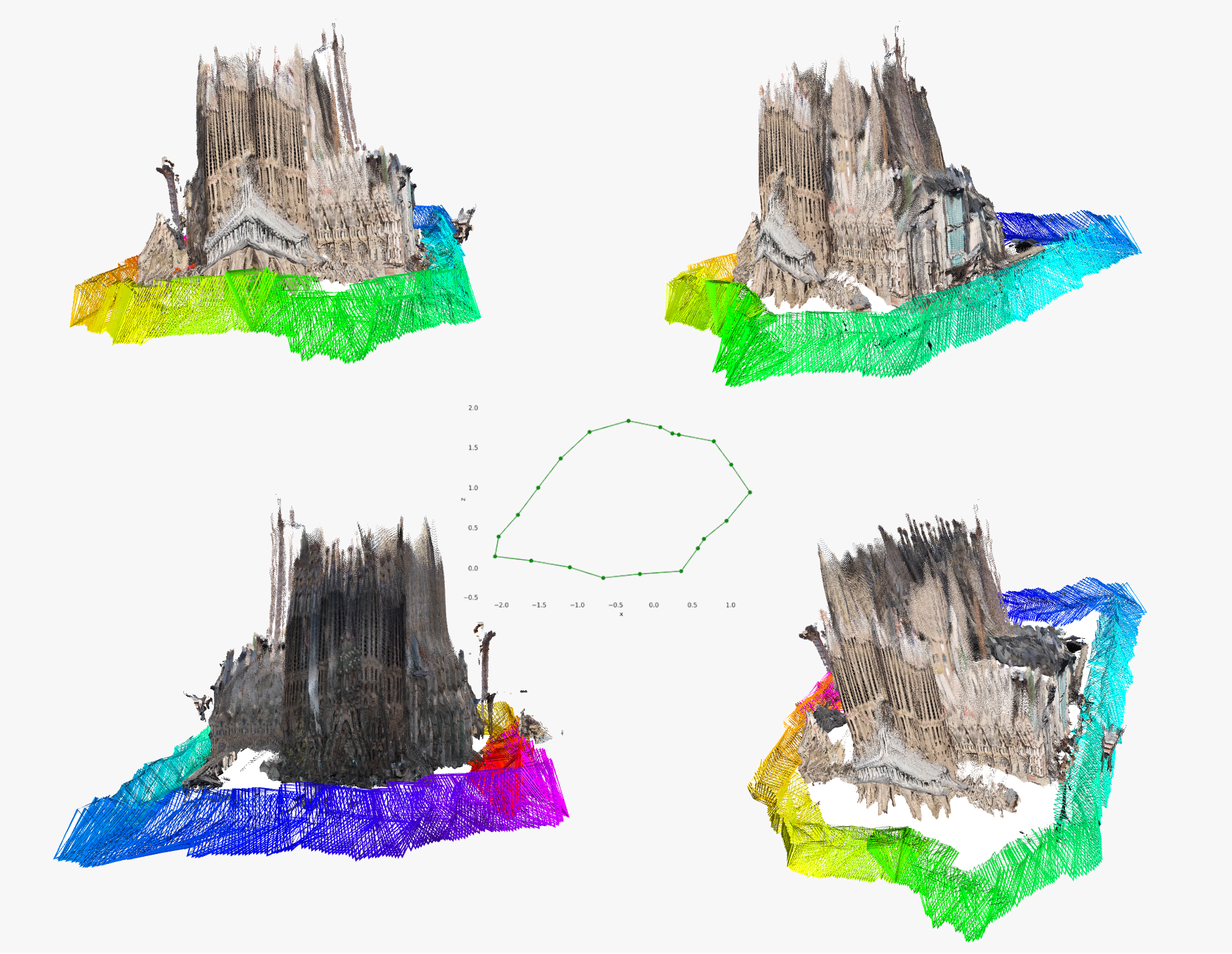}
  \caption{Example of a 3d reconstruction obtained by SS3D from the video \url{https://www.youtube.com/watch?v=_M2Q4lfoUHo&t=543s}. The coloured cones around the Sagrada Familia show the predicted camera poses, while the trajectory plot at the centre displays the predicted odometry. }
  \label{fig:sagrada}
\end{figure}

Fig. \ref{fig:sagrada} shows an example reconstruction obtained by SS3D.

\begin{figure}[H]
\centering
\setlength{\tabcolsep}{2pt}
\renewcommand{\arraystretch}{1.0}

\begin{tabular}{c c c c}
\scriptsize RGB &
\scriptsize &
\scriptsize &
\scriptsize \\
\includegraphics[width=0.22\linewidth]{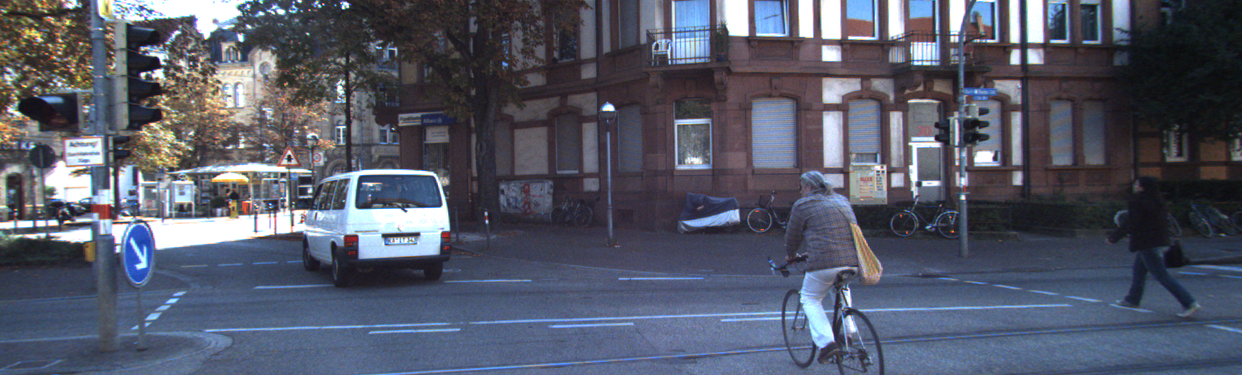} &
\includegraphics[width=0.22\linewidth]{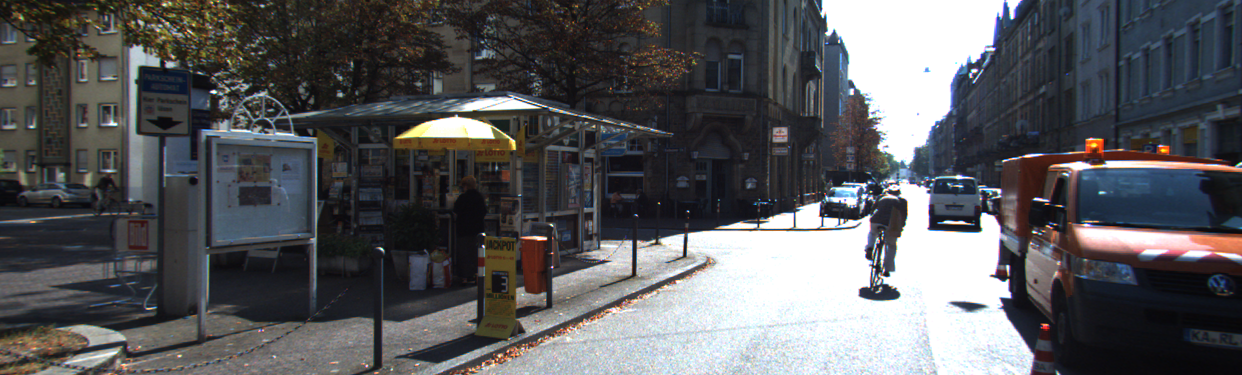} &
\includegraphics[width=0.22\linewidth]{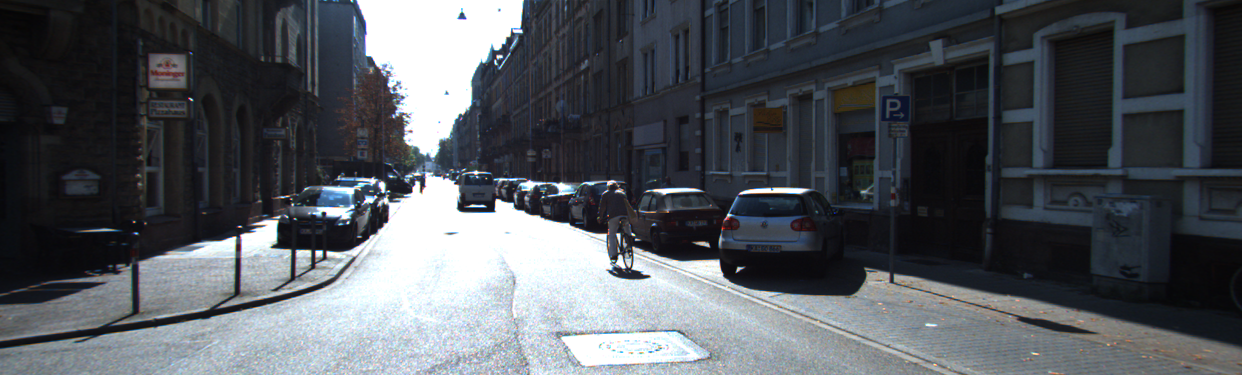} &
\includegraphics[width=0.22\linewidth]{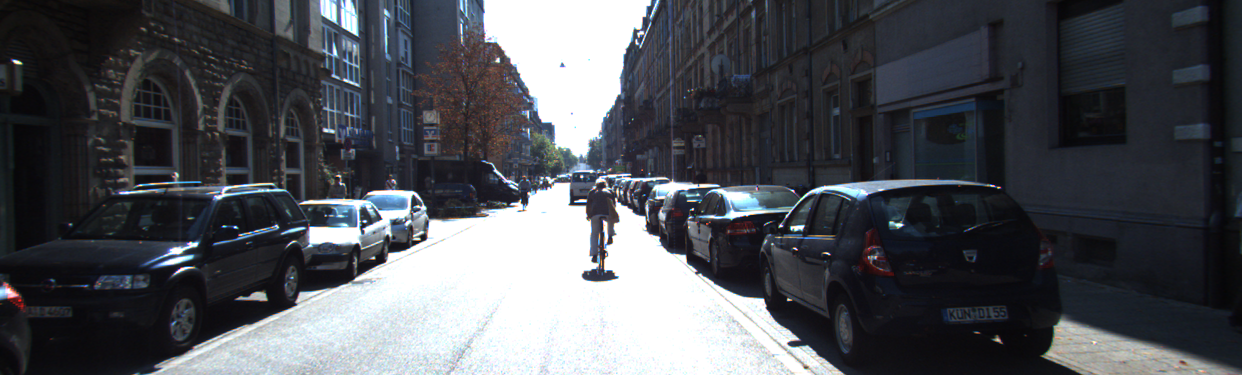} \\[1mm]

\scriptsize Monodepth2 &
\scriptsize &
\scriptsize &
\scriptsize \\
\includegraphics[width=0.22\linewidth]{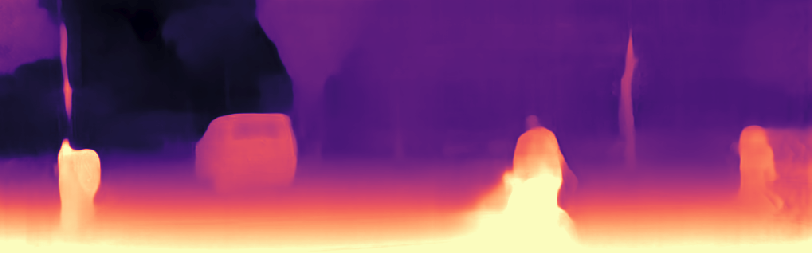} &
\includegraphics[width=0.22\linewidth]{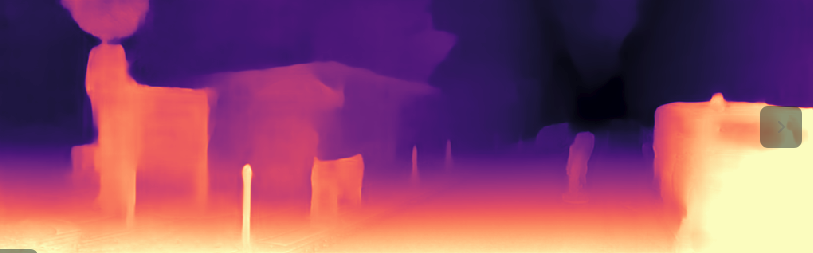} &
\includegraphics[width=0.22\linewidth]{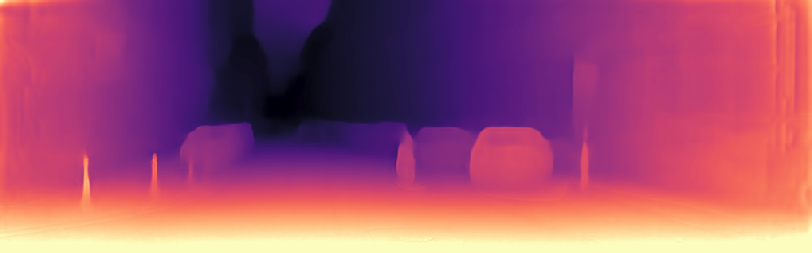} &
\includegraphics[width=0.22\linewidth]{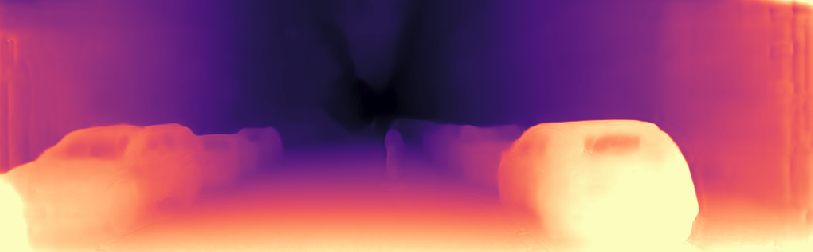} \\[1mm]

\scriptsize FeatDepth &
\scriptsize &
\scriptsize &
\scriptsize \\
\includegraphics[width=0.22\linewidth]{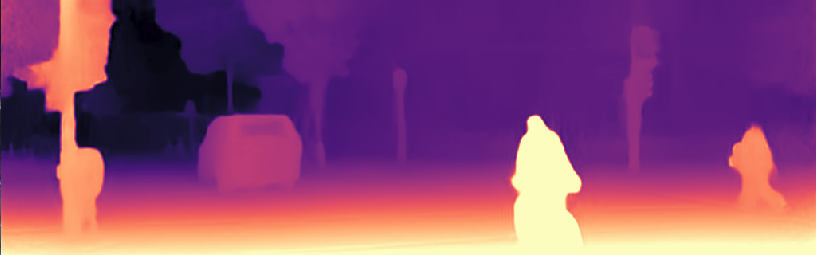} &
\includegraphics[width=0.22\linewidth]{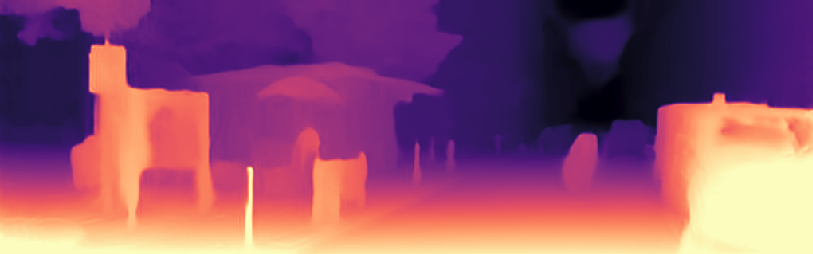} &
\includegraphics[width=0.22\linewidth]{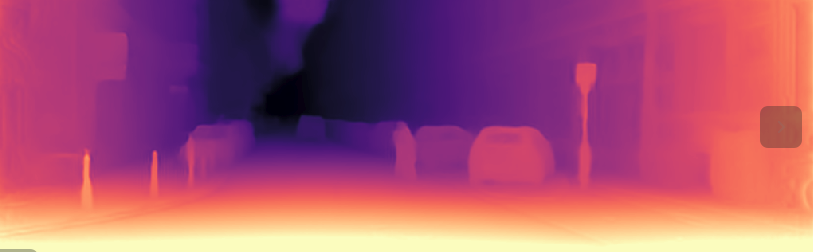} &
\includegraphics[width=0.22\linewidth]{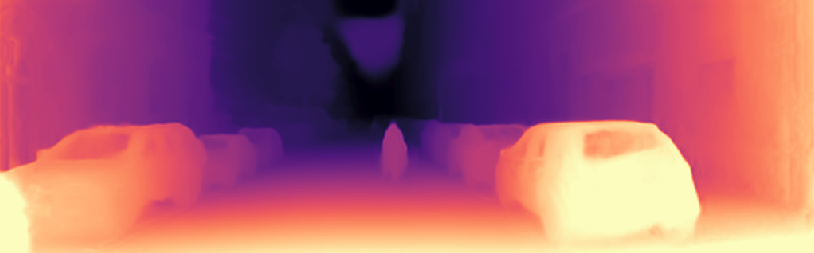} \\[1mm]

\scriptsize SS3D &
\scriptsize &
\scriptsize &
\scriptsize \\
\includegraphics[width=0.22\linewidth]{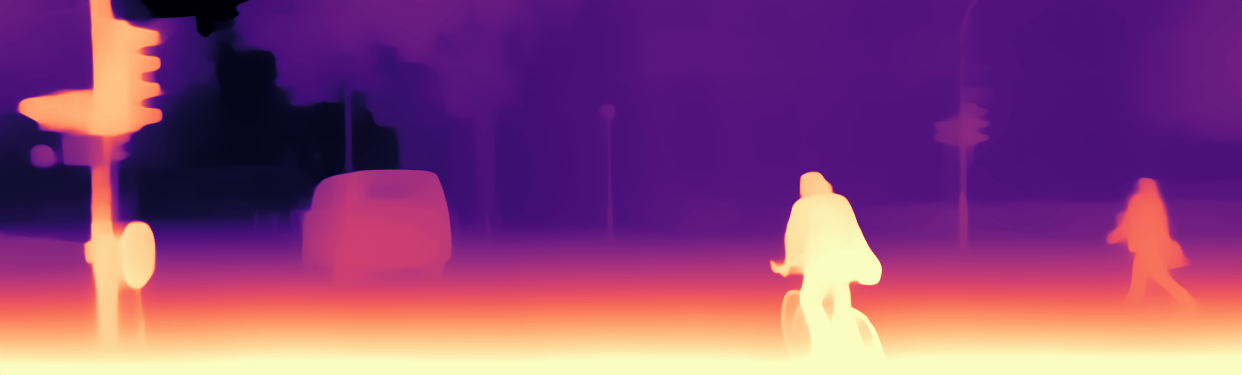} &
\includegraphics[width=0.22\linewidth]{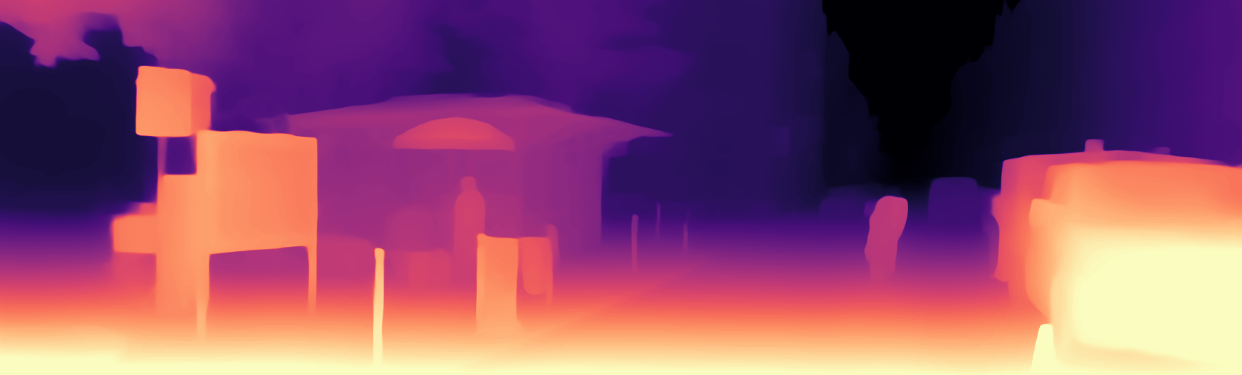} &
\includegraphics[width=0.22\linewidth]{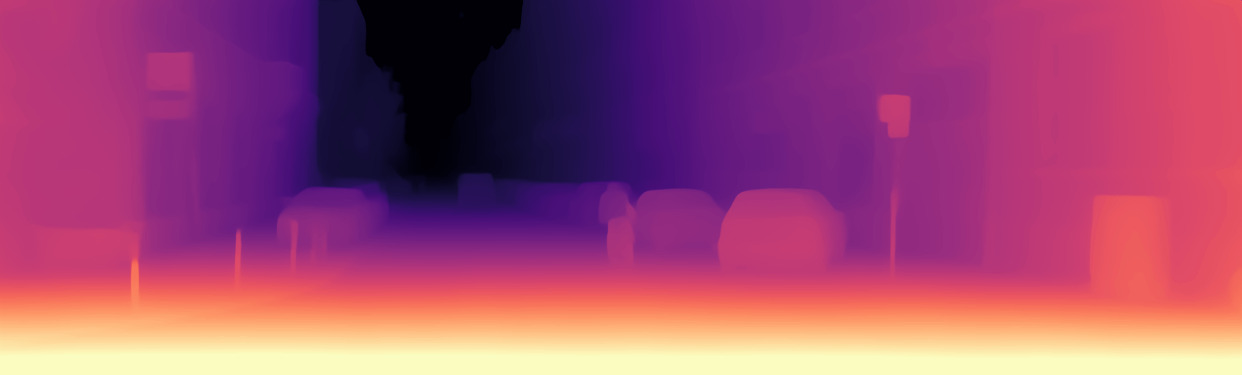} &
\includegraphics[width=0.22\linewidth]{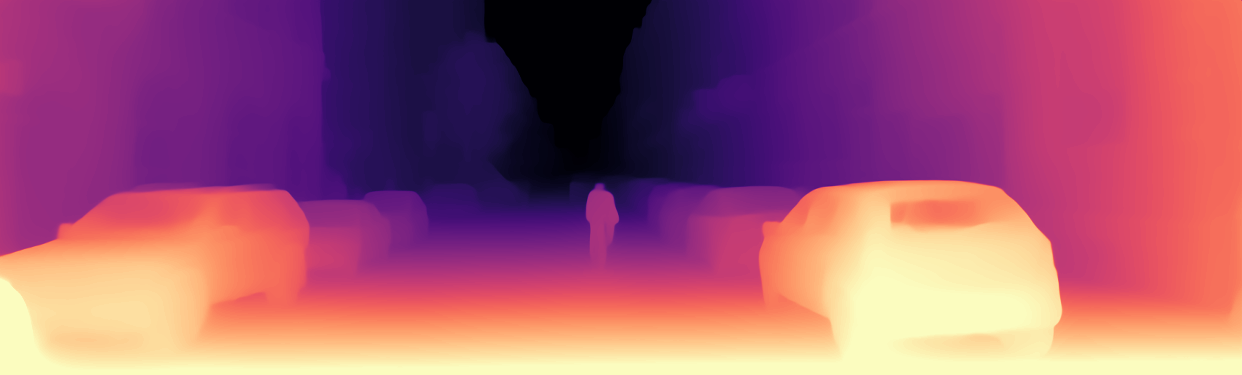} \\[1mm]

\end{tabular}

\vspace{2mm}

\begin{tabular}{cc}

\includegraphics[width=0.48\linewidth]{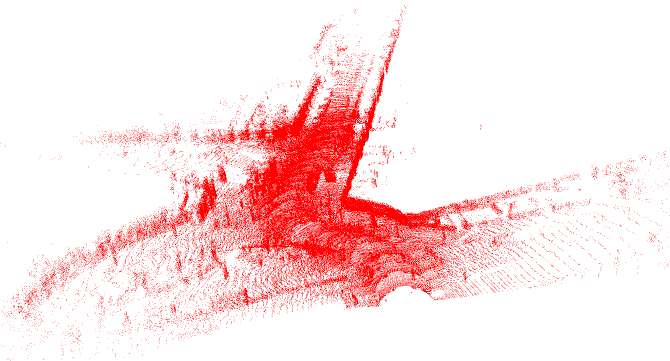} &
\includegraphics[width=0.48\linewidth]{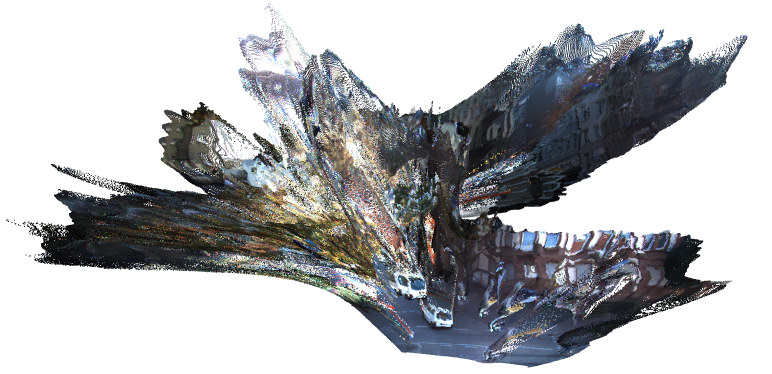} \\[1mm]
\includegraphics[width=0.48\linewidth]{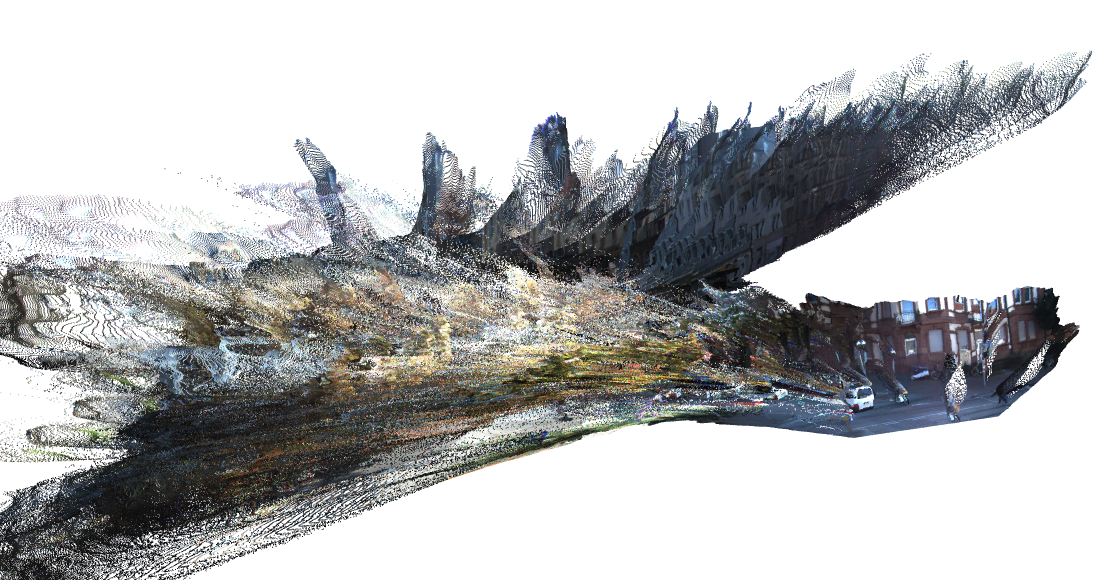} &
\includegraphics[width=0.48\linewidth]{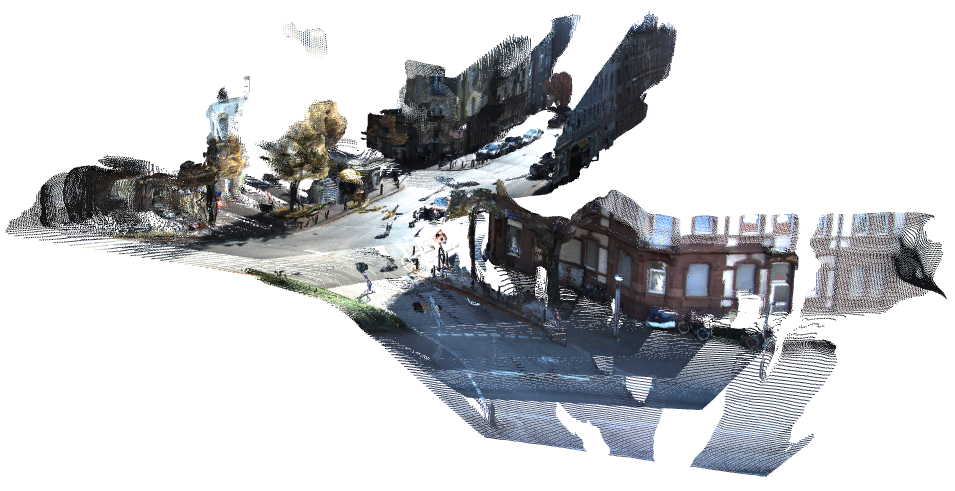}
\end{tabular}

\caption{\textbf{Sequence 1}. Qualitative comparison to strong self-supervised baselines FeatDepth~\cite{shu2020feature} and Monodepth2~\cite{godard2019digging}. \textbf{Top}: predicted depth maps. \textbf{Bottom}: 3D point clouds induced by the predicted depth, pose, and intrinsics. In the bottom block, the LiDAR point cloud is shown in red (top-left), with Monodepth2 (top-right), FeatDepth (bottom-left), and SS3D (bottom-right).}
\label{fig:qualitative_seq1}
\end{figure}

\begin{figure}[H]
\centering
\setlength{\tabcolsep}{2pt}
\renewcommand{\arraystretch}{1.0}

\begin{tabular}{c c c c}
\scriptsize RGB &
\scriptsize &
\scriptsize &
\scriptsize \\
\includegraphics[width=0.22\linewidth]{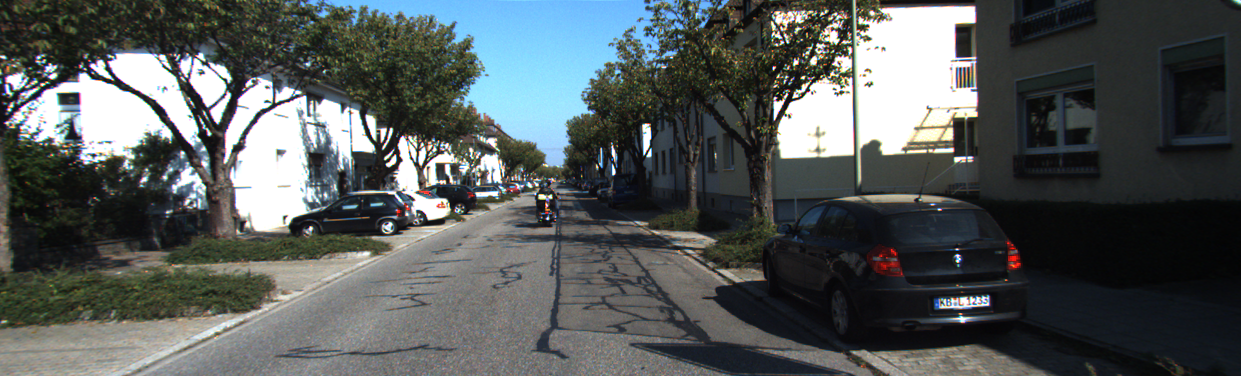} &
\includegraphics[width=0.22\linewidth]{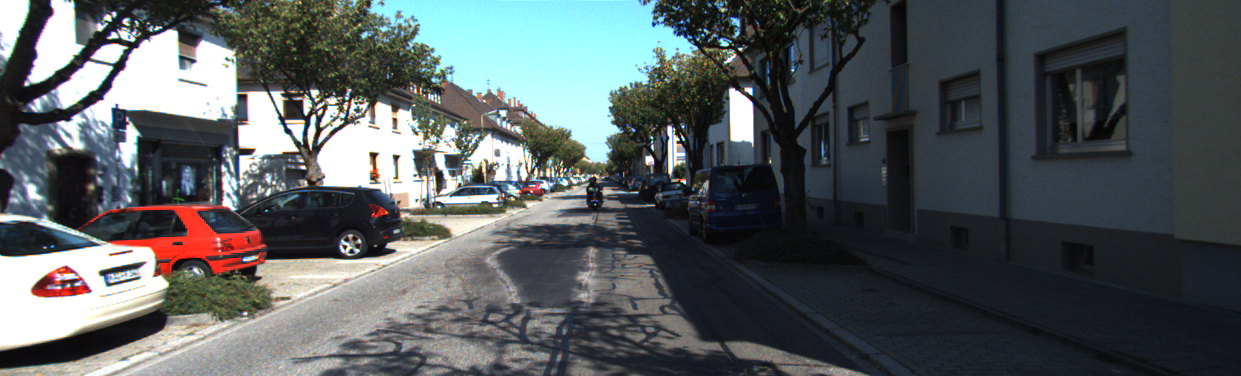} &
\includegraphics[width=0.22\linewidth]{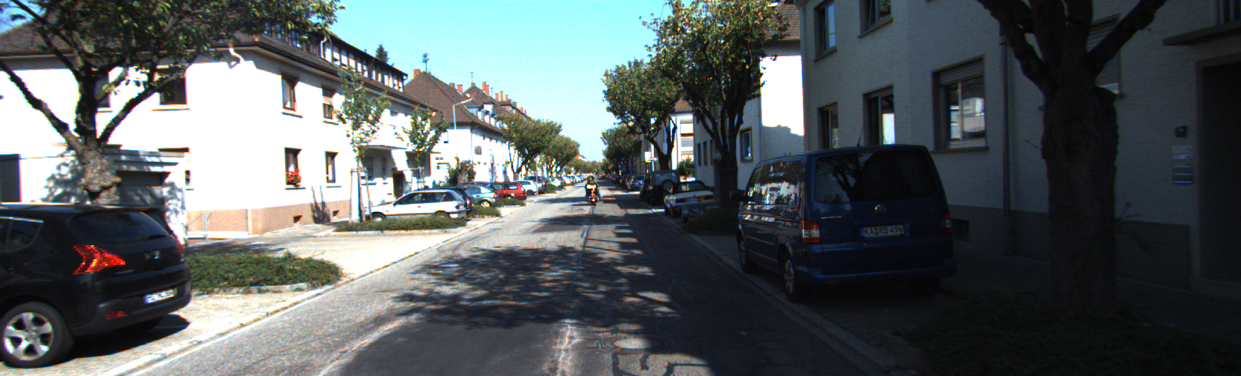} &
\includegraphics[width=0.22\linewidth]{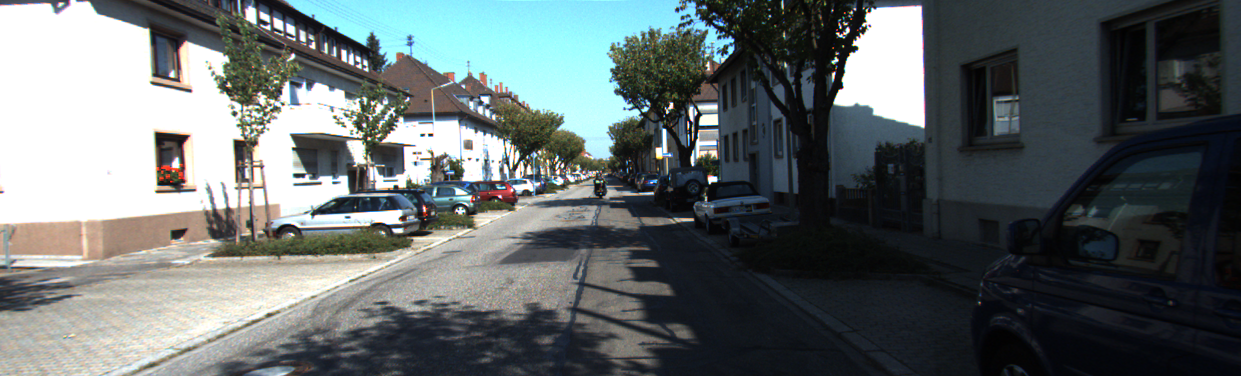} \\[1mm]

\scriptsize Monodepth2 &
\scriptsize &
\scriptsize &
\scriptsize \\
\includegraphics[width=0.22\linewidth]{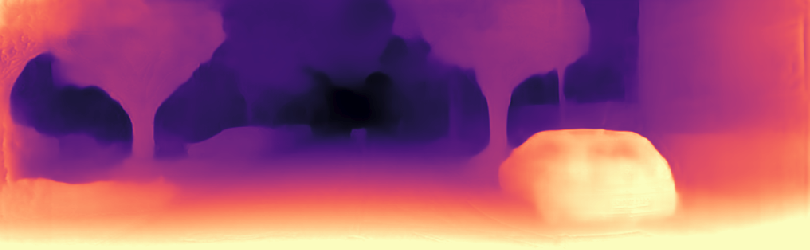} &
\includegraphics[width=0.22\linewidth]{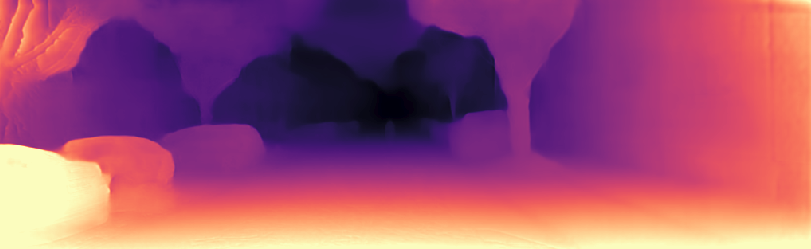} &
\includegraphics[width=0.22\linewidth]{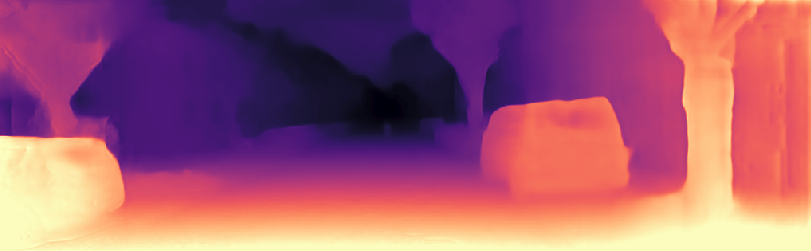} &
\includegraphics[width=0.22\linewidth]{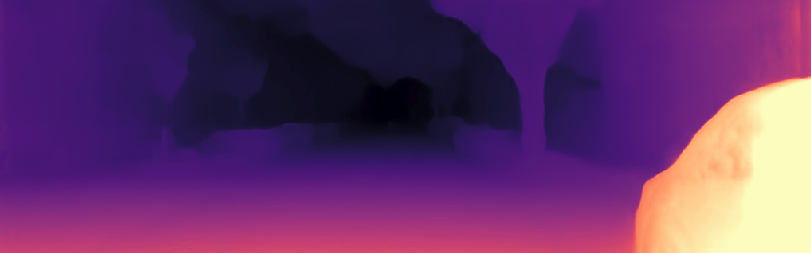} \\[1mm]

\scriptsize FeatDepth &
\scriptsize &
\scriptsize &
\scriptsize \\
\includegraphics[width=0.22\linewidth]{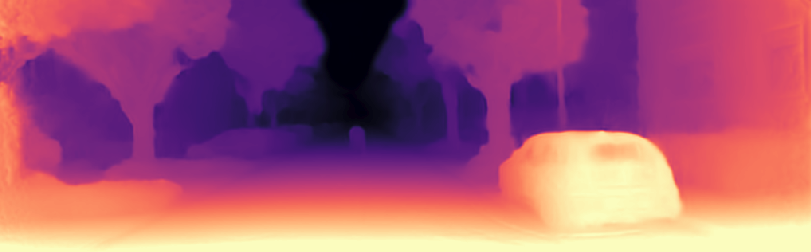} &
\includegraphics[width=0.22\linewidth]{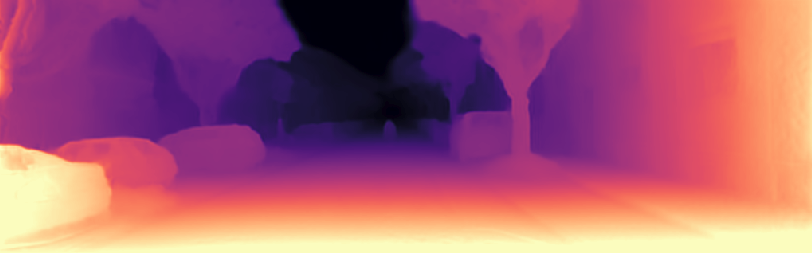} &
\includegraphics[width=0.22\linewidth]{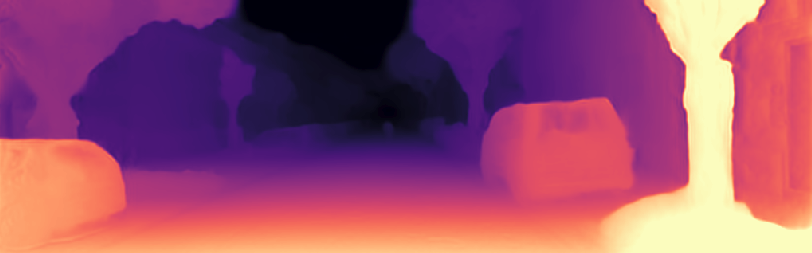} &
\includegraphics[width=0.22\linewidth]{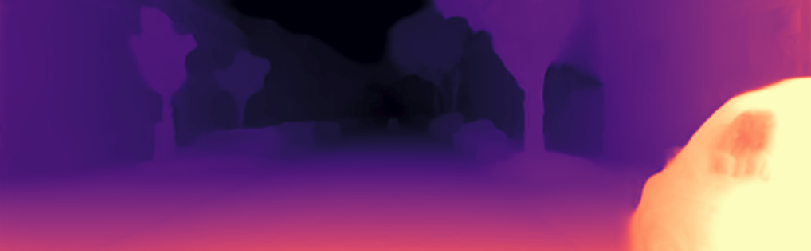} \\[1mm]

\scriptsize SS3D &
\scriptsize &
\scriptsize &
\scriptsize \\
\includegraphics[width=0.22\linewidth]{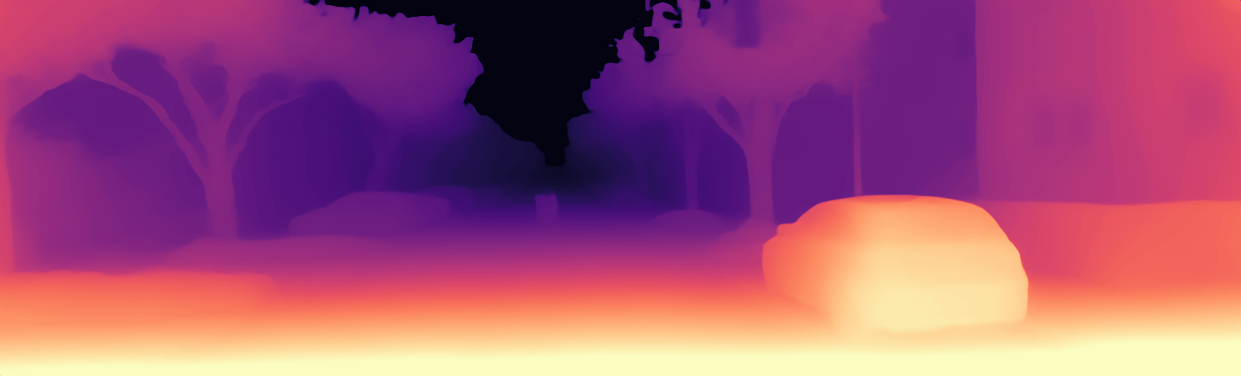} &
\includegraphics[width=0.22\linewidth]{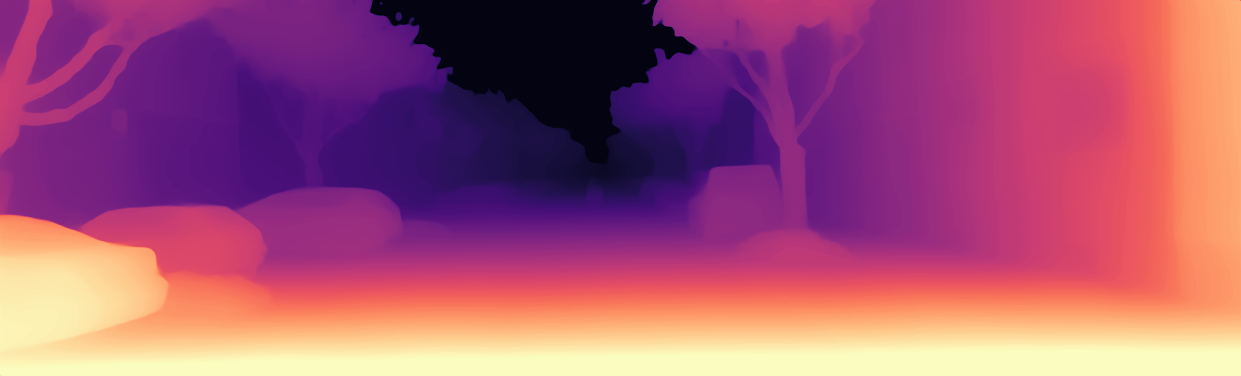} &
\includegraphics[width=0.22\linewidth]{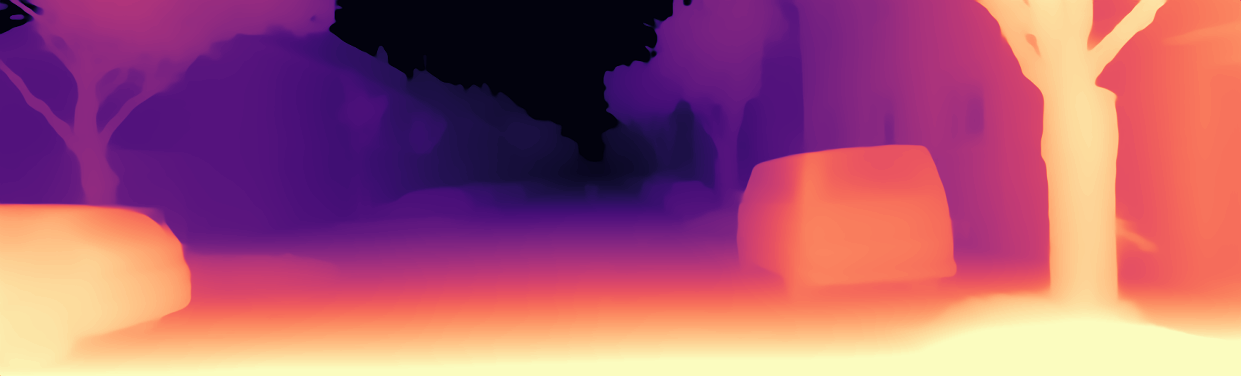} &
\includegraphics[width=0.22\linewidth]{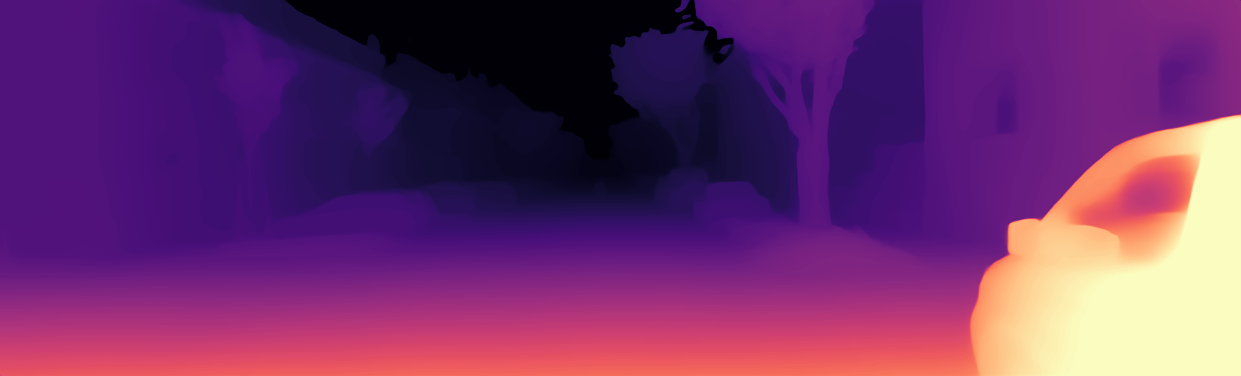} \\[1mm]

\end{tabular}

\vspace{2mm}

\begin{tabular}{cc}

\includegraphics[width=0.48\linewidth]{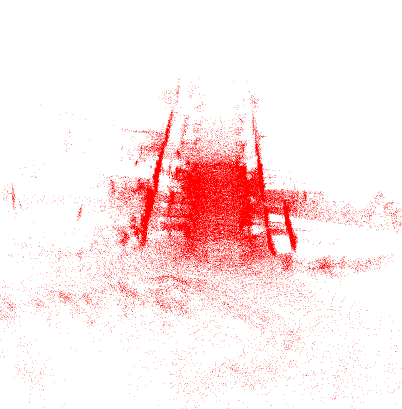} &
\includegraphics[width=0.48\linewidth]{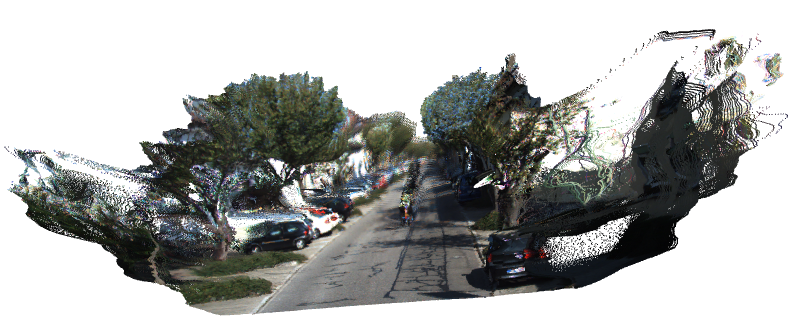} \\[1mm]
\includegraphics[width=0.48\linewidth]{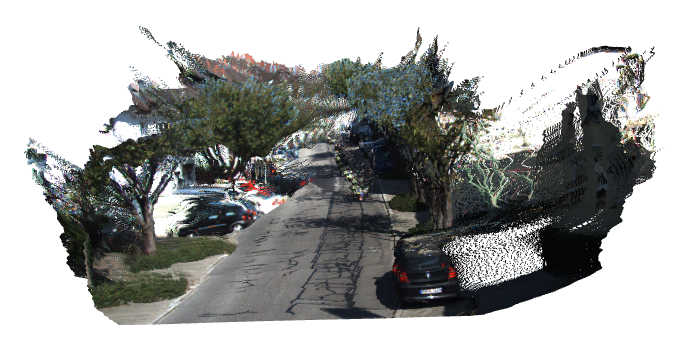} &
\includegraphics[width=0.48\linewidth]{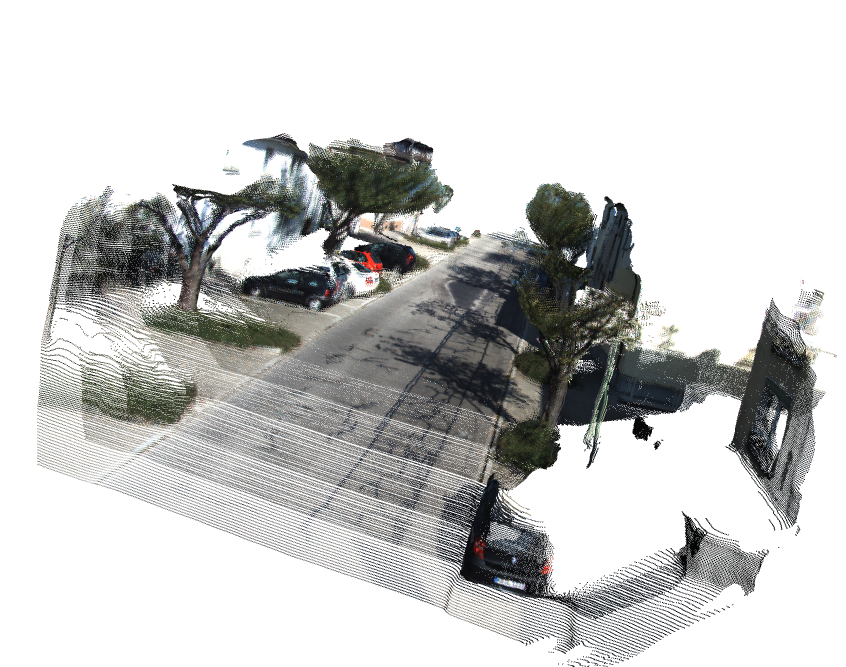}
\end{tabular}

\caption{\textbf{Sequence 2}. Qualitative comparison to strong self-supervised baselines FeatDepth~\cite{shu2020feature} and Monodepth2~\cite{godard2019digging}. \textbf{Top}: predicted depth maps. \textbf{Bottom}: 3D point clouds induced by the predicted depth, pose, and intrinsics. In the bottom block, the LiDAR point cloud is shown in red (top-left), with Monodepth2 (top-right), FeatDepth (bottom-left), and SS3D (bottom-right).}
\label{fig:qualitative_seq2}
\end{figure}

\begin{figure}[H]
\centering
\setlength{\tabcolsep}{2pt}
\renewcommand{\arraystretch}{1.0}

\begin{tabular}{c c c c}
\scriptsize RGB &
\scriptsize &
\scriptsize &
\scriptsize \\
\includegraphics[width=0.22\linewidth]{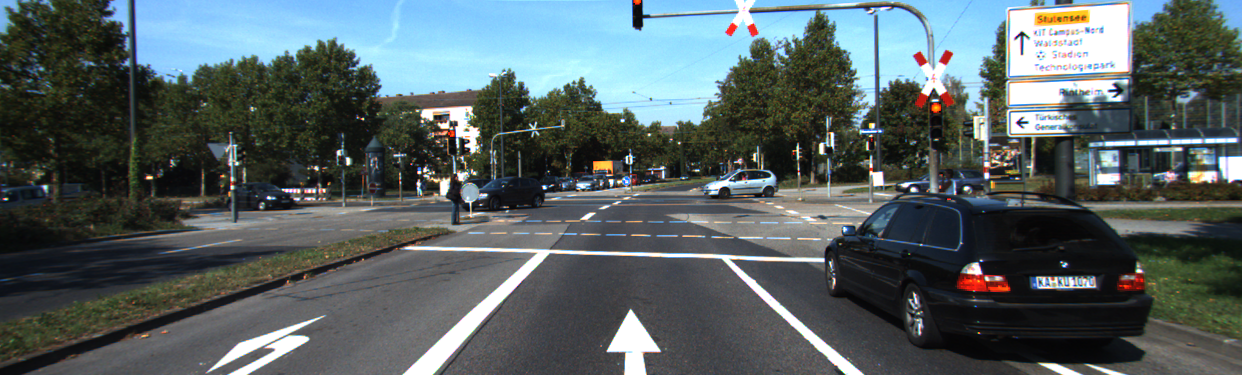} &
\includegraphics[width=0.22\linewidth]{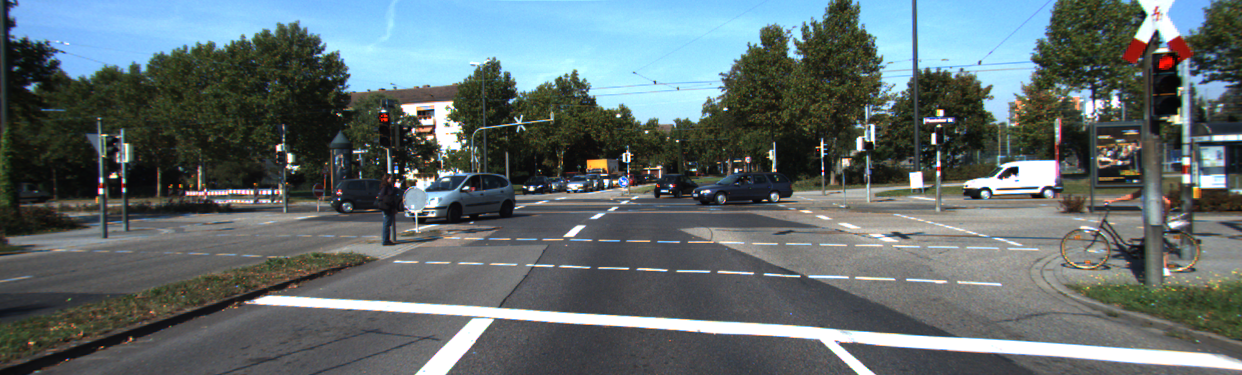} &
\includegraphics[width=0.22\linewidth]{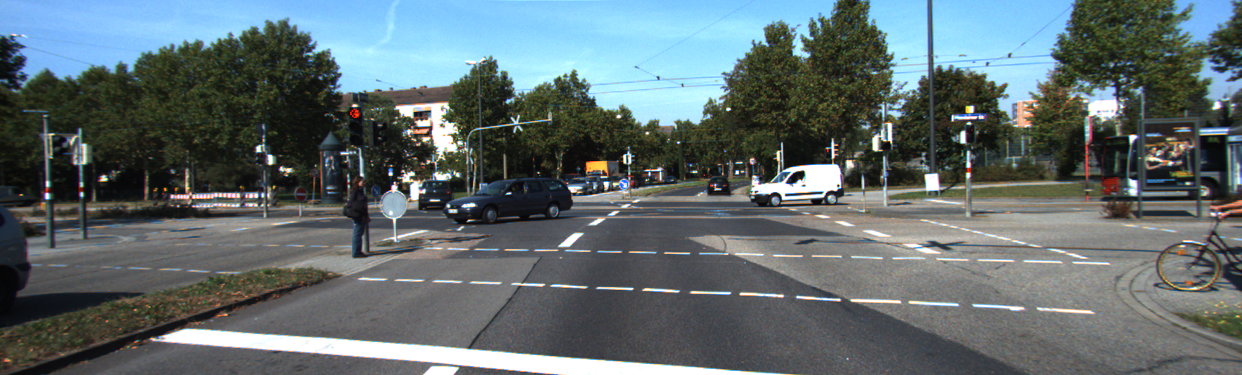} &
\includegraphics[width=0.22\linewidth]{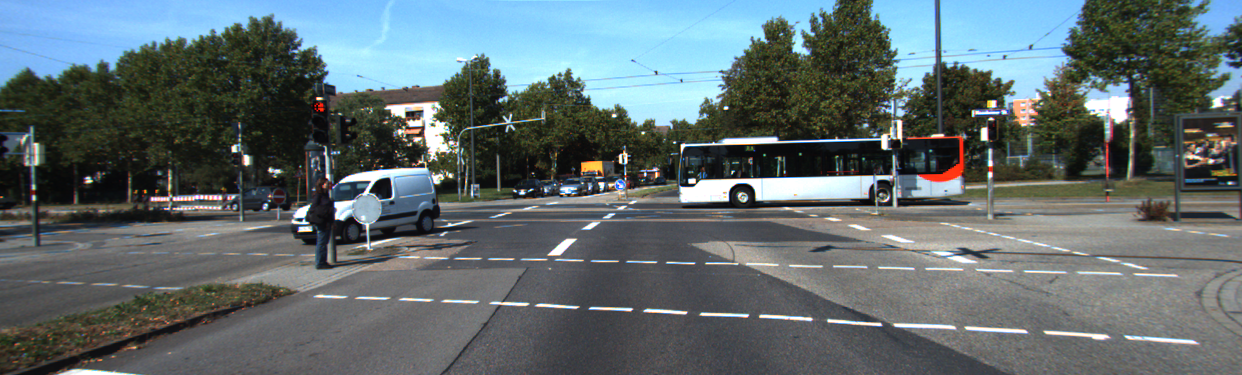} \\[1mm]

\scriptsize Monodepth2 &
\scriptsize &
\scriptsize &
\scriptsize \\
\includegraphics[width=0.22\linewidth]{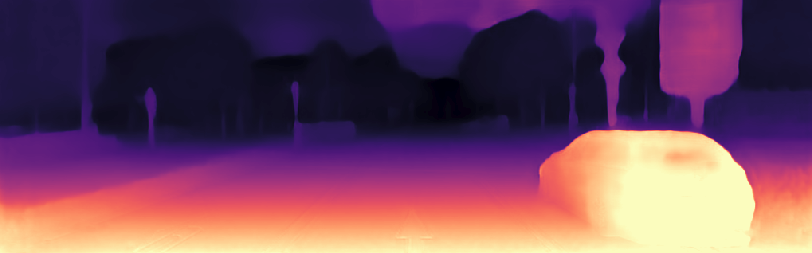} &
\includegraphics[width=0.22\linewidth]{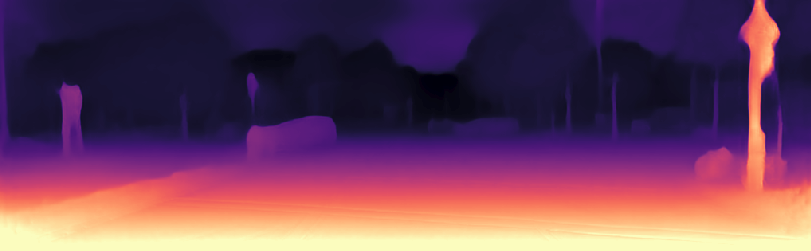} &
\includegraphics[width=0.22\linewidth]{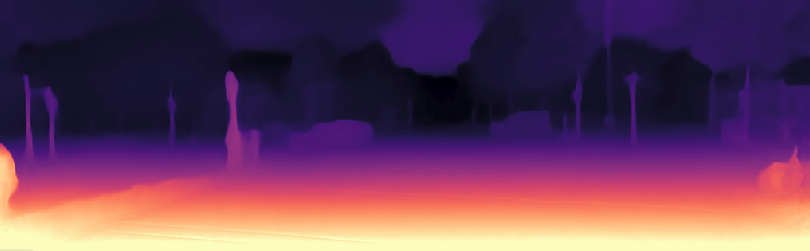} &
\includegraphics[width=0.22\linewidth]{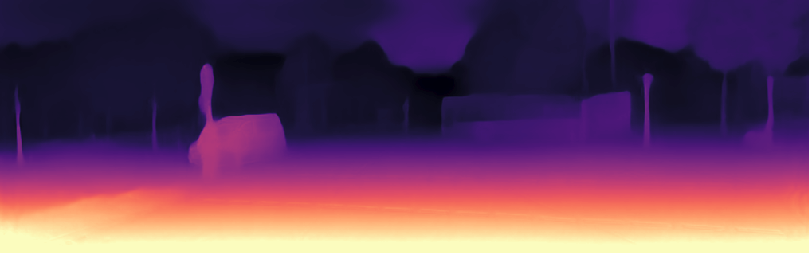} \\[1mm]

\scriptsize FeatDepth &
\scriptsize &
\scriptsize &
\scriptsize \\
\includegraphics[width=0.22\linewidth]{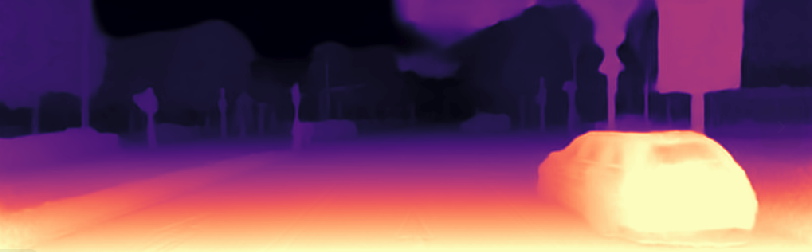} &
\includegraphics[width=0.22\linewidth]{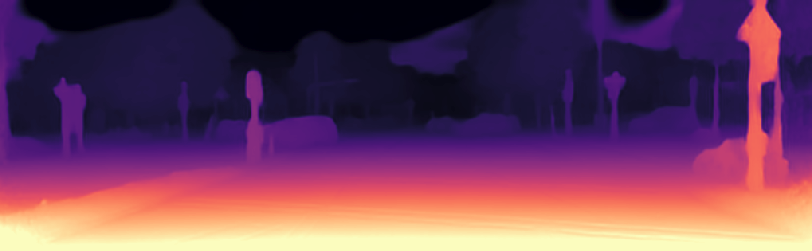} &
\includegraphics[width=0.22\linewidth]{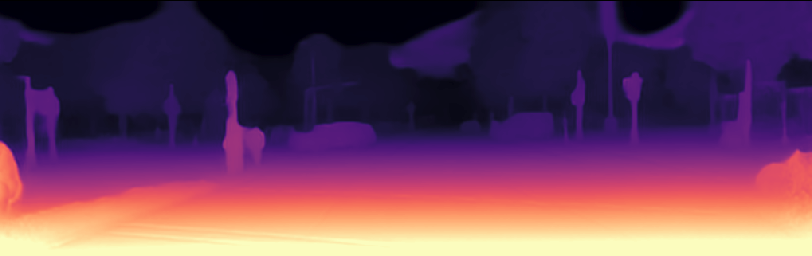} &
\includegraphics[width=0.22\linewidth]{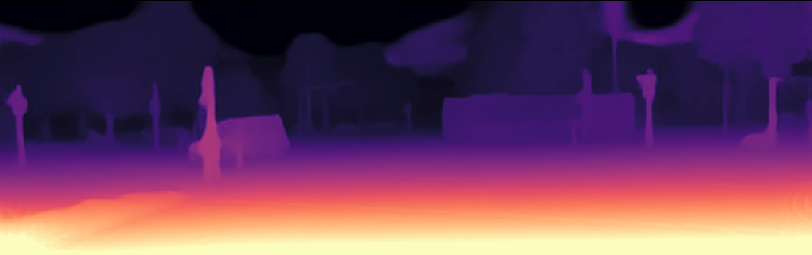} \\[1mm]

\scriptsize SS3D &
\scriptsize &
\scriptsize &
\scriptsize \\
\includegraphics[width=0.22\linewidth]{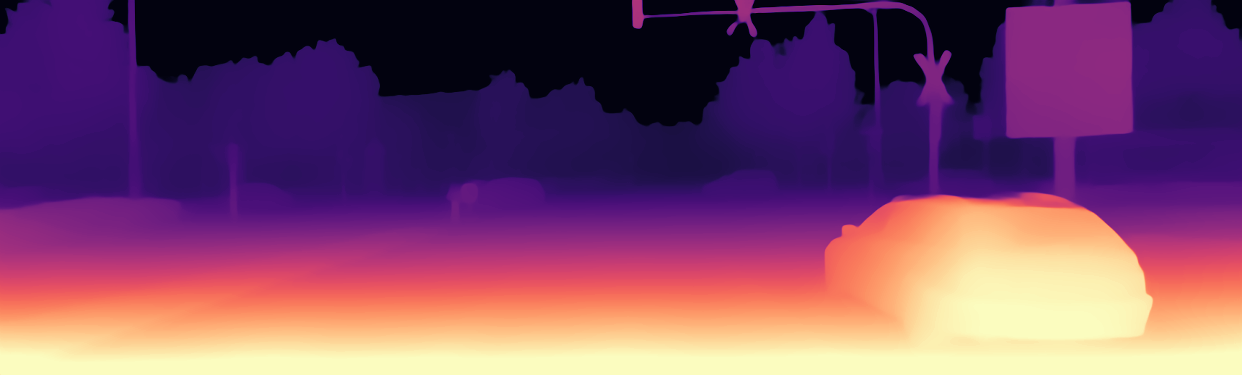} &
\includegraphics[width=0.22\linewidth]{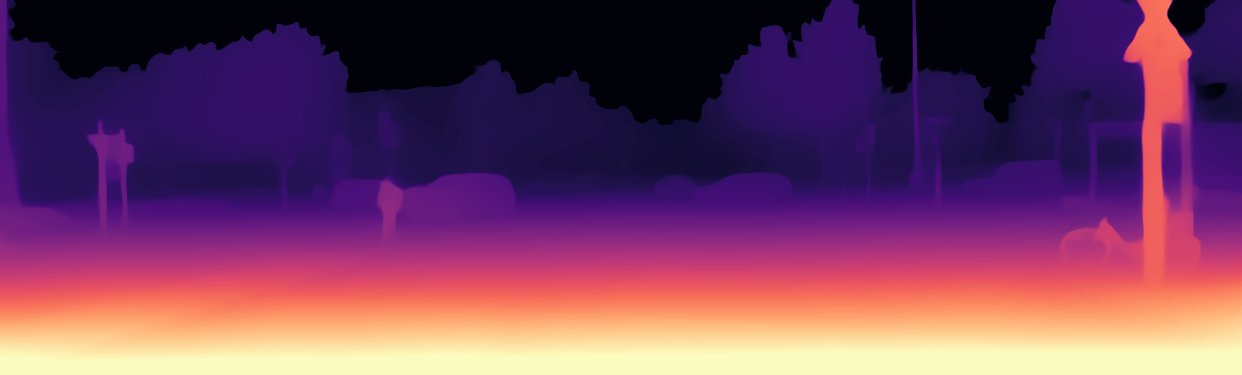} &
\includegraphics[width=0.22\linewidth]{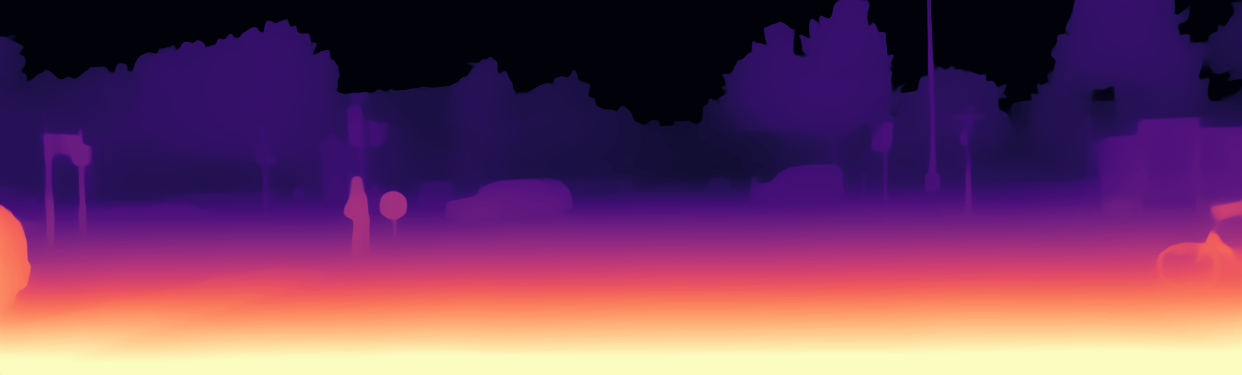} &
\includegraphics[width=0.22\linewidth]{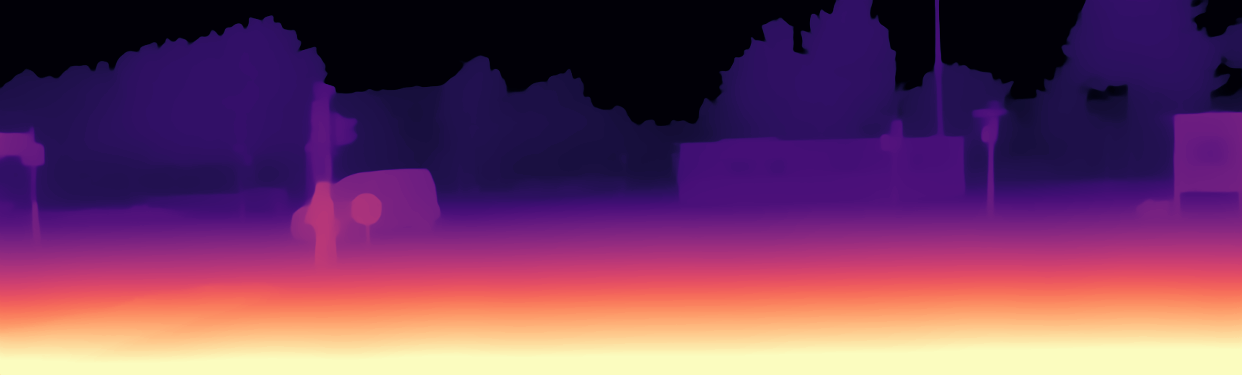} \\[1mm]

\end{tabular}

\vspace{2mm}

\begin{tabular}{cc}

\includegraphics[width=0.48\linewidth]{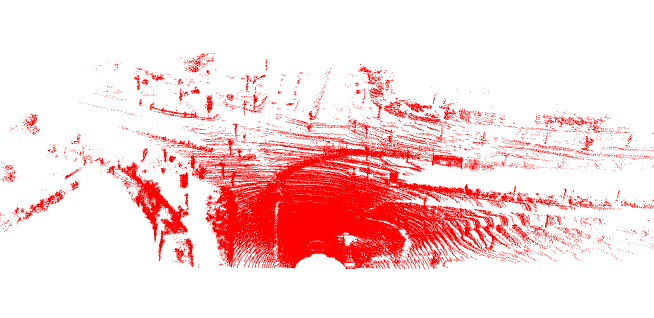} &
\includegraphics[width=0.48\linewidth]{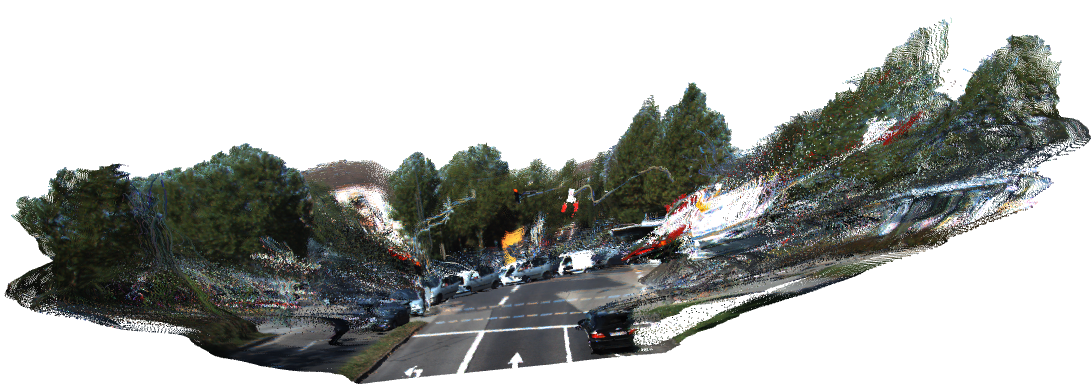} \\[1mm]
\includegraphics[width=0.48\linewidth]{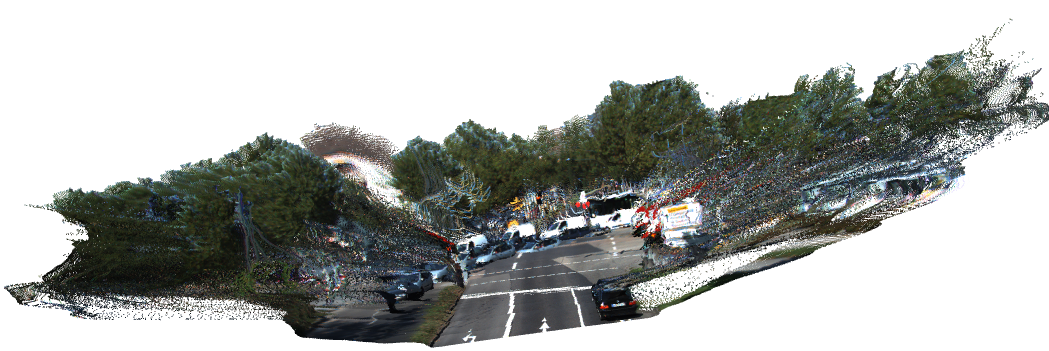} &
\includegraphics[width=0.48\linewidth]{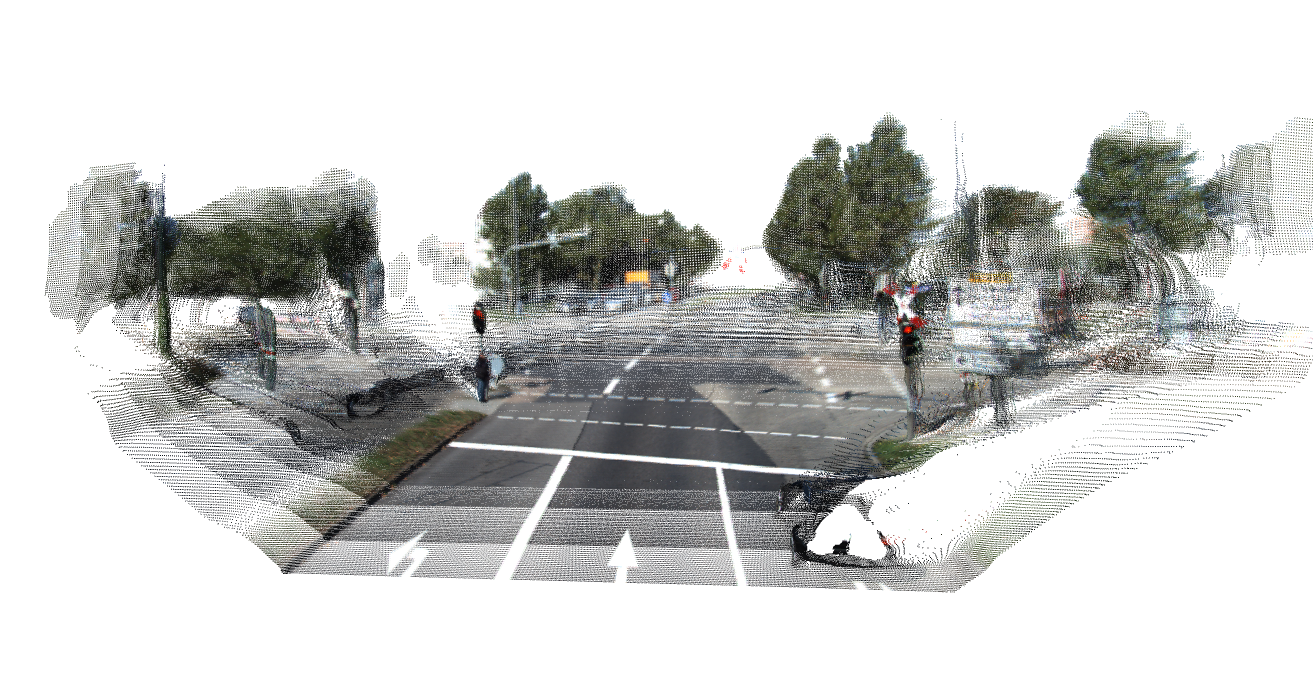}
\end{tabular}

\caption{\textbf{Sequence 3}. Qualitative comparison to strong self-supervised baselines FeatDepth~\cite{shu2020feature} and Monodepth2~\cite{godard2019digging}. \textbf{Top}: predicted depth maps. \textbf{Bottom}: 3D point clouds induced by the predicted depth, pose, and intrinsics. In the bottom block, the LiDAR point cloud is shown in red (top-left), with Monodepth2 (top-right), FeatDepth (bottom-left), and SS3D (bottom-right).}
\label{fig:qualitative_seq3}
\end{figure}

\begin{figure}[H]
\centering
\setlength{\tabcolsep}{2pt}
\renewcommand{\arraystretch}{1.0}

\begin{tabular}{c c c c}
\scriptsize RGB &
\scriptsize &
\scriptsize &
\scriptsize \\
\includegraphics[width=0.22\linewidth]{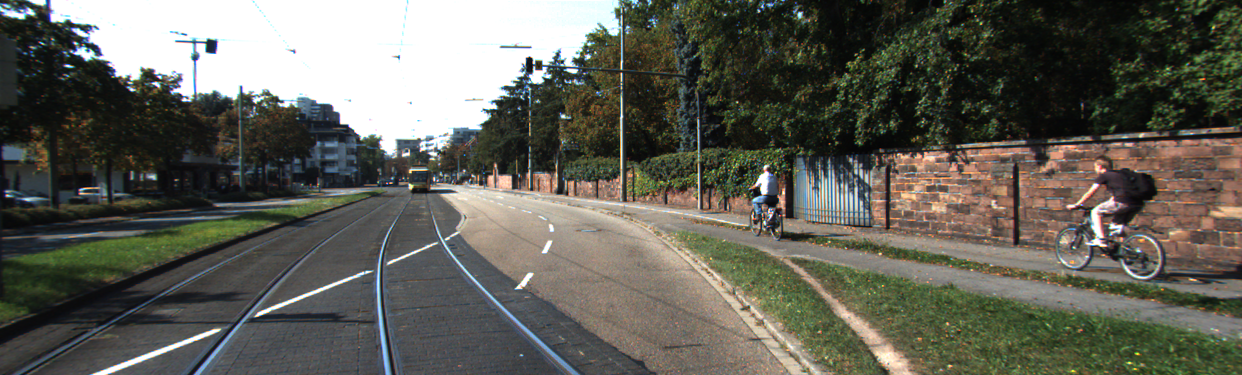} &
\includegraphics[width=0.22\linewidth]{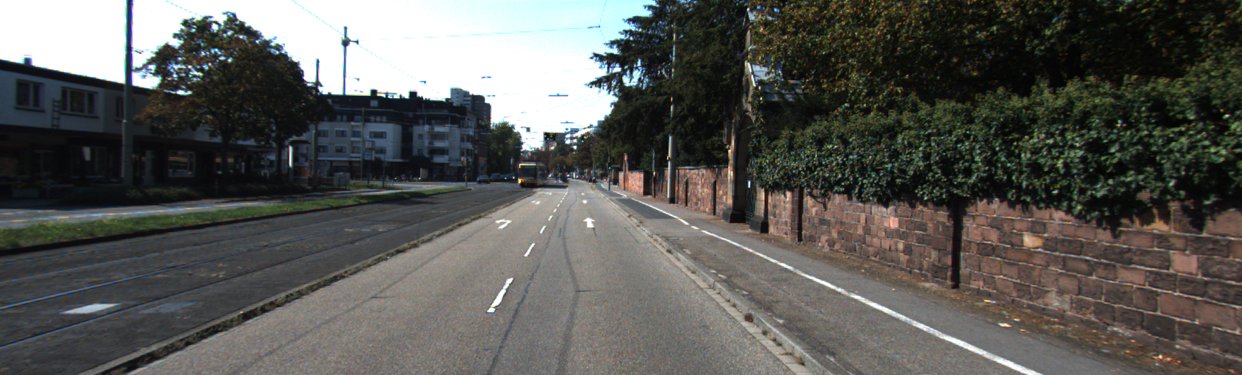} &
\includegraphics[width=0.22\linewidth]{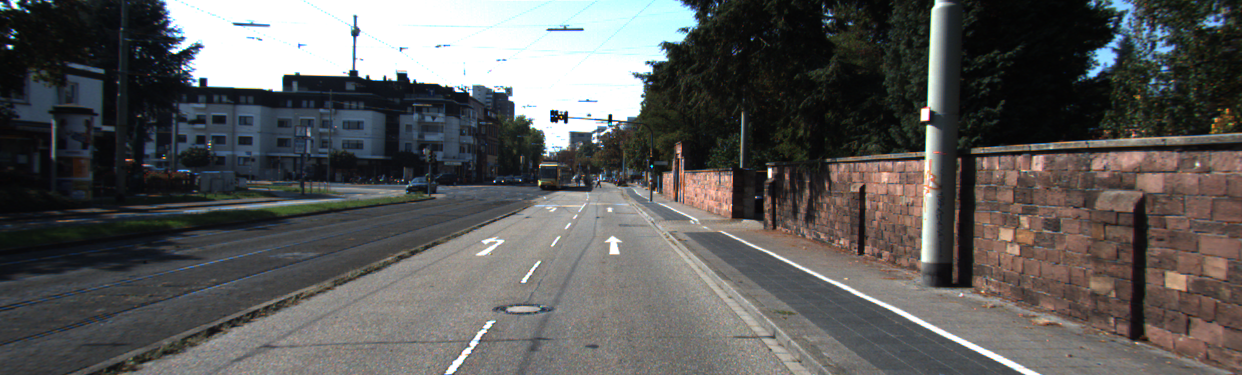} &
\includegraphics[width=0.22\linewidth]{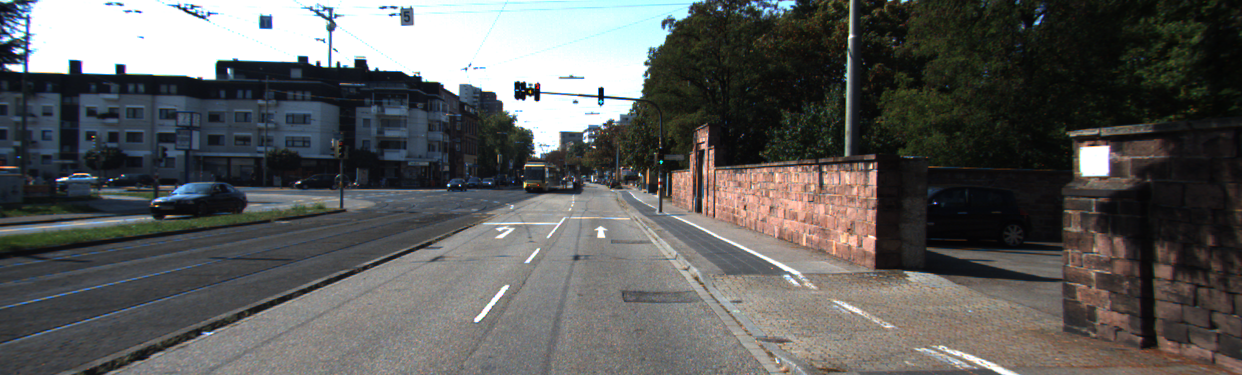} \\[1mm]

\scriptsize Monodepth2 &
\scriptsize &
\scriptsize &
\scriptsize \\
\includegraphics[width=0.22\linewidth]{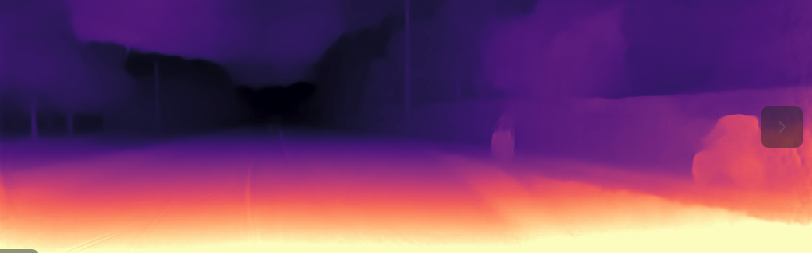} &
\includegraphics[width=0.22\linewidth]{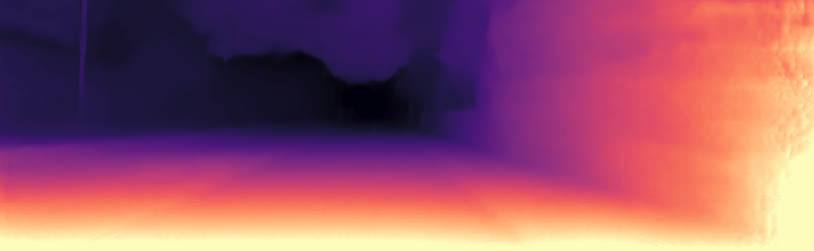} &
\includegraphics[width=0.22\linewidth]{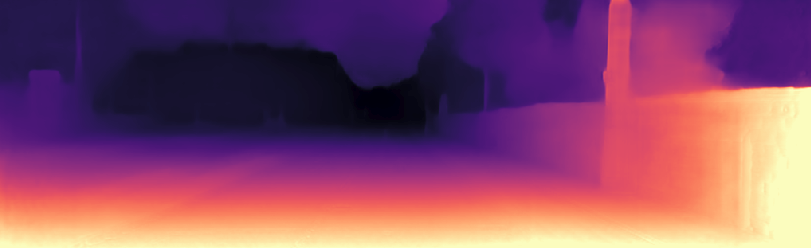} &
\includegraphics[width=0.22\linewidth]{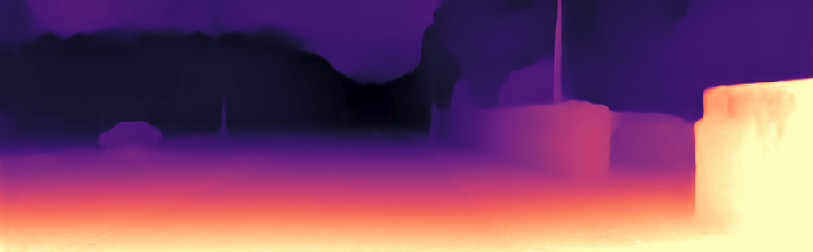} \\[1mm]

\scriptsize FeatDepth &
\scriptsize &
\scriptsize &
\scriptsize \\
\includegraphics[width=0.22\linewidth]{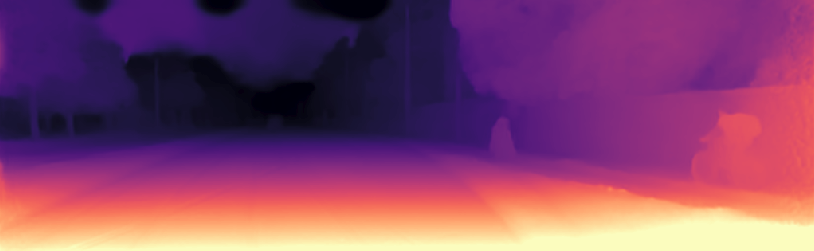} &
\includegraphics[width=0.22\linewidth]{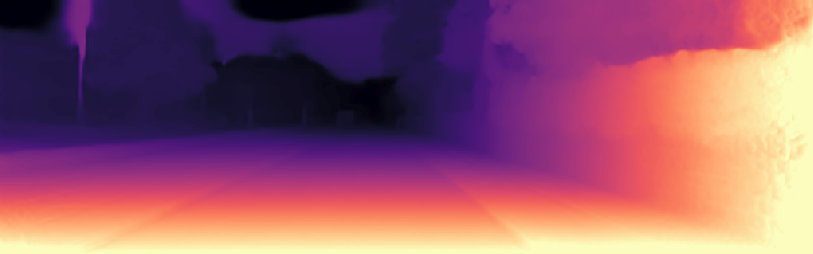} &
\includegraphics[width=0.22\linewidth]{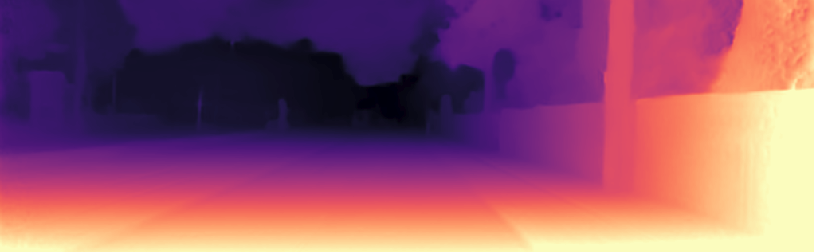} &
\includegraphics[width=0.22\linewidth]{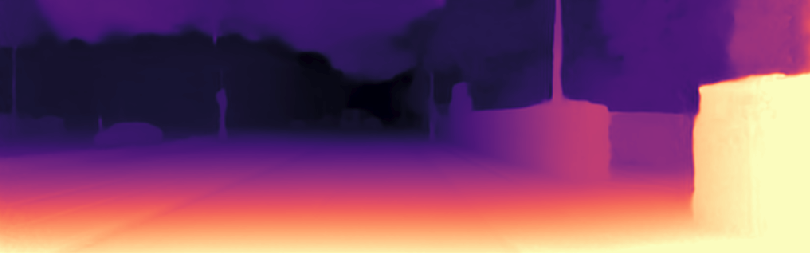} \\[1mm]

\scriptsize SS3D &
\scriptsize &
\scriptsize &
\scriptsize \\
\includegraphics[width=0.22\linewidth]{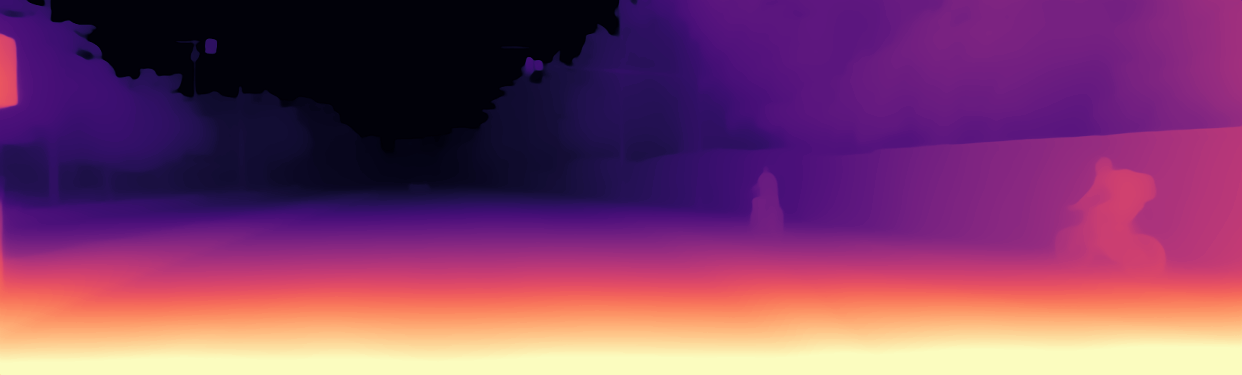} &
\includegraphics[width=0.22\linewidth]{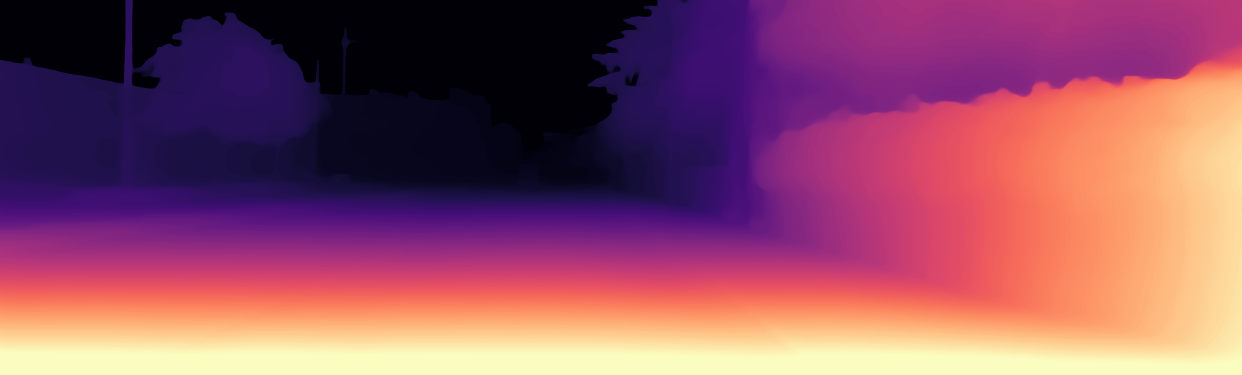} &
\includegraphics[width=0.22\linewidth]{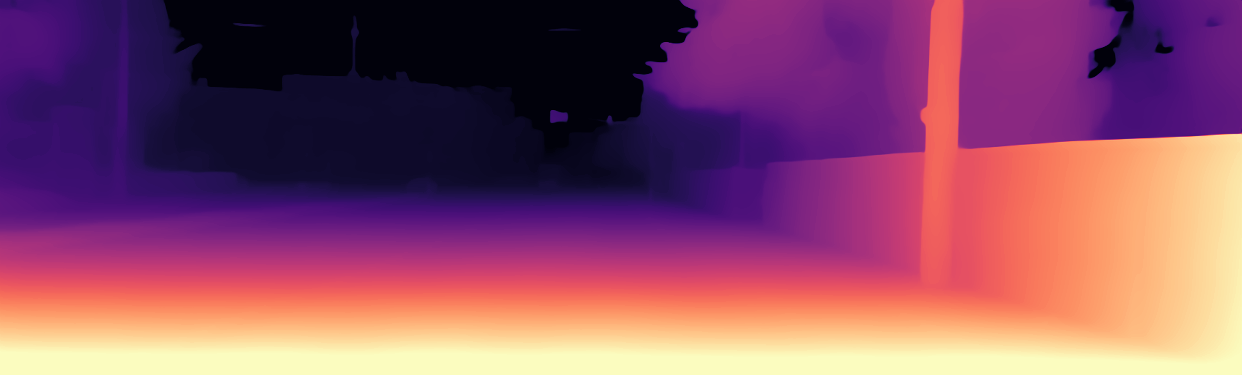} &
\includegraphics[width=0.22\linewidth]{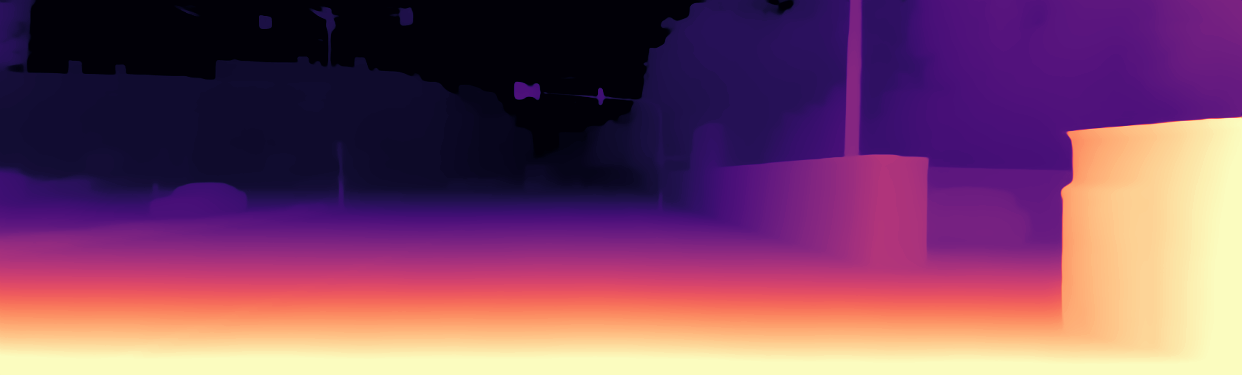} \\[1mm]

\end{tabular}

\vspace{2mm}

\begin{tabular}{cc}

\includegraphics[width=0.48\linewidth]{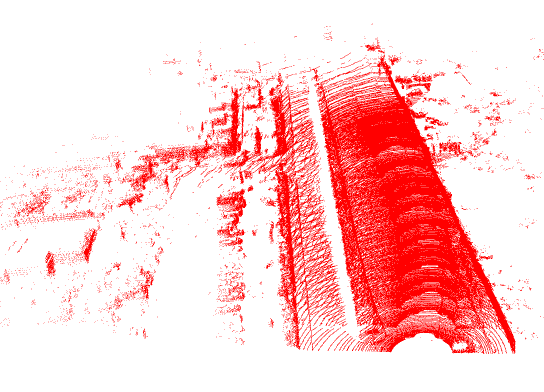} &
\includegraphics[width=0.48\linewidth]{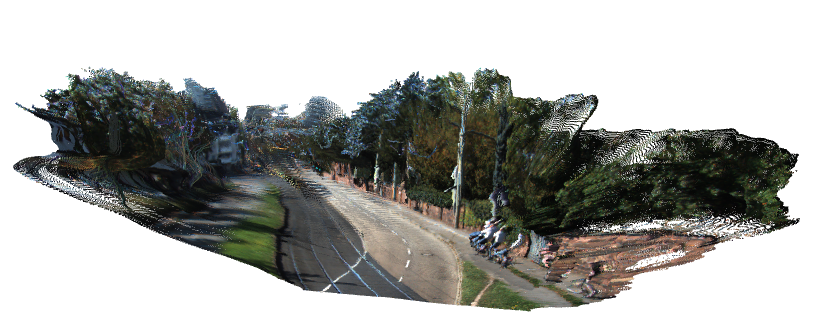} \\[1mm]
\includegraphics[width=0.48\linewidth]{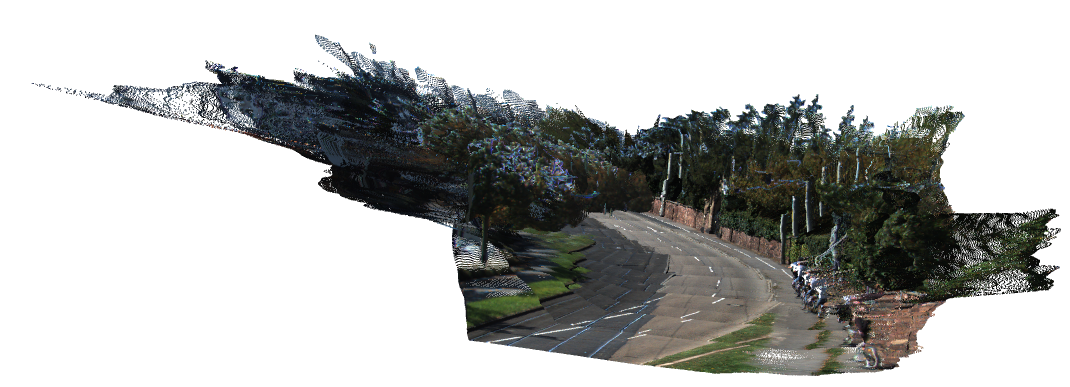} &
\includegraphics[width=0.48\linewidth]{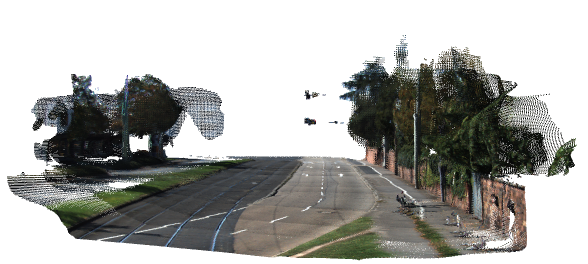}
\end{tabular}

\caption{\textbf{Sequence 4}. Qualitative comparison to strong self-supervised baselines FeatDepth~\cite{shu2020feature} and Monodepth2~\cite{godard2019digging}. \textbf{Top}: predicted depth maps. \textbf{Bottom}: 3D point clouds induced by the predicted depth, pose, and intrinsics. In the bottom block, the LiDAR point cloud is shown in red (top-left), with Monodepth2 (top-right), FeatDepth (bottom-left), and SS3D (bottom-right).}
\label{fig:qualitative_seq4}
\end{figure}

\begin{figure}[H]
\centering
\setlength{\tabcolsep}{1pt}

\begin{tabular}{cccc}
\includegraphics[width=0.235\linewidth]{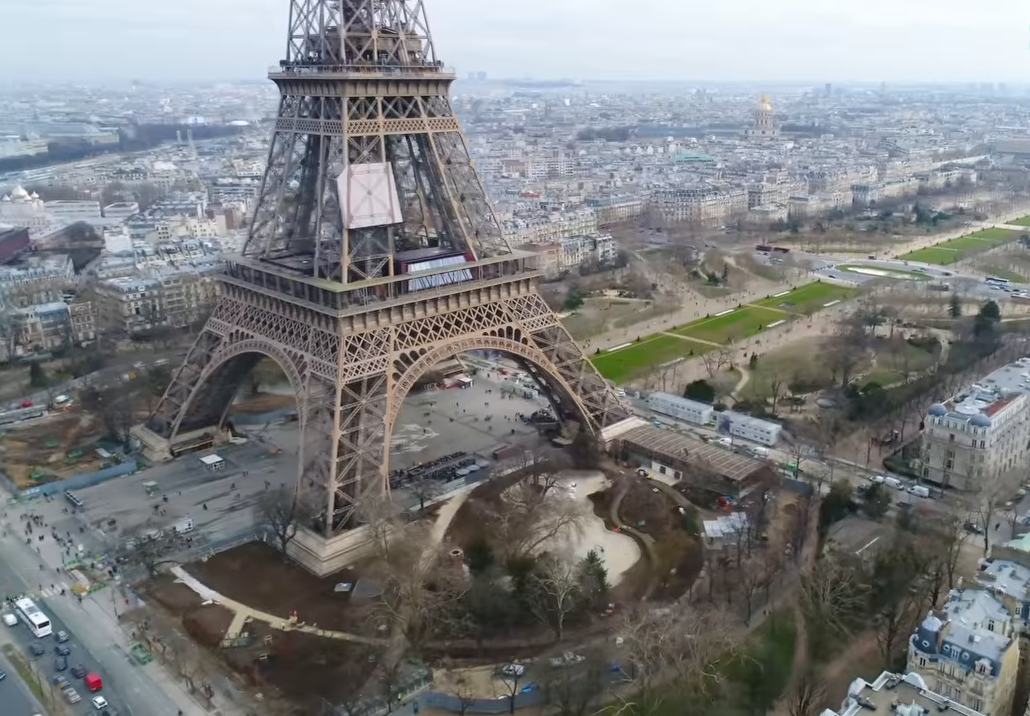} &
\includegraphics[width=0.235\linewidth]{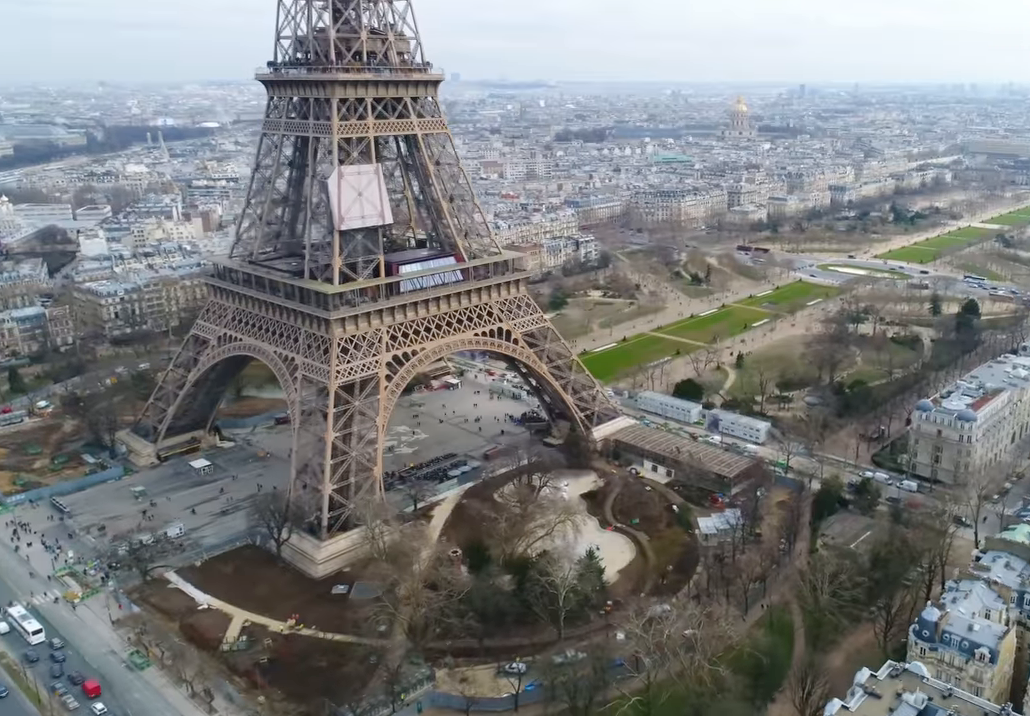} &
\includegraphics[width=0.235\linewidth]{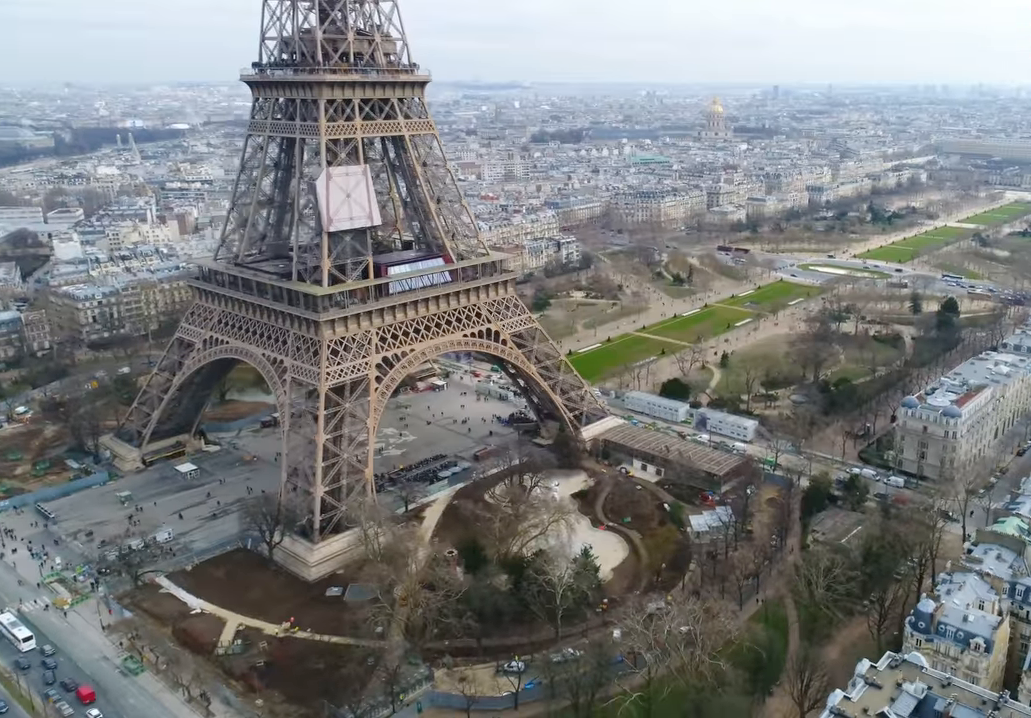} &
\includegraphics[width=0.235\linewidth]{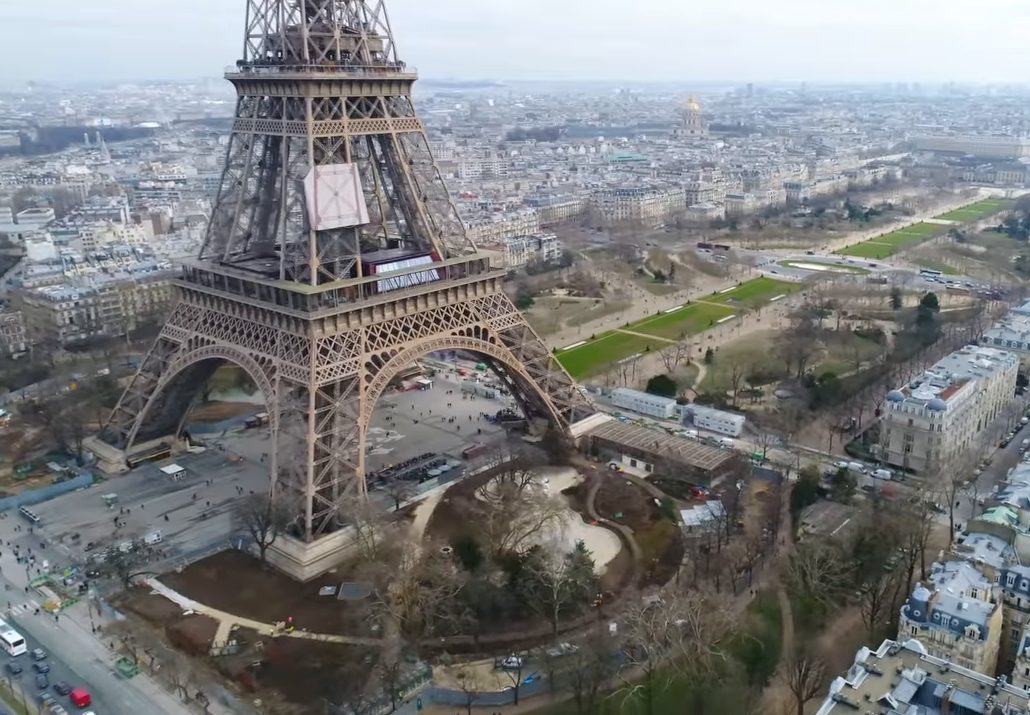} \\[1mm]

\includegraphics[width=0.235\linewidth]{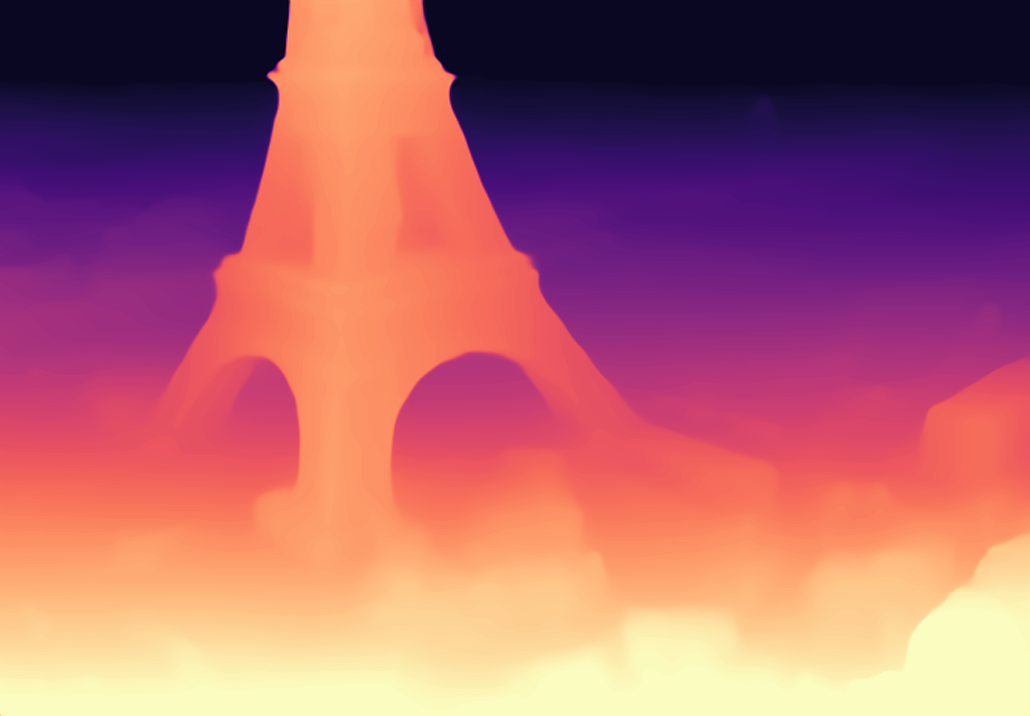} &
\includegraphics[width=0.235\linewidth]{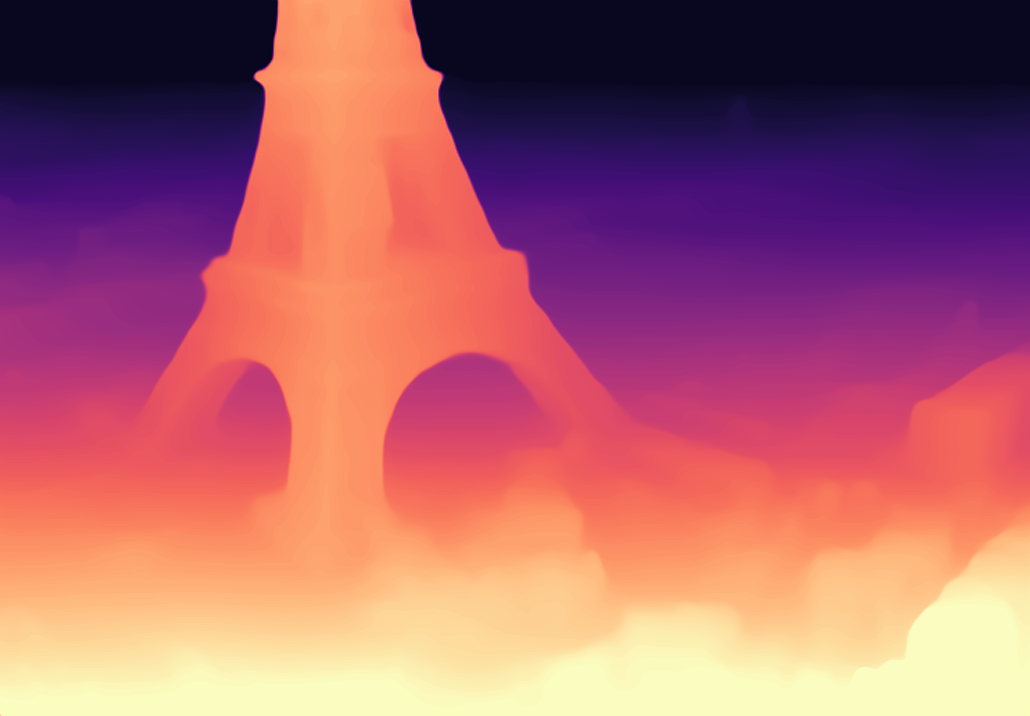} &
\includegraphics[width=0.235\linewidth]{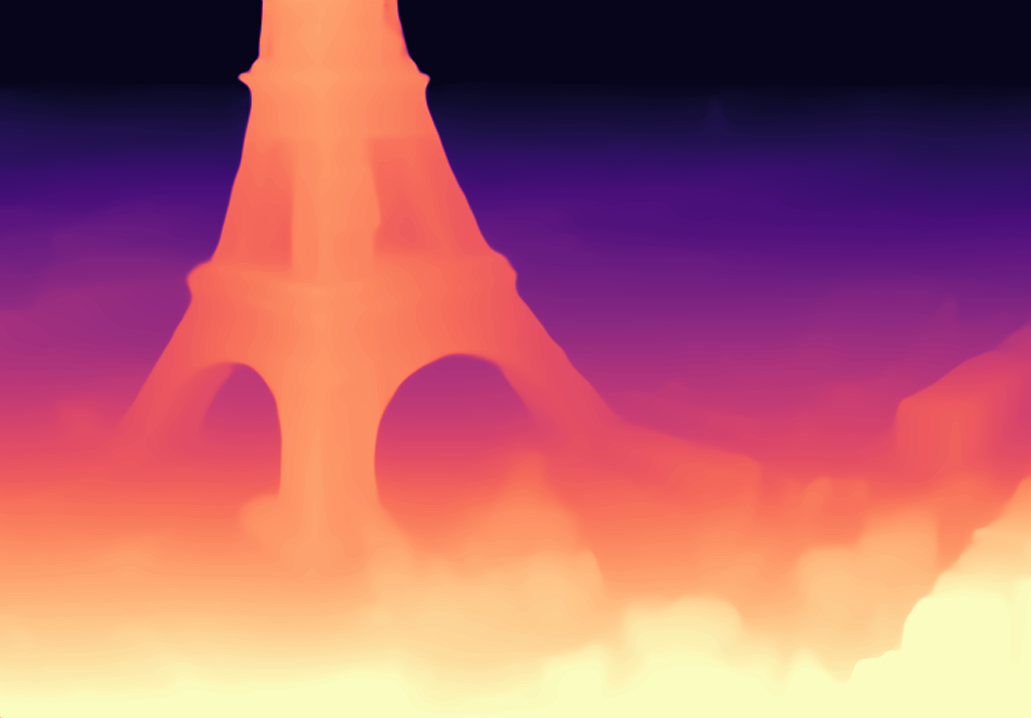} &
\includegraphics[width=0.235\linewidth]{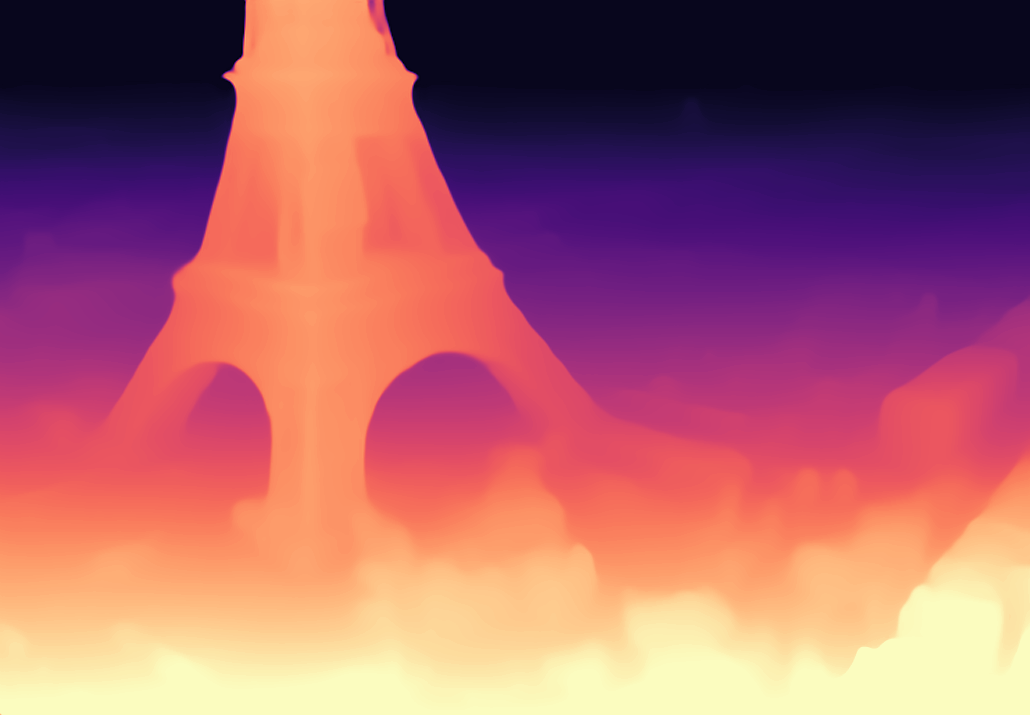}
\end{tabular}

\vspace{2mm}

\includegraphics[width=0.55\linewidth]{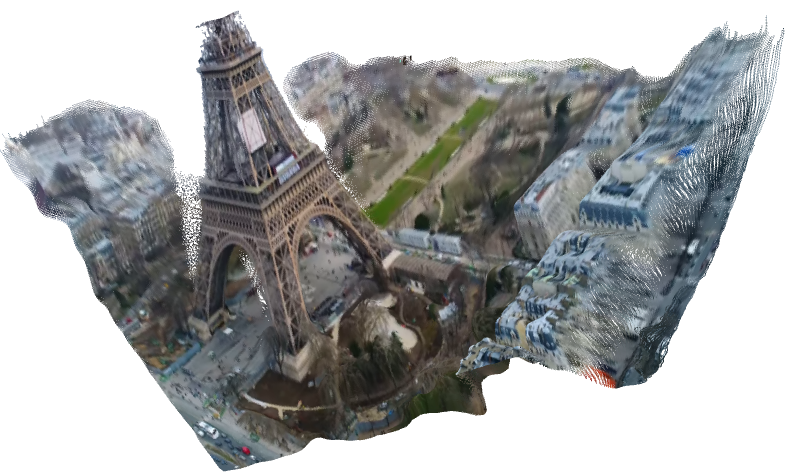}

\caption{We apply SS3D to a casual drone video of the Eiffel Tower.
\textbf{Top:} predicted depth.
\textbf{Bottom:} the induced 3D reconstruction obtained by fusing the predicted depth with the estimated camera poses and intrinsics.
Video source: \url{https://www.youtube.com/watch?v=ibUscnfrfEo}.}
\label{fig:qualitative_casual1}
\end{figure}

\begin{figure}[H]
\centering
\setlength{\tabcolsep}{1pt}

\begin{tabular}{cccc}
\includegraphics[width=0.235\linewidth]{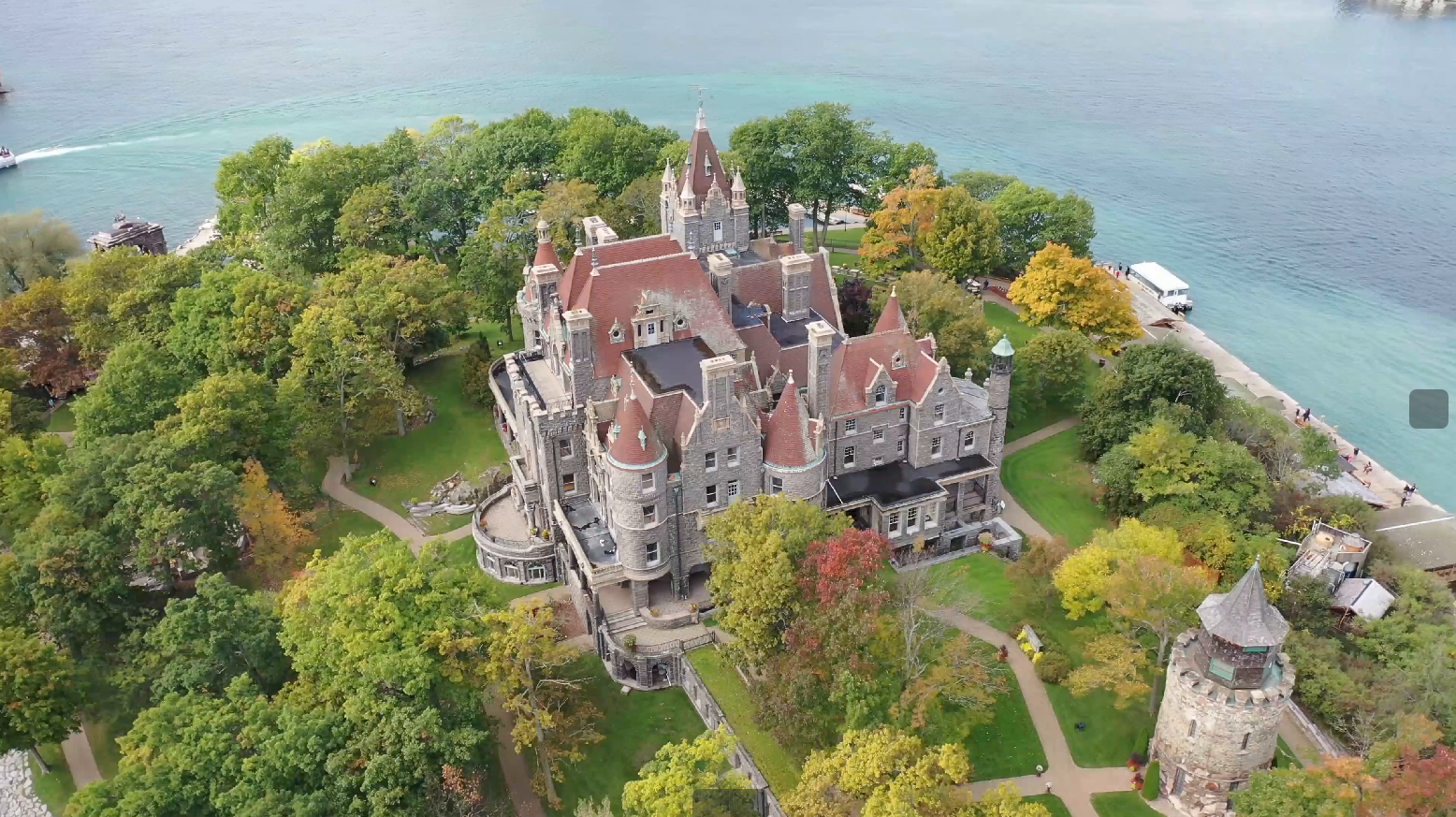} &
\includegraphics[width=0.235\linewidth]{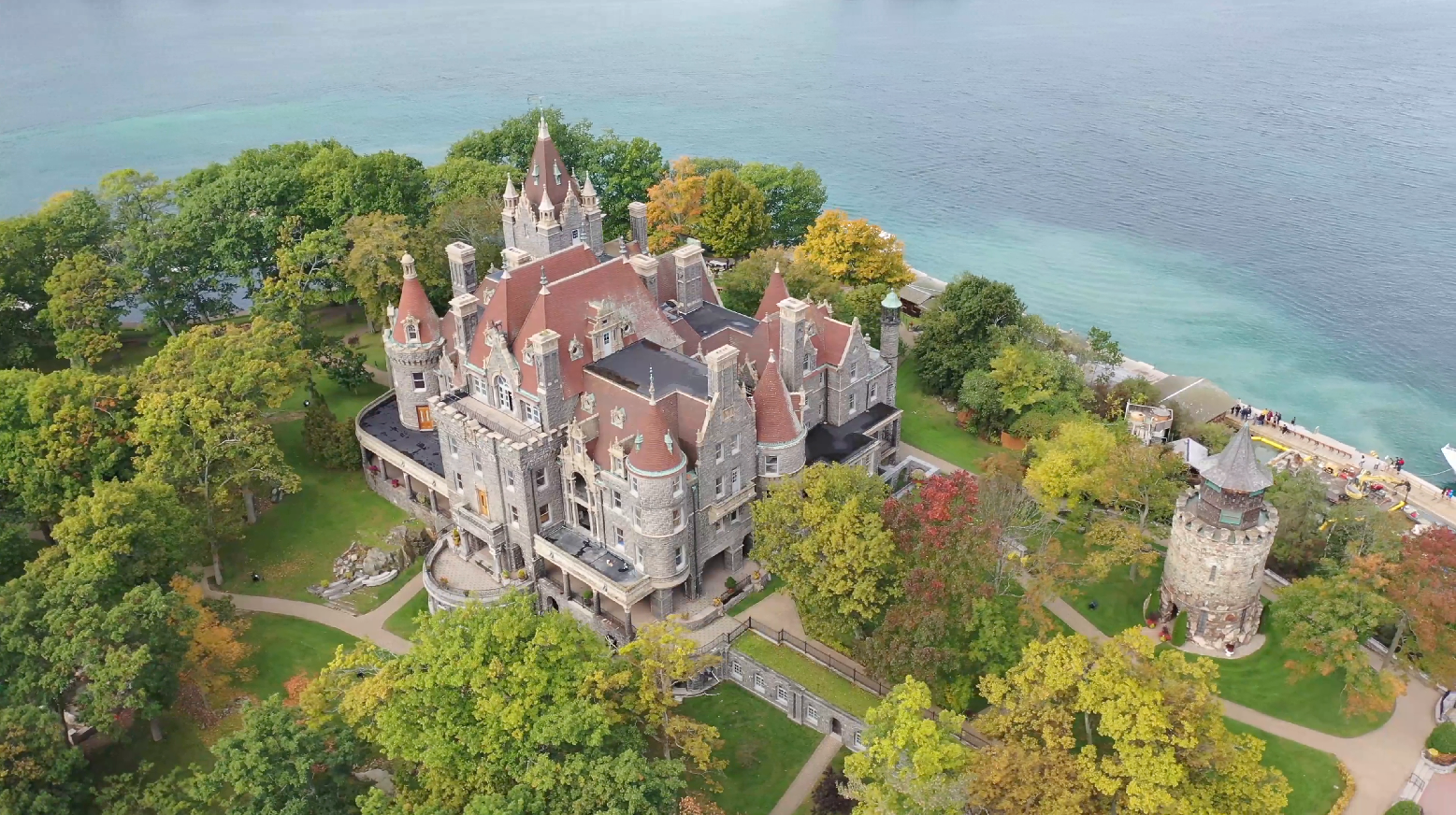} &
\includegraphics[width=0.235\linewidth]{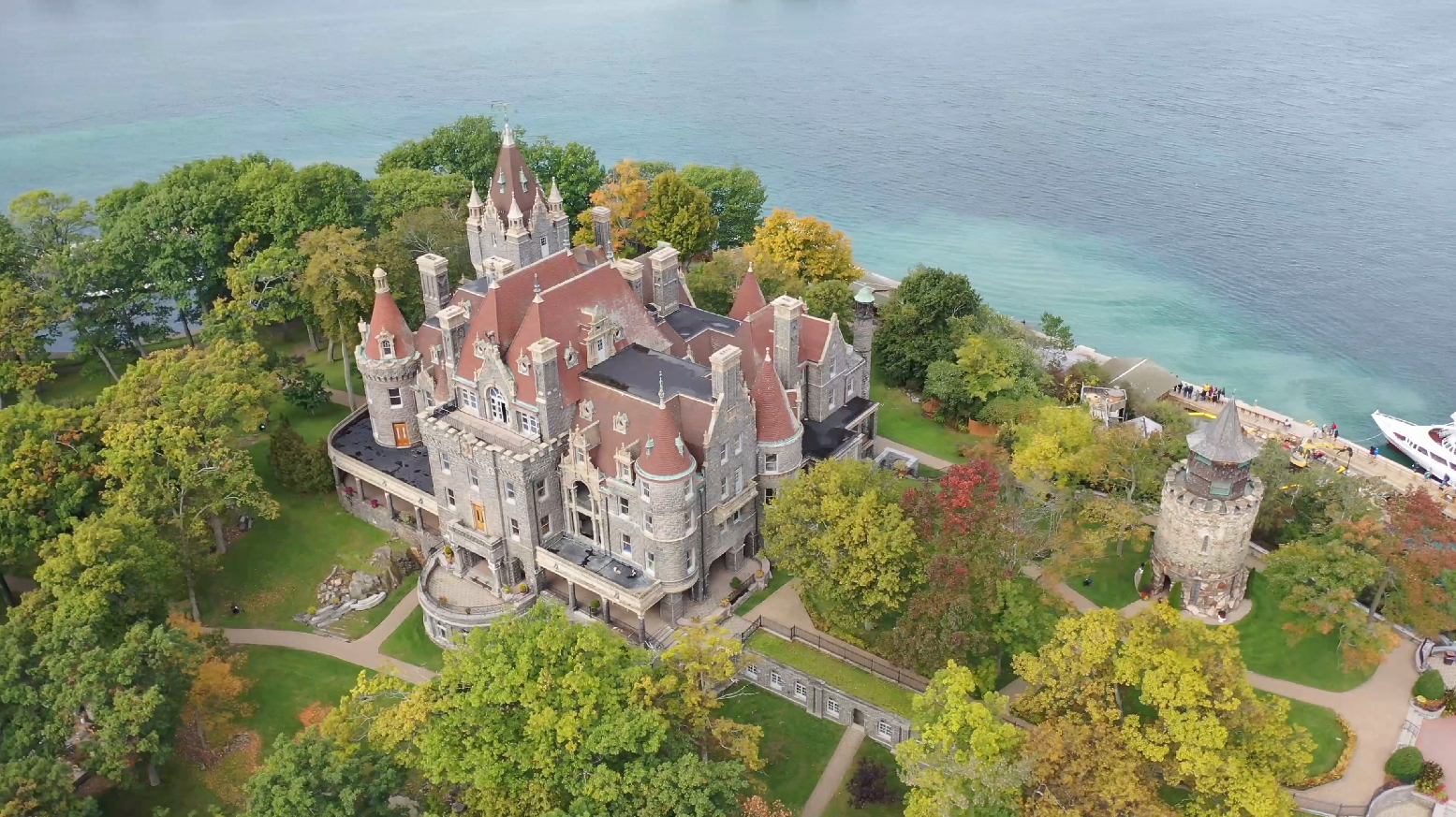} \\[1mm]

\includegraphics[width=0.235\linewidth]{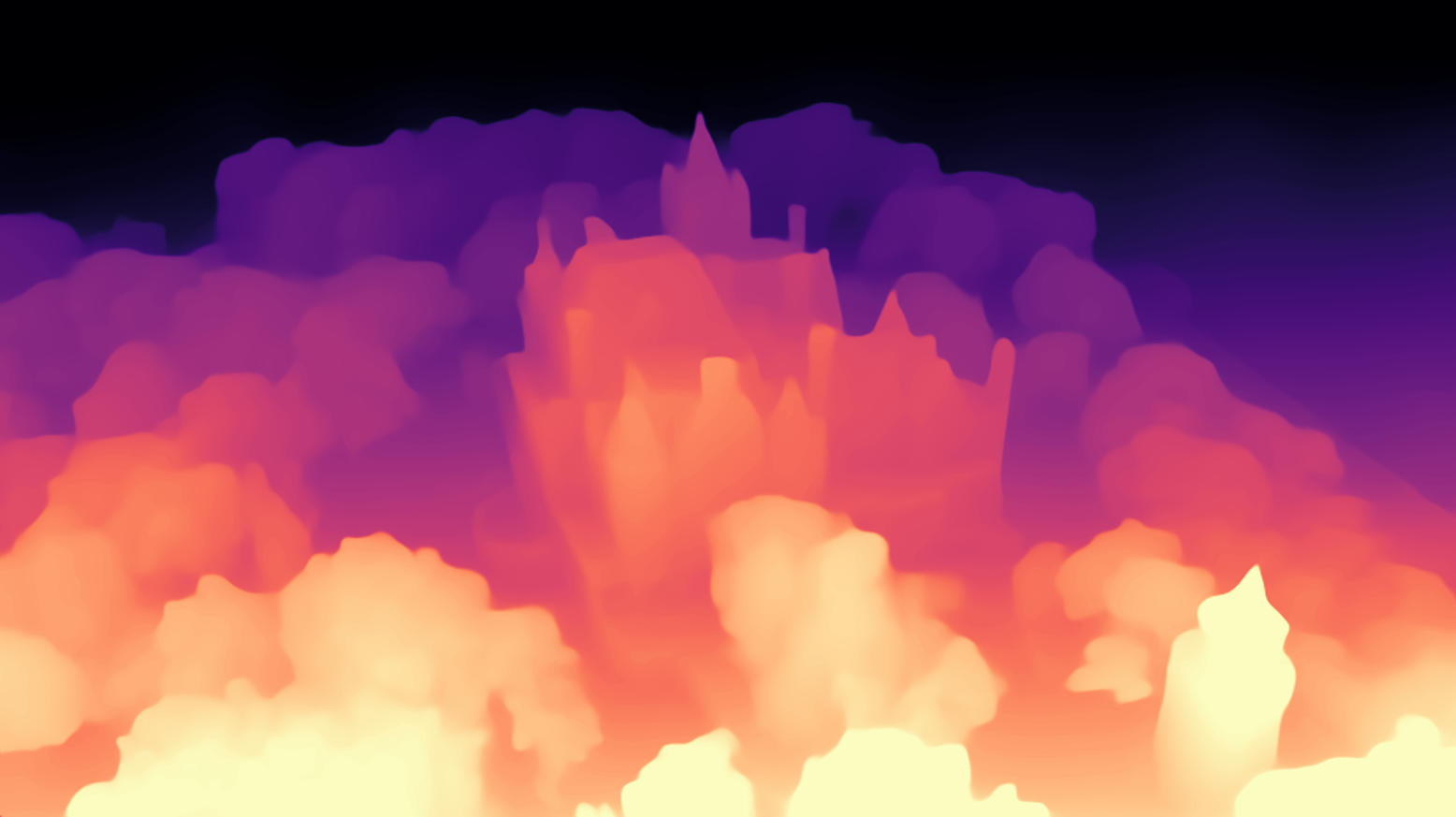} &
\includegraphics[width=0.235\linewidth]{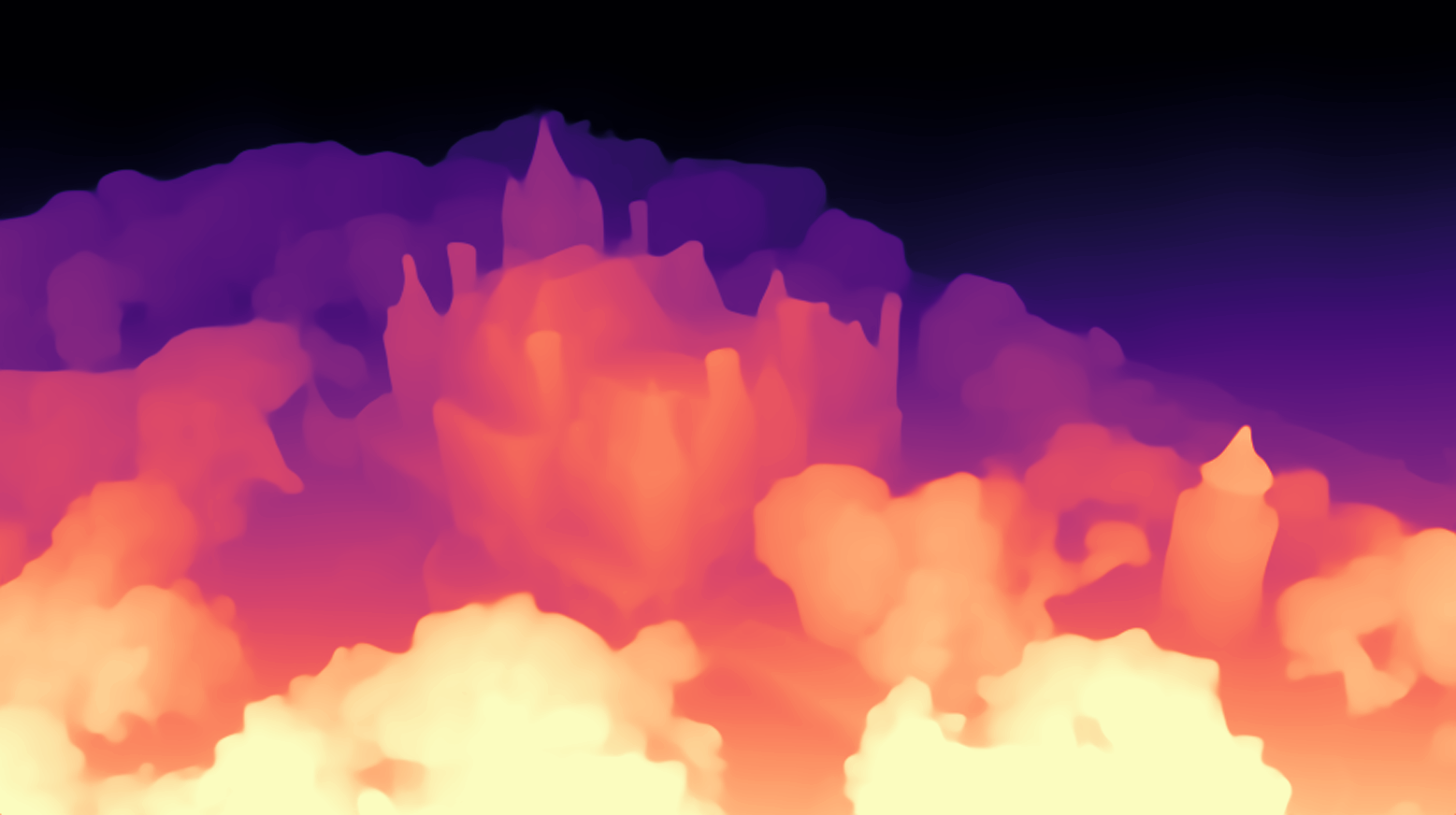} &
\includegraphics[width=0.235\linewidth]{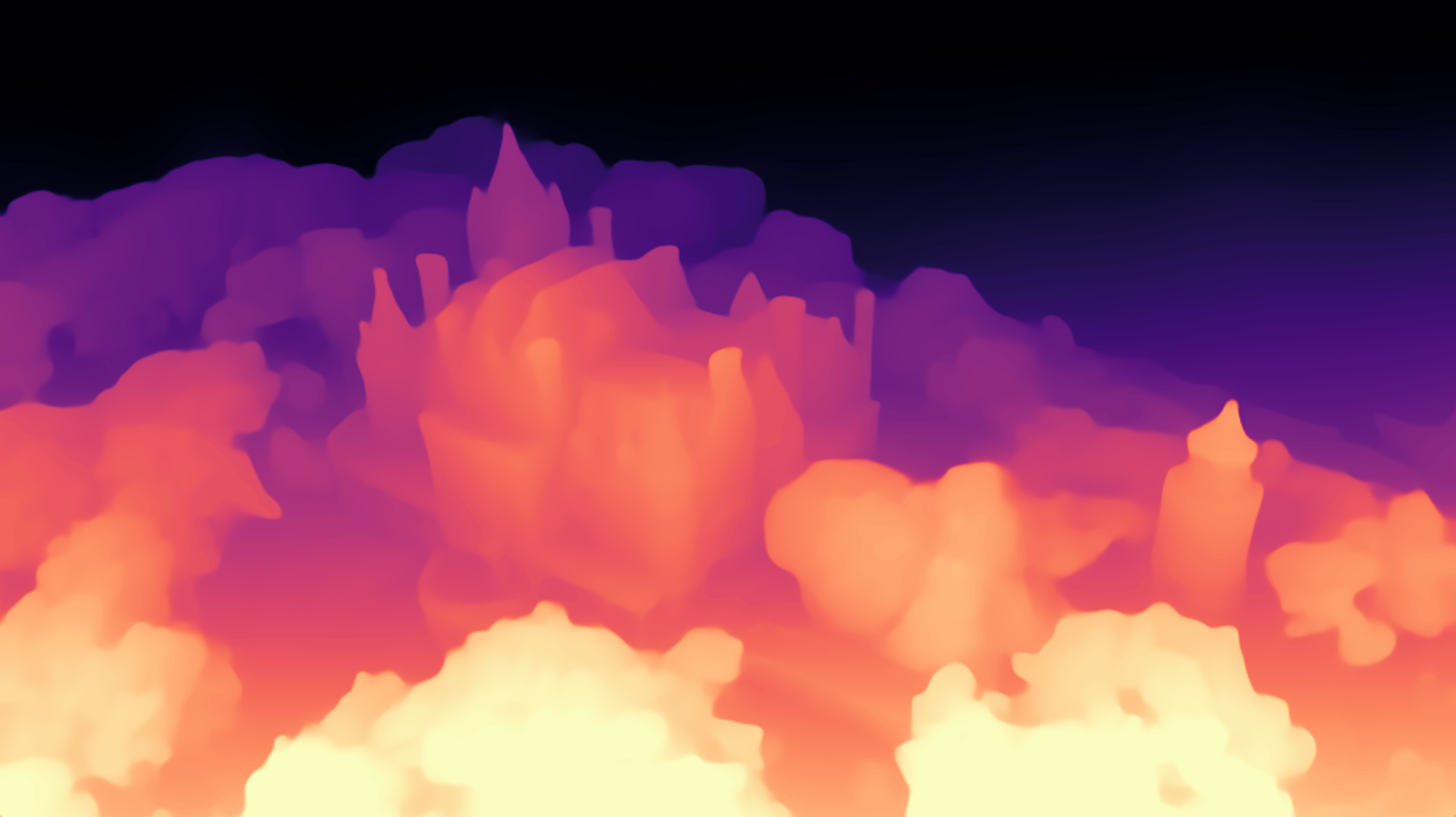}
\end{tabular}

\vspace{2mm}

\includegraphics[width=0.55\linewidth]{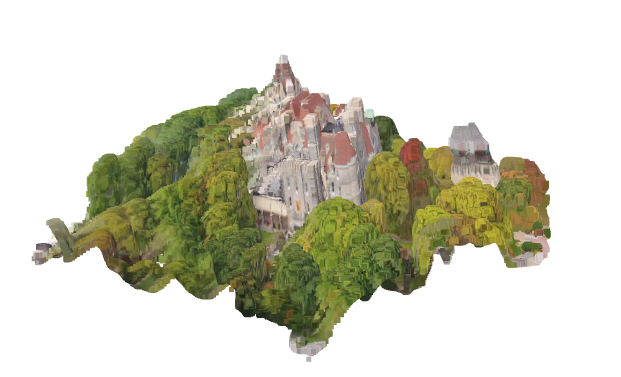}

\caption{We apply SS3D to a casual drone video of a Castle.
\textbf{Top:} predicted depth.
\textbf{Bottom:} the induced 3D reconstruction obtained by fusing the predicted depth with the estimated camera poses and intrinsics.}
\label{fig:qualitative_casual2}
\end{figure}

\begin{figure}[H]
\centering
\setlength{\tabcolsep}{1pt}

\begin{tabular}{cccc}
\includegraphics[width=0.235\linewidth]{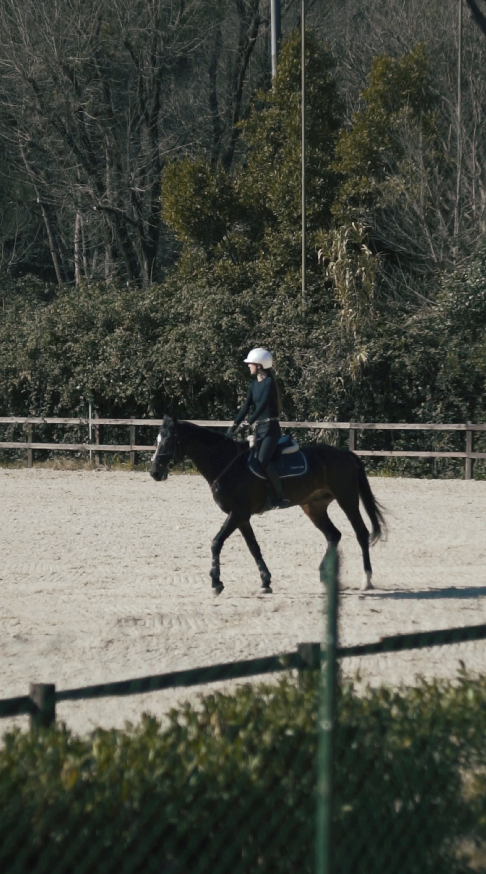} &
\includegraphics[width=0.235\linewidth]{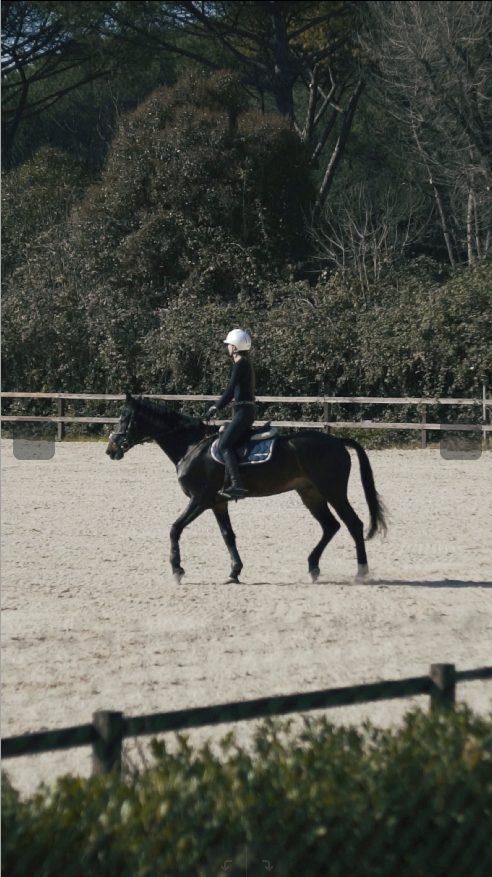} &
\includegraphics[width=0.235\linewidth]{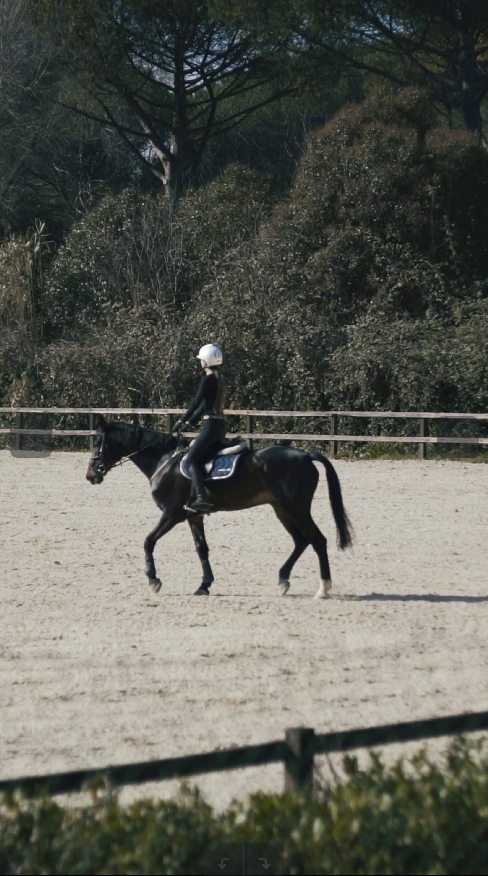} &
\includegraphics[width=0.235\linewidth]{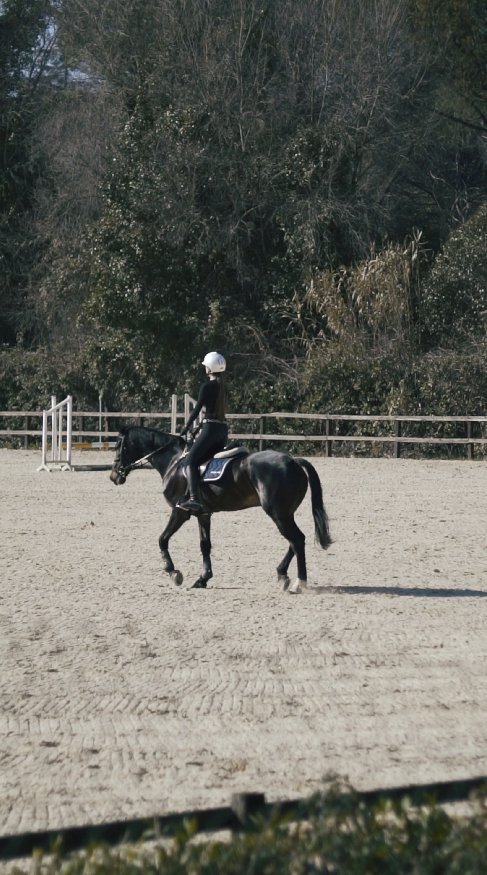} \\[1mm]

\includegraphics[width=0.235\linewidth]{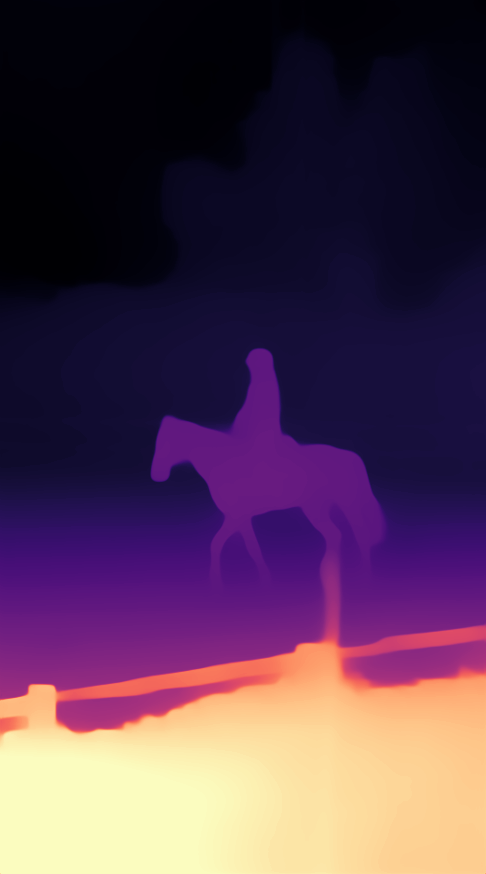} &
\includegraphics[width=0.235\linewidth]{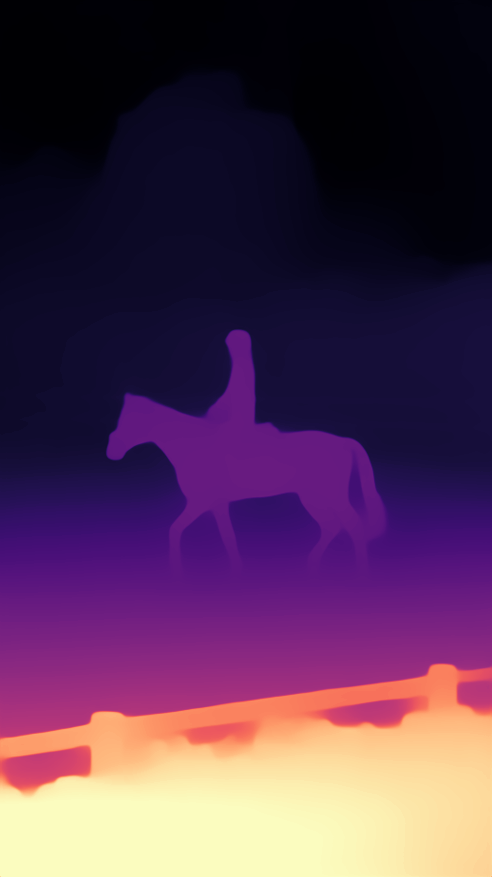} &
\includegraphics[width=0.235\linewidth]{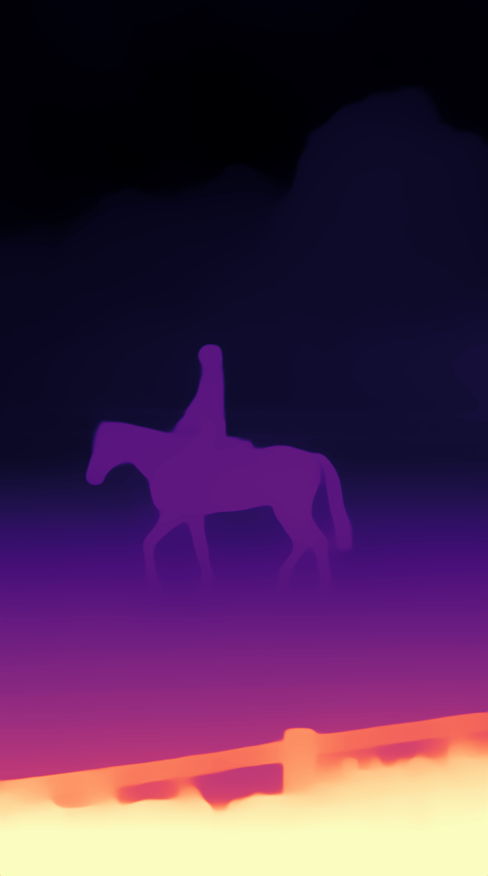} &
\includegraphics[width=0.235\linewidth]{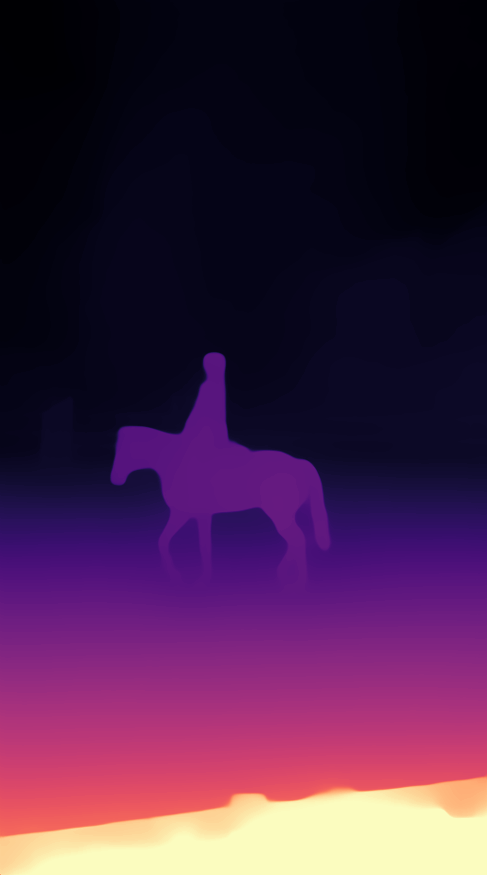}
\end{tabular}

\vspace{2mm}

\includegraphics[width=0.55\linewidth]{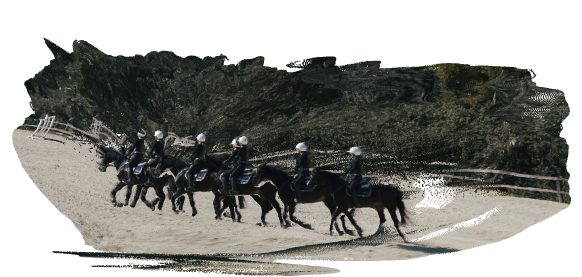}

\caption{We apply SS3D to a casual equestrian video.
\textbf{Top:} predicted depth.
\textbf{Bottom:} the induced 3D reconstruction obtained by fusing the predicted depth with the estimated camera poses and intrinsics.}
\label{fig:qualitative_casual3}
\end{figure}

We compare SS3D against two strong self-supervised baselines, FeatDepth~\cite{shu2020feature} and Monodepth2~\cite{godard2019digging}, on the four sequences introduced in section \ref{sec:append:data}. 
For each sequence, we report both the predicted depth maps and the induced 3D reconstructions obtained by fusing the predicted depth with the estimated camera poses and intrinsics. 
Figs.~\ref{fig:qualitative_seq1}, \ref{fig:qualitative_seq2}, \ref{fig:qualitative_seq3}, and \ref{fig:qualitative_seq4} show that SS3D produces sharper depth and, more importantly, substantially cleaner and more geometrically consistent 3D reconstructions than the baselines. 
Note that SS3D uses its \emph{estimated} intrinsics, whereas FeatDepth and Monodepth2 use the provided calibrated intrinsics.\\
We use a sky segmentation network to mask out sky regions.\\
We additionally present qualitative results on casual in-the-wild videos in Figs.~\ref{fig:qualitative_casual1}, \ref{fig:qualitative_casual2}, and \ref{fig:qualitative_casual3}, illustrating that SS3D generalizes to diverse capture conditions and remains robust outside curated benchmarks.

\section{Data}
\label{sec:append:data}
We focus some of our analyses on a small set of representative KITTI sequences: three from KITTI Raw: 
    (1)~\textbf{2011\_09\_26\_0005} - particularly challenging due to its length and substantial camera rotation, making it prone to drift -
     (2)~\textbf{2011\_09\_26\_0011},
     and (3)~\textbf{2011\_09\_26\_0002}.
And one from KITTI Odometry: \textbf{Sequence 00}.

These sequences are used consistently across our studies of intrinsics stability Figure~\ref{fig:intrinsic_stability},  and our 3D point cloud quantitative evaluation in Table~\ref{tab:seq_results}.

\section{More Implementation details}
\label{sec:supp:implementation_details}

\textbf{Architecture.}
We follow the VGGT backbone design and retain only the depth, pose, and intrinsics heads.
The network consists of 24 attention blocks, each comprising a frame-wise self-attention layer and a global self-attention layer.
Following the ViT-L configuration used in DINOv2, all attention layers operate on 1024-dimensional features with 16 attention heads.
To improve training stability, we use QKNorm and LayerScale in every attention layer, with LayerScale initialized to 0.01.\\
\noindent
For tokenization, we extract DINOv2 features and add positional embeddings.
Multi-scale features from the 4th, 11th, 17th, and 23rd blocks are forwarded to a DPT-style decoder for upsampling.\\

\noindent
\textbf{Video preprocessing.}
Shot boundary detection, frame-rate normalization, and basic frame filtering are performed with PyAV.\\

\noindent
\textbf{Student training.}
Each training batch is constructed by first sampling 4 clusters and then sampling 8 clips from each cluster (total batch size 32).
We balance the photometric and distillation objectives using Pareto-optimal gradient combination following \cite{sener2018multi}.
Images are resized so that the shorter side is 518 pixels, after which we extract a $518 \times 518$ crop.\\

\noindent
\textbf{Hyperparameters.}
We set $\lambda_{\text{distill}} = 0.2$.\\

\noindent
\textbf{Occlusions and dynamic objects.}
We handle occlusions using the masking strategy of \cite{godard2019digging}, and mitigate moving-object artifacts using the rebalancing approach of \cite{hariat2023rebalancing}.

\noindent
\textbf{Clustering.}
We partition the corpus using K-means on clip-level CLIP embeddings, obtained by averaging frame-level CLIP features within each clip.
We use $K=5$ clusters, train one expert per cluster, and distill their predictions into a single student model.\\

\noindent
\textbf{Warm-up.}
In the warm-up phase described in Sec.~3.2, we initialize the depth and pose heads by training on KITTI while keeping intrinsics fixed to the provided calibration.\\

\noindent
\textbf{Intrinsics.} We first predict the two fields of view, \(fov_x\) and \(fov_y\), independent of the resolution, and then derive the focal length:

\[
\begin{aligned}
f_x &= \frac{W/2}{\tan\left(\mathrm{fov}_x/2\right)} \\
f_y &= \frac{H/2}{\tan\left(\mathrm{fov}_y/2\right)}
\end{aligned}
\]

Following common practice in monocular depth and view-synthesis methods, we assume a centered principal point, i.e.,
\[
c_x = W/2, \qquad c_y = H/2 .
\]
The resulting camera intrinsics matrix is therefore
\[
K =
\begin{bmatrix}
f_x & 0 & c_x \\
0 & f_y & c_y \\
0 & 0 & 1
\end{bmatrix}.
\]

\noindent
\textbf{Multi-View Signal Proxy MVSP.}
We use standard SIFT descriptors for local feature extraction and matching. These matches are used only to compute MVSP scores.\\

\textbf{Monocular depth evaluation protocol.}
For KITTI, we follow the standard Eigen split~\cite{eigen2014depth} and evaluate using the commonly adopted monocular depth metrics: AbsRel, SqRel, RMSE, RMSE log, and threshold accuracies \(\delta < 1.25^i\), for \(i \in \{1,2,3\}\).
Predictions are median-scaled to the ground-truth depth before evaluation, following the standard monocular setting.
For NYUv2, we follow the standard indoor monocular depth evaluation protocol.
We use the official training and validation splits, which include 302 and 33 sequences, respectively.
For testing, we use the officially provided 654 images with dense labelled depth maps.
We evaluate using the same monocular depth metrics as KITTI.\\

\textbf{Preprocessing YTB8M.}
\label{sec:supp:ytb_preprocessing}
We rely on a distributed system designed to process millions of YouTube-8M videos efficiently and reliably. All components run inside \textbf{Kubernetes}\footnote{\url{https://kubernetes.io/}}, which simply ensures that the system stays alive, restarts components when needed, and distributes work across machines. New workers are created automatically using \textbf{KEDA}\footnote{\url{https://keda.sh/}}, which increases or decreases the number of processing nodes depending on how many videos are waiting to be processed.
\noindent
Tasks are stored in a \textbf{Kafka}\footnote{\url{https://kafka.apache.org/}} message queue: each message corresponds to one video to download and process. Kafka acts as a buffer that absorbs large fluctuations in workload and guarantees that every video is processed exactly once. Workers consume tasks from Kafka, download the video, extract frames and visual features, and save all results into a shared \textbf{S3-compatible storage}\footnote{\url{https://aws.amazon.com/s3/}}. A small \textbf{Redis}\footnote{\url{https://redis.io/}} database enforces rate-limiting on YouTube downloads, keeps track of which videos have already been processed, which ones failed, and the current progress of each worker. System-level metrics (CPU usage, queue length, failures, throughput) are monitored through \textbf{Prometheus}\footnote{\url{https://prometheus.io/}}, giving a clear real-time picture of the entire pipeline.
\noindent
Workers do not communicate with each other; they only read from Kafka and write to S3. The system scales thus linearly.\\
Each worker operates asynchronously on a subset of videos, performing all local tasks: frame extraction at a fixed stride, feature computation, geometric estimation (fundamental and homography models), and calculation of parallax and quality scores. Workers are dynamically assigned to CPU or GPU machines based on task requirements.\\
Preprocessing proceeds in three stages:
\begin{enumerate}
    \item A global pass that computes MVSP scores (defined in equation 12), along with CLIP embeddings; afterwards, the raw videos are discarded to save storage;
    \item Clustering and statistical analysis;
    \item Preprocessing and persistent downloading of the selected videos according to the curriculum percentile schedule.
\end{enumerate}
The pipeline includes several safety mechanisms to ensure reliability at scale.\\
First, Kafka topics include a \textbf{Dead Letter Queue (DLQ)}: if a video repeatedly fails (for example, because it is corrupted or suddenly unavailable), it is automatically redirected to this DLQ so that the rest of the pipeline keeps running smoothly. All downloads use retry strategies with exponential backoff and checksum verification, and we explicitly handle \textbf{YouTube rate limits} by respecting ``retry-after'' signals and distributing requests across workers to avoid temporary bans.\\
To avoid partial results or inconsistent outputs, every worker writes to storage using a simple \emph{atomic commit}: intermediate files are saved under a temporary name and only renamed once processing succeeds. If a worker crashes, Kubernetes restarts it automatically, and Kafka simply reassigns the unfinished video to another node. KEDA prevents overload by automatically increasing the number of workers when the queue grows and shrinking it when the workload decreases.\\
Together, these strategies ensure a stable, predictable, and fully fault-tolerant preprocessing pipeline that can run for days or weeks without interruption, even under high traffic, corrupted videos, or sudden spikes in workload.

\section{More results}
\noindent
\subsection{Effect of distillation}
We isolate expert distillation on a standard SfM-based reprojection baseline (ResNet-50, in-domain on KITTI/NYUv2) using \(K=5\) experts clustered within the target dataset. Importantly, we do not use any other component of SS3D: no unified transformer architecture and no MVSP-based filtering/curriculum and still improve a strong baseline~\cite{hariat2025improved} (Tab.~\ref{table:ablation_clip_distill_kitti_nyu_metrics}), showing our gains come from the training strategy, not web-scale data.

\begin{table}[ht]
\centering
\resizebox{\linewidth}{!}{%
\renewcommand{\arraystretch}{1.25}
\fontsize{6}{8}\selectfont
\setlength{\tabcolsep}{5pt}

\newcommand{\gcheck}{\textcolor{green!60!black}{\ensuremath{\checkmark}}}

\begin{tabular}{c  r r r  r r r}
\hline
\rowcolor{headergray}
\multicolumn{1}{c}{\textbf{Loss components}} &
\multicolumn{6}{c}{\textbf{Metrics}} \\
\cline{1-1}\cline{2-7}

\rowcolor{headergray}
$\mathcal{L}_{\text{distill}}$ &
\multicolumn{3}{c}{\textbf{KITTI}} &
\multicolumn{3}{c}{\textbf{NYU}} \\
\cline{2-4}\cline{5-7}

\rowcolor{headergray}
 &
\textcolor{lightblue}{\textbf{Abs Rel $\downarrow$}} &
\textcolor{lightblue}{\textbf{RMSE $\downarrow$}} &
\textcolor{lightblue}{\textbf{$\delta_1 \uparrow$}} &
\textcolor{lightblue}{\textbf{Abs Rel $\downarrow$}} &
\textcolor{lightblue}{\textbf{RMSE $\downarrow$}} &
\textcolor{lightblue}{\textbf{$\delta_1 \uparrow$}} \\
\hline

\multicolumn{7}{l}{\textbf{Baseline}} \\[-2pt]
\hline
 & 0.104 & 0.180 & 0.885 & 0.115 & 0.458 & 0.859 \\
\hline

\multicolumn{7}{l}{\textbf{Adding distillation from contextual experts}} \\[-2pt]
\hline
\gcheck & 0.098 & 0.174 & 0.892 & 0.109 & 0.449 & 0.860 \\
\hline
\end{tabular}%
}
\caption{\textbf{Ablation of expert distillation} on a strong SfM-based reprojection baseline \cite{hariat2025improved} (ResNet-50, KITTI/NYUv2). Using \(K=5\). Without other SS3D components.}
\label{table:ablation_clip_distill_kitti_nyu_metrics}
\end{table}

\subsection{Odometry}

\begin{figure}[ht]
  \begin{tabular}{c|c}
     \includegraphics[width=0.5\linewidth]{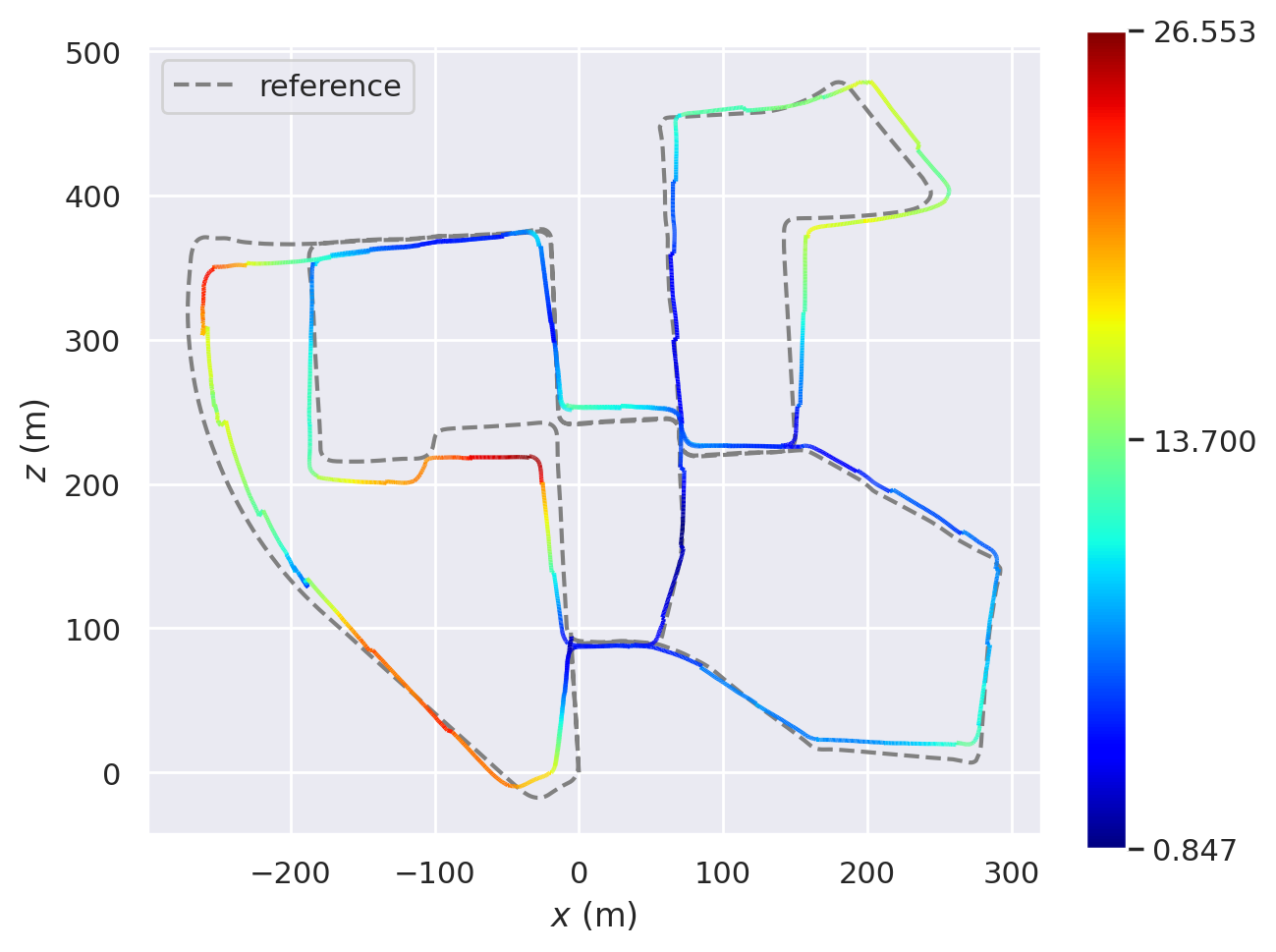} &  
     \includegraphics[width=0.5\linewidth]{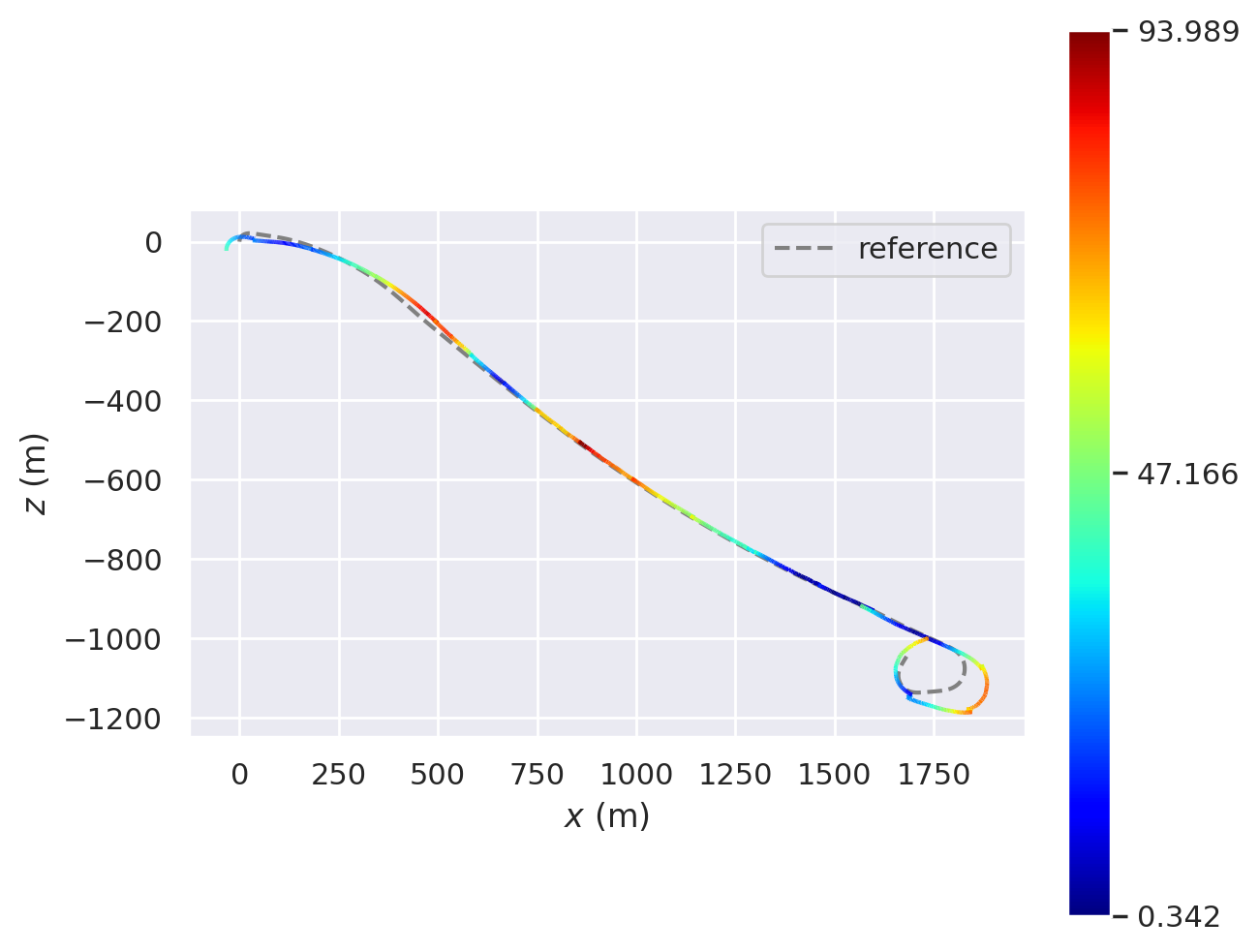} \\
     Seq.00 (3724m) & Seq.01 (2453m)\\
     \includegraphics[width=0.5\linewidth]{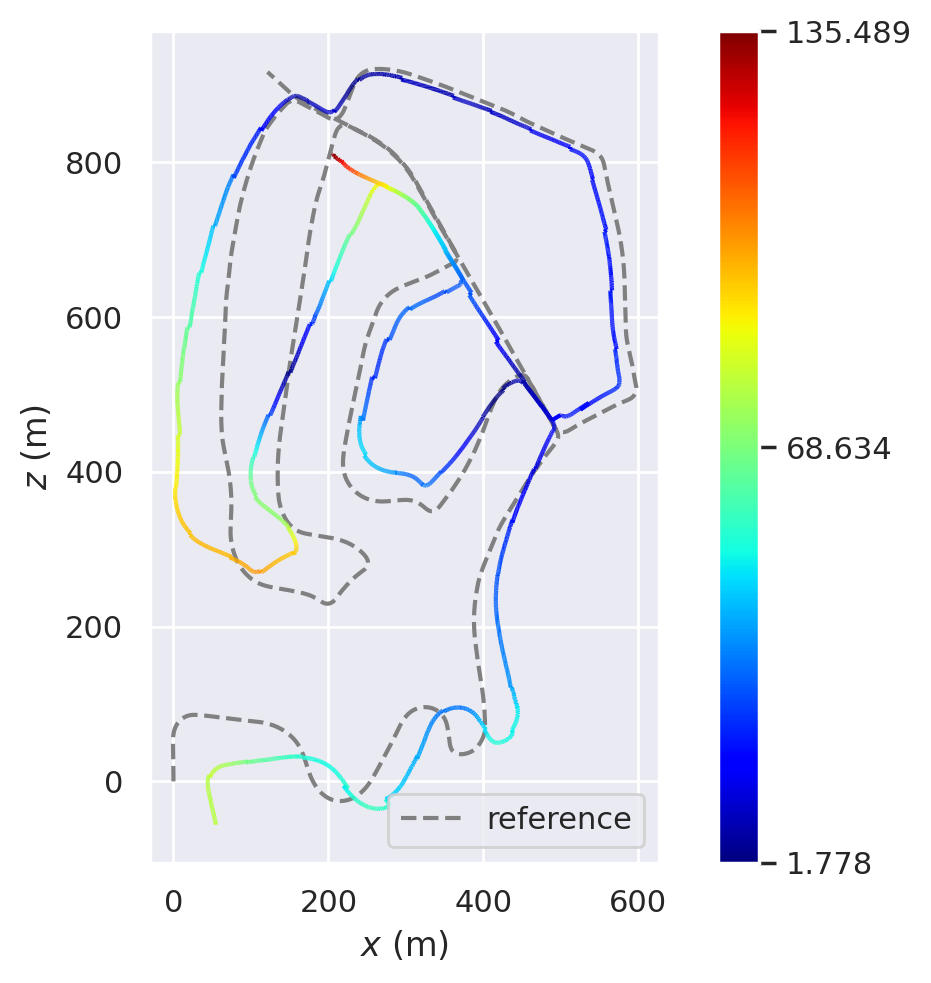} &  
     \includegraphics[width=0.5\linewidth]{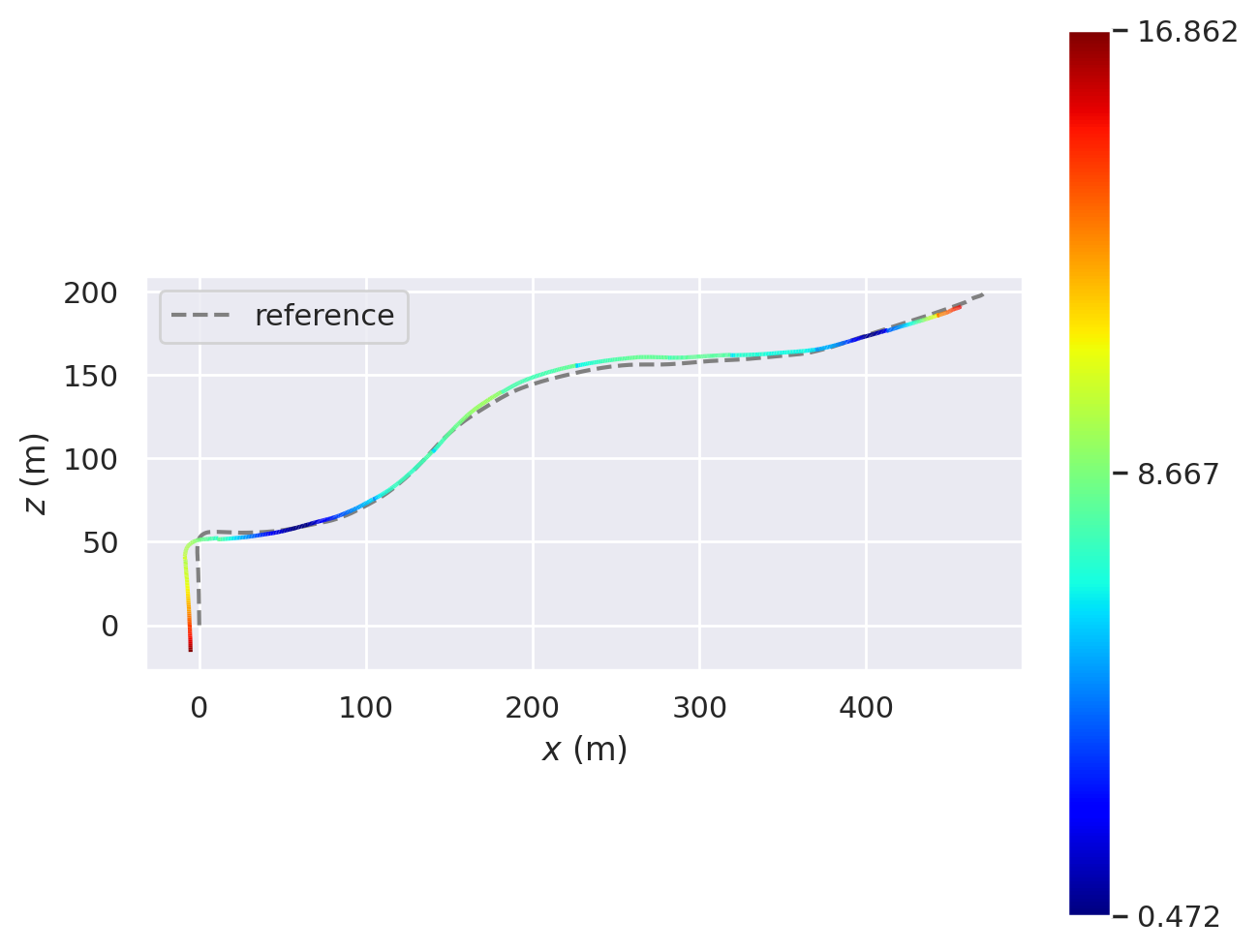} \\
     Seq.02 (5067m) & Seq.03 (561m)
  \end{tabular}
 \caption{{\bf Trajectory plots} for Sequences 00 to 03 of KITTI. The ramp on the right shows metric error scales.}
    \label{fig:odometry_trajectories1}  
\end{figure}

\begin{figure}[ht]
  \begin{tabular}{c|c}
     \includegraphics[width=0.5\linewidth]{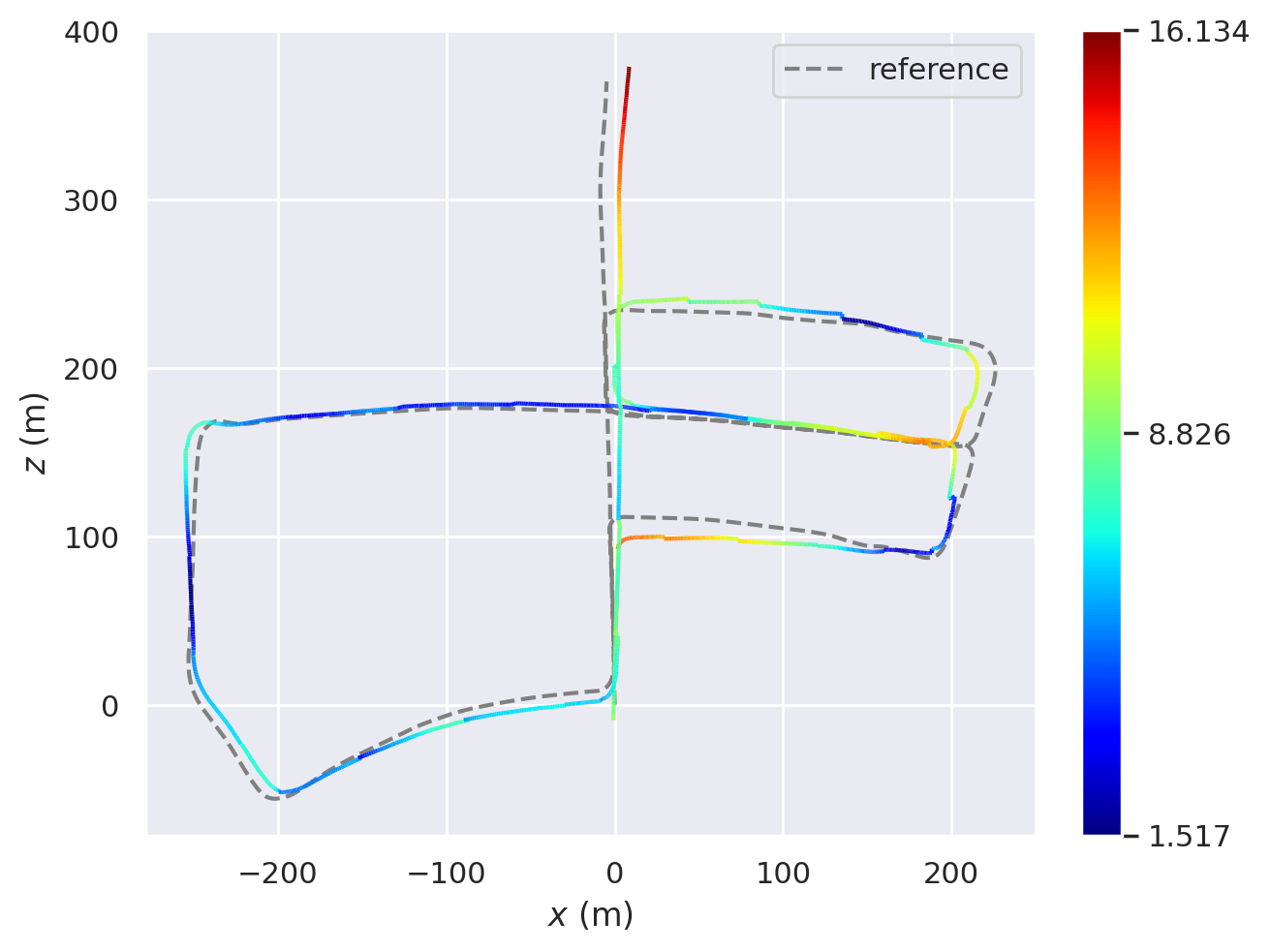} &  
     \includegraphics[width=0.5\linewidth]{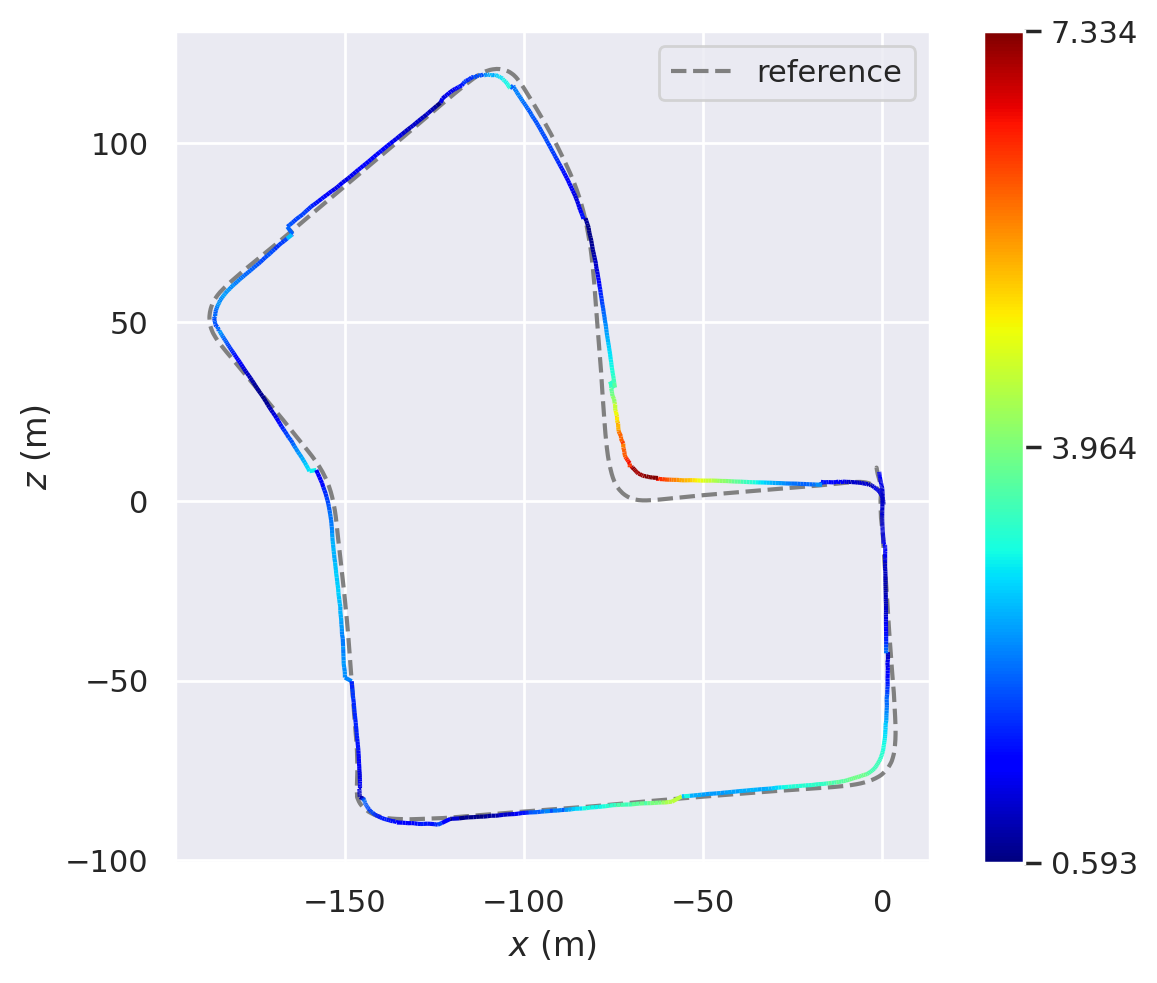} \\
     Seq.05 (2206m) & Seq.07 (650m)\\
     \includegraphics[width=0.5\linewidth]{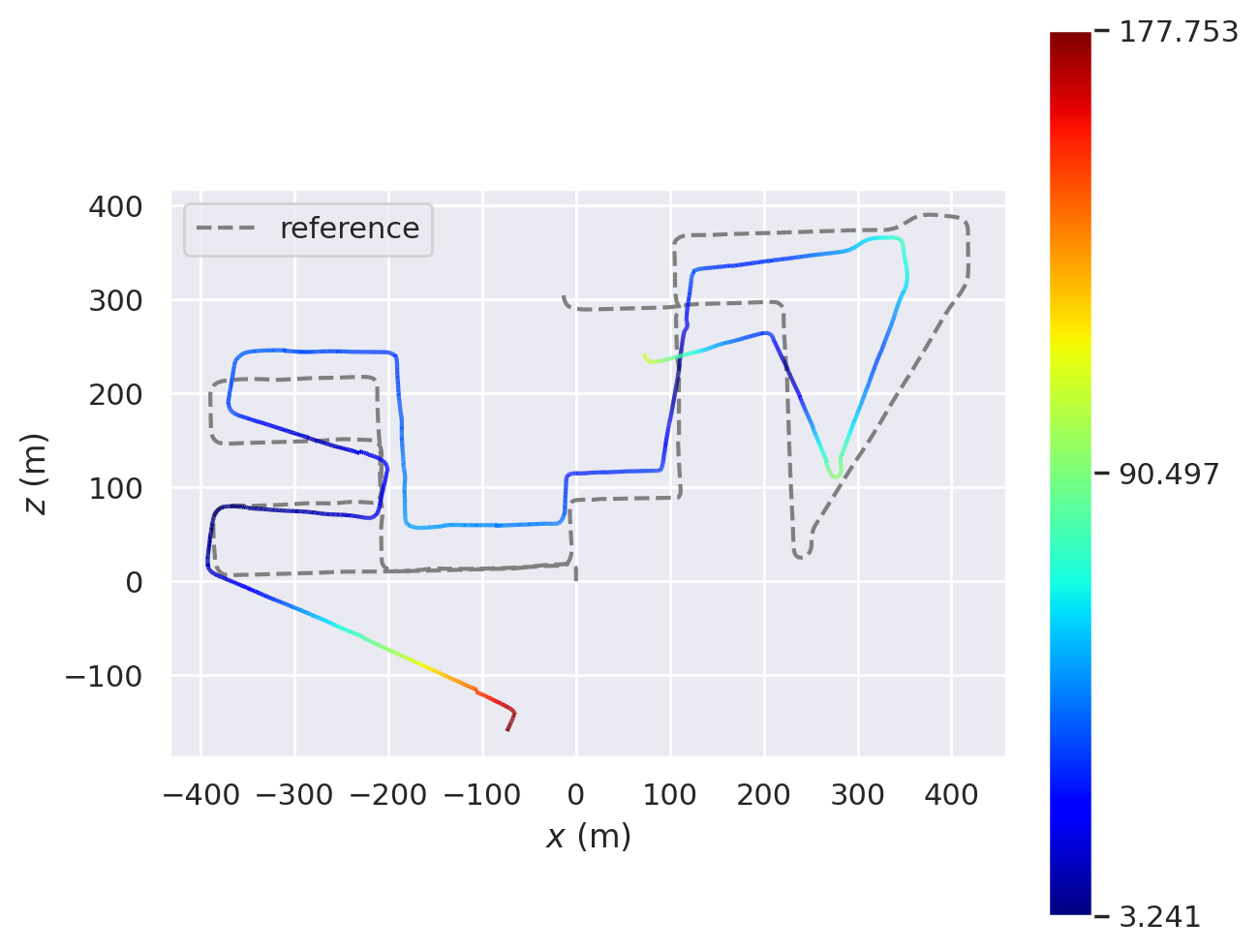} &  
     \includegraphics[width=0.5\linewidth]{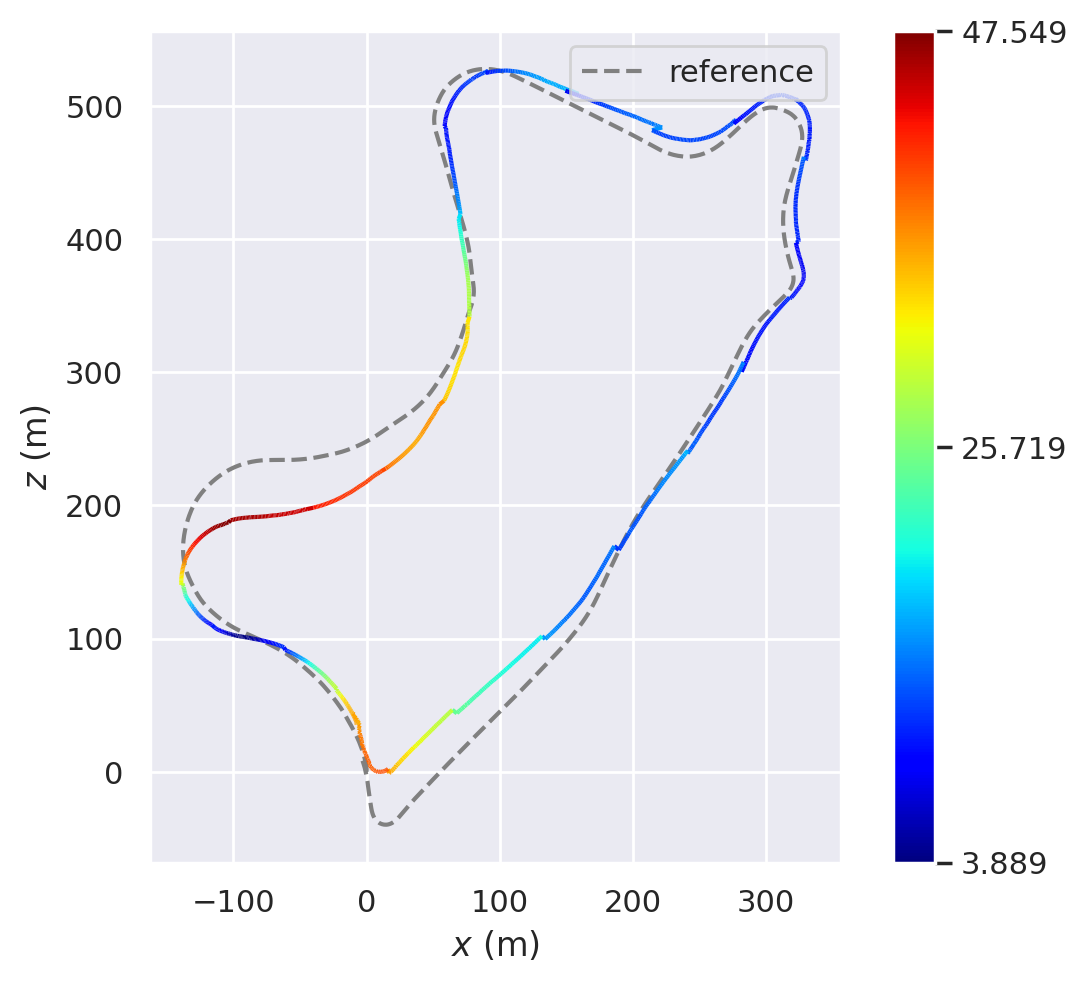} \\
     Seq.08 (3223m) & Seq.09 (1705m)
  \end{tabular}
 \caption{{\bf Trajectory plots} for Sequences 05, 07, 08 and 09 of KITTI. The ramp on the right shows metric error scales. Note that  Sequences 04 and 06 are discarded as very easy (straight line).}
    \label{fig:odometry_trajectories2}  
\end{figure}

In Figures~\ref{fig:odometry_trajectories1} and~\ref{fig:odometry_trajectories2}, we visualize the predicted trajectories on eight of the first ten sequences of the KITTI Odometry dataset. These results show that \textbf{SS3D} accurately captures the overall camera trajectory, even on long sequences spanning several kilometers.

\subsection{Intrinsics}

Tab.~\ref{tab:afe_rfe} compare our intrinsics estimation with the state of the art approaches.

\begin{table}[t]
\centering

\begin{minipage}[t]{0.75\linewidth}
    \centering
    \small
    \renewcommand{\arraystretch}{1.15}
    \begin{tabular}{lcc}
        \toprule
        \textbf{Method} & \textbf{AFE (px)$\downarrow$} & \textbf{RFE (\%)$\downarrow$} \\
        \midrule
        UniDepth & 447.4 & 0.357 \\
        Dust3r   & 434.0 & 0.364 \\
        AnyCam   & \textbf{252.2} & 0.181 \\
        \textbf{Ours} & 256.6 & \textbf{0.167} \\
        \bottomrule
    \end{tabular}
    \caption{\textbf{Intrinsic parameter estimation on Sintel.}
    Mean absolute focal error (AFE) and mean relative focal error (RFE).}
    \label{tab:afe_rfe}
\end{minipage}

\end{table}

\subsection{Point cloud quality evaluation}
\label{sec:app:pointcloudquality}
To quantitatively evaluate the reconstructed 3D geometry under global scale ambiguity, we first normalize the predicted and ground-truth point clouds independently by subtracting their centroids and dividing by a single global scale factor, defined as the root-mean-square distance of the points from the centroid. We then perform ICP in the normalized space to align the predicted point cloud with the ground truth. Finally, the aligned prediction is mapped back to the ground-truth coordinate system using the corresponding ground-truth centroid and scale, and all evaluation metrics are computed in that space. We finally compute the bidirectional RMSE, i.e., the nearest-neighbor RMSE from prediction to ground truth and from ground truth to prediction, and report their average as the final reconstruction error. Results are shown in Tab.~\ref{tab:seq_results}.